\documentclass{article} 
\usepackage{iclr2027_conference,times}

\usepackage{amsmath,amsfonts,bm}

\def\eqref#1{equation~\ref{#1}}

\def\1{\bm{1}}

\DeclareMathAlphabet{\mathsfit}{\encodingdefault}{\sfdefault}{m}{sl}
\SetMathAlphabet{\mathsfit}{bold}{\encodingdefault}{\sfdefault}{bx}{n}

\usepackage{hyperref}
\hypersetup{
  pdftitle={ReAP: From Visual Failure Recognition to Action-Conditioned Risk Prediction for Runtime Failure Monitoring},
  pdfauthor={Jingyuan Liu, Yunpeng Qing, Changxu Wei, Zhi Hou}
}
\usepackage{url}

\usepackage{tabularx}
\usepackage{array}
\usepackage{booktabs}
\usepackage{multirow}
\usepackage{xcolor}
\usepackage{graphicx}
\usepackage[table]{xcolor}
\usepackage{makecell}
\usepackage{bm}
\usepackage{adjustbox}
\usepackage{subcaption}
\usepackage{wrapfig}
\usepackage{capt-of}
\usepackage{placeins}
\usepackage{amsmath}
\usepackage{bbm}

\newcommand{\res}[2]{\ensuremath{#1^{\scriptscriptstyle \pm #2}}}
\newcommand{\bestres}[2]{\ensuremath{\bm{#1}^{\scriptscriptstyle \pm #2}}}
\newcommand{\secondres}[2]{\ensuremath{\underline{#1}^{\scriptscriptstyle \pm #2}}}

\title{ReAP: From Visual Failure Recognition to Action-Conditioned Risk Prediction for Runtime Failure Monitoring}

\author{
Jingyuan Liu$^{1,2}$\thanks{Email:
\texttt{23302010010@m.fudan.edu.cn}} \quad
Yunpeng Qing$^{1,3}$ \quad
Changxu Wei$^{1,4}$ \quad
Zhi Hou$^{1}$\\[0.35em]
{\normalfont\small
$^{1}$ ACE Robotics
$^{2}$ Fudan University
$^{3}$ Zhejiang University
$^{4}$ PolyU
}
}

\newcommand{\methodname}{ReAP}

\iclrfinalcopy 
\begin{document}

\maketitle
\lhead{Preprint}

\begin{abstract}
Despite rapid progress, generalist robot policies remain brittle during execution. Reliable deployment therefore requires runtime monitoring that provides timely warnings of failure. Existing monitors commonly estimate risk from current observations or policy signals without explicitly predicting the future consequences of proposed actions. To anticipate such risks before action execution, we introduce ReAP, a future-aware runtime monitoring framework that combines visual failure recognition with action-conditioned failure prediction. Specifically, ReAP first trains a visual detector on successful and failed rollouts to learn a failure-relevant representation space and estimate current risk. An action-conditioned predictor then forecasts the resulting future representation in this space to estimate future failure risk. At runtime, the detector and predictor provide complementary signals for current-state failure recognition and future-aware risk assessment. Experiments across diverse robot policies in simulation and the real world demonstrate that ReAP effectively balances failure-detection accuracy and warning timeliness.
\end{abstract}

\section{Introduction}

\begin{wrapfigure}{r}{0.54\columnwidth}
    \centering
    \vspace{-12pt}
    \includegraphics[width=\linewidth]{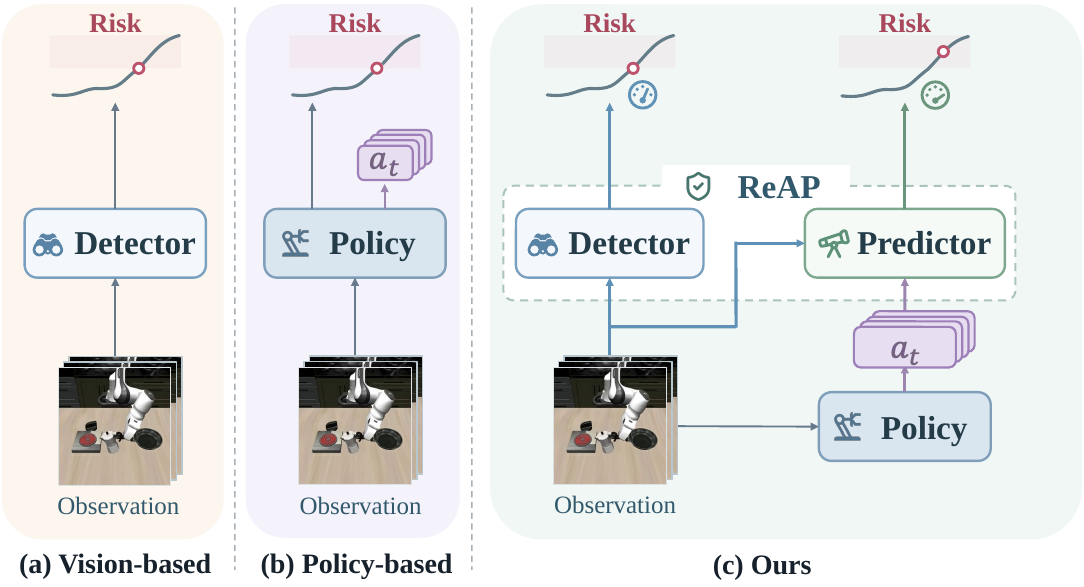}
    \caption{
    Vision-based monitors estimate failure risk from the current observation,
    while policy-based monitors rely on signals from the deployed policy.
    \methodname{} retains observation-based detection for assessing current risk
    and augments it with action-conditioned prediction to evaluate the future
    risk induced by a proposed action before execution.
    }
    \label{fig:overview}
    \vspace{-8pt}
\end{wrapfigure}

Recent advances in generalist embodied models and policies have substantially expanded the range of manipulation tasks that robots can perform across diverse environments and embodiments~\citep{zitkovich2023rt2,kim2025openvla,
black2024pi0,nvidia2026cosmos3}.
Despite this progress, learned policies remain vulnerable to unexpected visual changes, imperfect contacts, and compounding action errors, where seemingly small execution deviations can quickly lead to irreversible task failures
~\citep{agia2025sentinel}.
Runtime failure monitoring is therefore critical for reliable deployment.
Timely warnings allow a robot to stop an unsafe action, replan, invoke a fallback strategy, or request human intervention before a recoverable deviation escalates into an irreversible task failure.

Existing runtime monitors use different signals to estimate failure risk.
Vision-based methods~(Figure~\hyperref[fig:overview]{\ref*{fig:overview}a}) assess failure from the observed execution state~\citep{duan2025aha, zhou2026rcnf}, while policy-based methods (Figure~\hyperref[fig:overview]{\ref*{fig:overview}b}) exploit signals such as policy-internal representations~\citep{gu2025safe,park2026hideandseek} or uncertainty in the action-generation process~\citep{lee2025diffdagger,rao2026accel}.
These approaches largely assess failure risk from evidence available at the current decision state.
However, an execution may still appear nominal even when the proposed action would lead to failure, with the resulting deviation becoming visually apparent only after the action begins. Even from the same state, different actions may steer the execution toward diverse future outcomes, such as recovery or further deviation.
Thus, runtime monitoring calls for explicitly looking beyond current evidence and assessing the failure risk of the predicted future state before execution.

Motivated by this insight, we introduce \textbf{\methodname{}}, a future-aware runtime monitoring framework combining visual failure \underline{Re}cognition with \underline{A}ction-conditioned risk \underline{P}rediction. 
The key idea is to estimate the future failure risk through action-conditioned prediction in a compact visual representation space learned for failure recognition.
Specifically, we first train a visual detector on successful and
failed rollouts to learn this representation that distinguishes successful
and failed observations. 
Building on this representation, we follow the JEPA-style latent prediction paradigm~\citep{assran2023ijepa} to train an action-conditioned predictor.
Given the current visual representation and a proposed action chunk, the predictor models its consequences by estimating the detector representation of the resulting future observation.
The frozen detector head is then reused to map this prediction to a future failure-risk score.
As illustrated in Figure~\hyperref[fig:overview]{\ref*{fig:overview}c}, the detector assesses the observed state, whereas the predictor evaluates the proposed action before its outcome becomes observable.

We evaluate \methodname{} across four simulated policy--benchmark settings, covering OpenVLA~\citep{kim2025openvla} and $\pi_0$~\citep{black2024pi0} on LIBERO-10~\citep{liu2023libero}, and $\pi_{0.5}$~\citep{physicalintelligence2025pi05} and Fast-WAM~\citep{yuan2026fastwam} on eight selected RoboTwin tasks~\citep{chen2025robotwin}. 
We further deploy the monitor on a real robot across three manipulation tasks using $\pi_{0.5}$. 
Across all four simulation settings, our combined monitor consistently outperforms existing baselines, with relative improvements
of up to \bm{$12.7\%$} over the strongest evaluated baseline.
Real-robot experiments further demonstrate the effectiveness of our \methodname{} in runtime monitoring with low inference overhead. To summarize, our main contributions are as follows:
\begin{enumerate}
    \item We highlight that many runtime monitors rely only on current execution signals, limiting their ability to predict failures arising from the future consequences of proposed actions.
    
    \item We introduce \methodname{}, a future-aware runtime monitoring framework that learns a failure-relevant visual representation and uses action-conditioned latent prediction to estimate the future failure risk associated with proposed actions (Section~\ref{sec:method}).

    \item We demonstrate strong monitoring performance across diverse simulated
    policy--benchmark settings (Section~\ref{sec:offline}) and real-world manipulation tasks (Section~\ref{sec:online}), together with
    detailed analyses of the detector and predictor components (Section~\ref{sec:analysis}).
\end{enumerate}

\section{Related Work}
\label{sec:related_work}

\paragraph{Runtime Failure Monitoring}

Existing runtime failure monitors estimate risk from different execution-time signals. Based on how these signals are processed to assess failure risk, existing
approaches can be broadly grouped into four families.
OOD-based methods detect deviations from successful or in-distribution behavior
\citep{burda2019exploration,wong2022erroraware,xu2025faildetect,zhou2026rcnf}.
Action-uncertainty methods use signals from generated actions or the action-generation process,
including action resampling, diffusion objectives, and flow-matching dynamics
\citep{lee2025diffdagger,rao2026accel}.
Supervised detectors learn failure scores from labeled executions using visual, temporal, or policy
representations \citep{farid2022failure,gu2025safe,park2026hideandseek}.
Foundation-model-based approaches use pretrained language or vision-language models for semantic
failure detection and reasoning \citep{sinha2024aesop,duan2025aha,pacaud2025guardian,lin2025failsafe}.
Hybrid methods combine multiple signals. FIPER~\citep{romer2025fiper} combines observation novelty
with action uncertainty, while SENTINEL~\citep{agia2025sentinel} combines action uncertainty with
VLM-based progress monitoring.

A separate line of predictive monitoring models future states under proposed
actions. ~\citet{liu2024modelbased} simulate future policy rollouts
with latent dynamics. Concurrent works include Foresight~\citep{zhang2026foresight}, which uses
action-conditioned world-model features for failure detection, and
ContactGuard~\citep{zheng2026contactguard}, which scores predicted post-contact
latents with a failure probe. In comparison, \methodname{} first learns a failure-relevant representation space and models action consequences in this space to estimate future failure risk, considering both current visual signals and future action risks.
Further comparisons are provided in Appendix~\ref{app:concurrent_world_model}.

\paragraph{Test-Time Action Evaluation.}
A growing body of work evaluates candidate actions proposed by robot policies before execution.
Recent approaches use learned verifiers, reward models, history-aware verification, or model-internal confidence signals
to score and select among multiple action proposals
\citep{kwok2025robomonkey,dai2025rover,li2025have,
yang2025taco,jang2026mgselect}.
These methods score candidate actions without explicitly predicting their
future consequences.
A complementary line of work predicts the consequences of candidate actions before evaluation. FOREWARN~\citep{wu2025forewarn} uses a latent world
model to imagine candidate outcomes and aligns the predicted latent states with
a VLM for consequence evaluation. DynaGuide~\citep{west2025dynaguide} predicts
future latent states and uses their similarity to desired or undesired outcomes
to guide action generation, while Do What You Say~\citep{wu2026dowhatyousay}
forward-simulates candidate action sequences and evaluates whether the predicted
outcomes agree with the policy's textual reasoning. 
Our method is most closely related to this predictive line, as we use this predictive
capability as a runtime monitoring signal by evaluating the estimated representation in
the failure-relevant space learned by the visual detector to assess risk before action execution.

\section{Problem Setup}
\label{sec:problem}

We consider runtime failure monitoring for robotic manipulation policies.
When queried at environment timestep $t$, the deployed policy receives
the visual observation $O_t$ and proposes an action chunk $A_t^\pi$
of length $H_\pi$.
The observation-based detector estimates current-state risk
$r_t^{\mathrm{det}}$, while the action-conditioned predictor estimates
future risk $r_{t+H|t}^{\mathrm{pred}}$ associated with executing the first $H$ actions of $A_t^\pi$, denoted by $A_t=(a_{t|t},\ldots,a_{t+H-1|t})$, where $H$ is the prediction horizon.
Together, they support failure recognition from observed states and
anticipation of risky action consequences before execution. The monitoring objective is to detect failed executions as early as possible while limiting false alarms on successful rollouts.

Both models are trained on successful and failed rollouts collected from the policy, using only episode-level outcomes.
For rollout $i$ of length $T_i$, the risk target is $y_{i,t}=0$
if the rollout succeeds and $y_{i,t}=t/(T_i-1)$ if it fails,
where $t\in\{0,\ldots,T_i-1\}$.
This weak supervision avoids failure-onset annotation and assigns lower risk to early frames that may
still appear nominal, with the target increasing to one at the end of a failed rollout.



\begin{figure}[tbp]
    \centering
    \includegraphics[width=\linewidth]{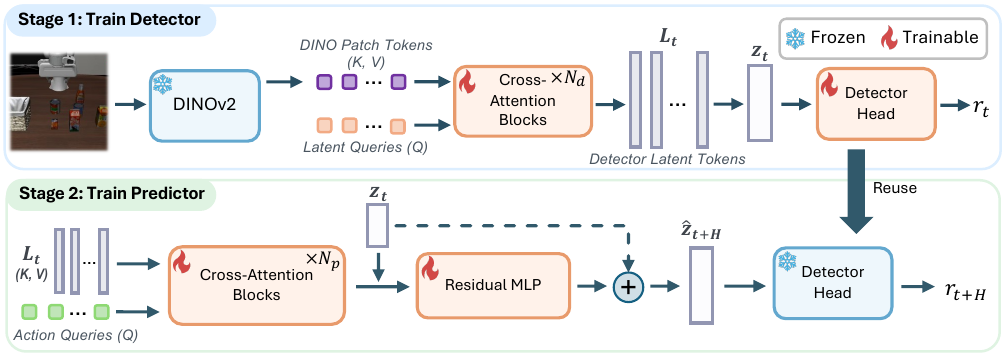}
    \caption{
    Overview of \methodname{}. Stage~1 learns an observation-based failure
    representation and detector. Stage~2 freezes the detector and learns an
    action-conditioned transition model that predicts future representations,
    which are evaluated by the same detector head.
    }
    \label{fig:pipeline}
\end{figure}

\section{Methods}
\label{sec:method}

As shown in Figure~\ref{fig:pipeline}, our framework is trained in two stages.
In Stage~1, an observation-only failure detector learns a failure-relevant
visual representation using episode-level outcome supervision and maps it to a scalar risk score. In Stage~2, an
action-conditioned transition model is trained in the latent space learned in
Stage~1 to predict how the detector representation changes after executing a
candidate action chunk. The predicted future representation is evaluated by the
same frozen detector head, keeping the risk semantics of observed and predicted
representations fixed. After training, thresholds for the detector and predictor
are calibrated separately on held-out successful rollouts.

\subsection{Observation-Only Failure Representation Learning}
\label{sec:detector}

Stage~1 learns a policy-external representation of failure from the current visual observation
$O_t$, without accessing candidate actions or internal policy features.
Each observation may contain one or more synchronized camera views.
For each camera view, dense patch tokens are extracted using a frozen DINOv2~\citep{oquab2023dinov2} backbone.
The tokens are projected into a shared feature space and augmented with camera-specific and
spatial positional embeddings.
Tokens from all views are then concatenated to form the visual tokens of $O_t$.

To compress the dense visual features into a failure-relevant representation, we use a
Perceiver-style feature extractor~\citep{jaegle2021perceiver} with a fixed set of $M$
learnable latent queries.
The latent queries aggregate visual information through cross-attention, using the projected DINOv2 patch tokens as keys and values.
This produces a fixed set of \(M\) detector latent tokens, denoted by \(L_t\).
We use $M=32$ in all experiments.
The detector latent tokens are first mean-pooled into a single vector that summarizes the current observation. A learned linear projection then maps this vector to the compact state representation $z_t$. Finally, the detector head maps $z_t$ to the observation-level risk score
$r_t^{\mathrm{det}} \in [0,1]$.

During Stage~1 training, the DINOv2 backbone remains frozen, while the visual
projection, Perceiver-style feature extractor, and detector head are optimized using failure
supervision. To train the detector, we use a class-weighted binary
cross-entropy loss over valid observation timesteps:
\begin{equation}
\mathcal{L}_{\mathrm{det}}
=
-
\sum_i\sum_t
\left[
w_{+}y_{i,t}\log r_{i,t}^{\mathrm{det}}
+
w_{-}(1-y_{i,t})\log(1-r_{i,t}^{\mathrm{det}})
\right],
\label{eq:detector_loss}
\end{equation}
where $y_{i,t}$ is the risk target defined in Section~\ref{sec:problem}, and $w_{+}$ and
$w_{-}$ compensate for class imbalance.

\subsection{Action-Conditioned Latent Dynamics in Failure Space}
\label{sec:predictor}

Stage~2 models how a candidate action chunk changes the failure-relevant
representation learned in Stage~1. Inspired by JEPA-style latent prediction
\citep{assran2023ijepa,bardes2024vjepa}, we predict the representation of the
future observation directly in the learned latent space rather than
reconstructing the observation itself. The complete Stage~1 detector is
frozen throughout this stage.

Given the current compact representation $z_t$ and the action proposal $A_t$
available from the current policy query, each action is independently projected to the transition latent
dimension and augmented with a learnable step embedding that encodes its position within the action
chunk. The resulting action tokens serve as queries in stacked cross-attention blocks. To construct the keys and values, we apply linear projections independently to each token in $L_t$. The cross-attention blocks attend to these projected tokens and produce action-conditioned features over the prediction horizon.
We mean-pool both the action tokens and the resulting action-conditioned features,
and concatenate them with the current compact representation $z_t$. A residual transition MLP takes the concatenated features as input and predicts a latent update, which is added to the current representation \(z_t\) to obtain the future compact representation $\hat{z}_{t+H|t}$. The predicted representation is passed
through the frozen Stage~1 detector head to obtain the anticipated future risk
$r_{t+H|t}^{\mathrm{pred}}$.

During training, the actual future observation $O_{t+H}$ is processed by the same frozen detector to
obtain the target representation $z_{t+H}$. The transition model is trained to predict this future latent
representation using an MSE objective together with future-risk supervision:
\begin{equation}
\mathcal{L}_{\mathrm{pre}}
=
\frac{\lambda_{\mathrm{feat}}}{d}
\left\|
\hat{z}_{t+H|t}
-
\operatorname{sg}(z_{t+H})
\right\|_2^2
+
\lambda_{\mathrm{risk}}
\mathcal{L}_{\mathrm{det}}^{\mathrm{future}}.
\end{equation}

Here, $d$ denotes the compact state dimension and $\operatorname{sg}$ stops gradients through the target representation. The future-risk loss $\mathcal{L}_{\mathrm{det}}^{\mathrm{future}}$ follows Equation~\ref{eq:detector_loss} using
$r_{t+H|t}^{\mathrm{pred}}$. The two terms supervise future latent prediction and failure-risk estimation, respectively. We use $\lambda_{\mathrm{feat}}=\lambda_{\mathrm{risk}}=1$ by default.

\subsection{Threshold Calibration}
\label{sec:threshold}

Following prior work on time-dependent failure-score calibration~\citep{xu2025faildetect}, we calibrate monitoring thresholds using a held-out set of successful rollouts.
Scores estimated may vary across execution phases even in successful rollouts, motivating thresholds that vary with time. To align score trajectories for calibration, we consider three temporal parameterizations.
\emph{Absolute-time} calibration indexes thresholds by the raw
timestep $t$. \emph{Normalized-time} calibration represents each timestep by
its relative progress within the trajectory. \emph{Staged-time} calibration
partitions normalized execution progress into a fixed number of stages and
uses a shared threshold within each stage. 

For each temporal parameterization, we construct thresholds using either
\emph{pointwise} quantiles or a one-sided \emph{functional} conformal band,
resulting in six calibration variants. The detector and predictor are calibrated
independently under the same protocol. For further details, refer to Appendix~\ref{app:time_dependent_threshold}. Let $\delta_t^{\mathrm{det}}$ and
$\delta_t^{\mathrm{pred}}$ denote their respective thresholds. At runtime, the
monitor raises an alarm whenever either branch crosses its calibrated threshold:
\begin{equation}
b_t=
\mathbbm{1}\!\left[
r_t^{\mathrm{det}}\geq\delta_t^{\mathrm{det}}
\right]
\lor
\mathbbm{1}\!\left[
r_{t+H|t}^{\mathrm{pred}}\geq\delta_t^{\mathrm{pred}}
\right].
\label{eq:runtime_or}
\end{equation}

\section{Experiments}

We evaluate \methodname{} in simulation and on real robots to answer three questions. (1) Does ReAP achieve a favorable balance between failure-detection accuracy, warning timeliness, and false alarms across diverse policies and tasks?~(Section~\ref{sec:offline} and Appendix~\ref{app:generalization})
(2) Can ReAP maintain effective online monitoring on real robots while introducing low inference overhead?~(Section~\ref{sec:online})
(3) What enables its monitoring performance, and how do failure-relevant representation learning and action-conditioned prediction contribute to the final monitor?~(Section~\ref{sec:analysis} and Appendix~\ref{app:additional_component})






\subsection{Experiment Setup}

\label{sec:experiment_setup}

\paragraph{Baselines.}
We compare \methodname{} with representative runtime failure monitors from
three categories. 
\textit{Action-uncertainty methods} estimate failure risk from the policy's action-generation process. STAC~\citep{agia2025sentinel}
measures inconsistency across sampled actions,
Diff-DAgger~\citep{lee2025diffdagger} uses diffusion-model objectives as uncertainty signals, and ACCEL~\citep{rao2026accel} derives uncertainty from flow-matching trajectories.
\textit{OOD-based methods} identify failures from deviations relative to successful execution. We include RND~\citep{burda2019exploration} and
LogpZO~\citep{xu2025faildetect}, and additionally evaluate FIPER~\citep{romer2025fiper}, which augments an observation-based OOD signal with action uncertainty.
\textit{Classifier-based methods} directly learn failure risk from successful and failed rollouts.
We include SAFE-MLP and SAFE-LSTM~\citep{gu2025safe}, which operate on internal VLA representations extracted from the deployed policy.

\paragraph{Evaluation Protocols.}
Following ~\citet{romer2025fiper}, we evaluate runtime monitoring with three complementary metrics, balanced time-weighted accuracy (bTWA), balanced accuracy (bACC), and normalized detection time (T-det). bACC measures failure-detection accuracy while accounting for false alarms on successful rollouts, with
$\mathrm{bACC}=(\mathrm{TPR}+\mathrm{TNR})/2$. T-det measures warning timeliness as the mean normalized first-alarm time over all failed rollouts, with missed failures assigned a value of one. We use bTWA as the primary metric because it jointly rewards failure detection and early warning while penalizing missed failures and false alarms:
\begin{equation}
\mathrm{bTWA}
=
\frac{1}{2}
\left(
\frac{1}{N_{\mathrm{fail}}}
\sum_{i\in\mathrm{TP}}
\left(1-\frac{t_i}{T_i}\right)
+
\frac{N_{\mathrm{TN}}}{N_{\mathrm{succ}}}
\right),
\label{eq:btwa}
\end{equation}

Here, $\mathrm{TP}$ denotes the set of failed rollouts that are detected, and $t_i$ and $T_i$ denote the first-alarm time and rollout length for a correctly detected failure. $N_{\mathrm{fail}}$, $N_{\mathrm{succ}}$ and $N_{\mathrm{TN}}$ denote the numbers of failed rollouts, successful rollouts, and successful rollouts without false alarms, respectively. 

\subsection{Offline Failure Monitoring}
\label{sec:offline}




\begin{table*}[t]
\centering
\caption{
Offline failure monitoring across four policy--benchmark settings. Results are averaged over five calibration levels, $\alpha\in\{0.05,0.10,0.15,0.20,0.25\}$, with means and standard deviations reported over three seeds.
Best and second-best results for each metric are shown in \textbf{bold} and
\underline{underline}, respectively.
\methodname{} (OR) is the complete monitor, while the detector and predictor rows report standalone performance for reference.
A dash~(--) indicates that the method is not applicable to the corresponding policy in our evaluation setting, with further details provided in Appendix~\ref{app:implementation}. RoboTwin-8 denotes policy-specific eight-task subsets selected separately for
$\pi_{0.5}$ and Fast-WAM. 
}
\label{tab:main_offline_results}

\small
\setlength{\tabcolsep}{4.8pt}
\renewcommand{\arraystretch}{1.08}


\textbf{(a) LIBERO-10}

\vspace{3pt}

\begin{adjustbox}{max width=0.98\textwidth}
\begin{tabular}{
@{}
>{\raggedright\arraybackslash}m{1.65cm}
l
ccc
ccc
@{}
}
\toprule
& &
\multicolumn{3}{c}{\textbf{OpenVLA (Success Rate: 53.6\%)}}
&
\multicolumn{3}{c}{\textbf{$\bm{\pi_0}$ (Success Rate: 80.4\%)}}
\\
\cmidrule(lr){3-5}
\cmidrule(lr){6-8}

Category
& Method
& bTWA $\uparrow$
& bACC $\uparrow$
& T-det $\downarrow$
& bTWA $\uparrow$
& bACC $\uparrow$
& T-det $\downarrow$
\\
\midrule


\multirow[c]{3}{=}{Action Uncertainty}
& STAC
& --
& --
& --
& \res{0.490}{0.000}
& \res{0.578}{0.000}
& \secondres{0.235}{0.000}
\\

& Diff-DAgger
& --
& --
& --
& \res{0.454}{0.000}
& \res{0.531}{0.000}
& \bestres{0.163}{0.000}
\\

& ACCEL
& --
& --
& --
& \res{0.628}{0.000}
& \res{0.777}{0.000}
& \res{0.524}{0.000}
\\

\midrule


\multirow[c]{3}{=}{OOD-based}
& RND
& \res{0.584}{0.013}
& \res{0.715}{0.048}
& \secondres{0.596}{0.069}
& \res{0.751}{0.013}
& \bestres{0.874}{0.010}
& \res{0.410}{0.027}
\\

& LogpZO
& \res{0.475}{0.023}
& \res{0.531}{0.019}
& \res{0.809}{0.114}
& \res{0.714}{0.006}
& \res{0.862}{0.007}
& \res{0.454}{0.008}
\\

& FIPER
& \res{0.559}{0.004}
& \bestres{0.808}{0.035}
& \res{0.812}{0.015}
& \res{0.683}{0.005}
& \res{0.866}{0.005}
& \res{0.518}{0.022}
\\

\midrule


\multirow[c]{2}{=}{Classifier}
& SAFE-MLP
& \secondres{0.622}{0.004}
& \res{0.687}{0.001}
& \res{0.641}{0.019}
& \secondres{0.758}{0.005}
& \res{0.869}{0.005}
& \res{0.378}{0.005}
\\

& SAFE-LSTM
& \res{0.502}{0.004}
& \res{0.528}{0.037}
& \res{0.889}{0.077}
& \res{0.724}{0.014}
& \secondres{0.871}{0.010}
& \res{0.394}{0.024}
\\

\midrule


\rowcolor{blue!4}
&
\textbf{\methodname{}~(Detector)}
& \res{0.671}{0.021}
& \res{0.744}{0.020}
& \res{0.450}{0.041}
& \res{0.750}{0.005}
& \res{0.838}{0.013}
& \res{0.466}{0.014}
\\

\rowcolor{blue!4}
&
\textbf{\methodname{}~(Predictor)}
& \res{0.683}{0.007}
& \res{0.764}{0.017}
& \res{0.406}{0.038}
& \res{0.755}{0.009}
& \res{0.831}{0.010}
& \res{0.461}{0.017}
\\

\rowcolor{blue!4}
\multirow{-3}{=}{\textbf{Ours}}
&
\textbf{\methodname{}~(OR)}
& \bestres{0.701}{0.010}
& \secondres{0.770}{0.008}
& \bestres{0.326}{0.029}
& \bestres{0.780}{0.008}
& \res{0.865}{0.015}
& \res{0.395}{0.025}
\\

\bottomrule
\end{tabular}
\end{adjustbox}

\vspace{10pt}


\textbf{(b) RoboTwin-8 }

\vspace{3pt}

\begin{adjustbox}{max width=0.98\textwidth}
\begin{tabular}{
@{}
>{\raggedright\arraybackslash}m{1.65cm}
l
ccc
ccc
@{}
}
\toprule
& &
\multicolumn{3}{c}{\textbf{$\bm{\pi_{0.5}}$ (Success Rate: 48.4\%)}}
&
\multicolumn{3}{c}{\textbf{Fast-WAM (Success Rate: 66.6\%)}}
\\
\cmidrule(lr){3-5}
\cmidrule(lr){6-8}

Category
& Method
& bTWA $\uparrow$
& bACC $\uparrow$
& T-det $\downarrow$
& bTWA $\uparrow$
& bACC $\uparrow$
& T-det $\downarrow$
\\
\midrule


\multirow[c]{3}{=}{Action Uncertainty}
& STAC
& \res{0.507}{0.000}
& \res{0.554}{0.000}
& \res{0.349}{0.000}
& \res{0.487}{0.000}
& \res{0.506}{0.000}
& \bestres{0.269}{0.000}
\\

& Diff-DAgger
& \res{0.513}{0.000}
& \res{0.560}{0.000}
& \secondres{0.312}{0.000}
& \res{0.438}{0.044}
& \res{0.460}{0.034}
& \res{0.921}{0.012}
\\

& ACCEL
& \res{0.512}{0.000}
& \res{0.644}{0.000}
& \res{0.675}{0.000}
& \res{0.491}{0.000}
& \res{0.556}{0.000}
& \res{0.883}{0.000}
\\

\midrule


\multirow[c]{3}{=}{OOD-based}
& RND
& \secondres{0.667}{0.000}
& \res{0.733}{0.007}
& \res{0.636}{0.022}
& \res{0.537}{0.038}
& \res{0.585}{0.053}
& \res{0.605}{0.108}
\\

& LogpZO
& \res{0.645}{0.009}
& \res{0.775}{0.009}
& \res{0.556}{0.004}
& \res{0.545}{0.010}
& \res{0.561}{0.010}
& \res{0.771}{0.036}
\\

& FIPER
& \res{0.645}{0.024}
& \bestres{0.815}{0.035}
& \res{0.493}{0.065}
& \res{0.541}{0.005}
& \res{0.597}{0.012}
& \secondres{0.590}{0.039}
\\

\midrule


\multirow[c]{2}{=}{Classifier}
& SAFE-MLP
& \res{0.589}{0.004}
& \res{0.668}{0.006}
& \res{0.605}{0.004}
& \res{0.551}{0.014}
& \res{0.559}{0.017}
& \res{0.719}{0.035}
\\

& SAFE-LSTM
& \res{0.562}{0.010}
& \res{0.598}{0.012}
& \res{0.701}{0.003}
& \secondres{0.578}{0.009}
& \secondres{0.636}{0.012}
& \res{0.800}{0.007}
\\

\midrule


\rowcolor{blue!4}
&
\textbf{\methodname{}~(Detector)}
& \res{0.689}{0.021}
& \res{0.795}{0.018}
& \res{0.356}{0.027}
& \res{0.626}{0.013}
& \res{0.752}{0.015}
& \res{0.675}{0.038}
\\

\rowcolor{blue!4}
&
\textbf{\methodname{}~(Predictor)}
& \res{0.695}{0.017}
& \res{0.792}{0.019}
& \res{0.342}{0.027}
& \res{0.640}{0.015}
& \res{0.761}{0.008}
& \res{0.642}{0.053}
\\

\rowcolor{blue!4}
\multirow{-3}{=}{\textbf{Ours}}
&
\textbf{\methodname{}~(OR)}
& \bestres{0.701}{0.013}
& \secondres{0.785}{0.011}
& \bestres{0.306}{0.021}
& \bestres{0.651}{0.014}
& \bestres{0.779}{0.008}
& \res{0.592}{0.051}
\\

\bottomrule
\end{tabular}
\end{adjustbox}

\end{table*}

We evaluate offline failure monitoring across four policy–benchmark settings.
On LIBERO-10~\citep{liu2023libero}, we evaluate
OpenVLA~\citep{kim2025openvla} and $\pi_0$~\citep{black2024pi0}.
On RoboTwin 2.0~\citep{chen2025robotwin}, we evaluate
$\pi_{0.5}$~\citep{physicalintelligence2025pi05} and
Fast-WAM~\citep{yuan2026fastwam} on policy-specific subsets of eight tasks, referred to as RoboTwin-8.
For each setting, a number of successful rollouts are held out for threshold calibration, and all reported results are computed on a separate evaluation split.
Following \citet{park2026hideandseek}, we average performance over $\alpha \in \{0.05, 0.10, 0.15, 0.20, 0.25\}$
under the same calibration protocol unless otherwise specified.

\paragraph{Results Analysis.}

Table~\ref{tab:main_offline_results} shows that \methodname{} achieves the highest bTWA across all four policy--benchmark settings, with relative improvements of up to $12.7\%$ over the strongest evaluated baseline in each setting. Averaged across the four settings, \methodname{} achieves a bTWA of $0.708$, a bACC of $0.800$, and a T-det of $0.405$. These results demonstrate a favorable balance between failure-detection accuracy and warning timeliness across different policy architectures and manipulation benchmarks.

The baseline results reveal different limitations in achieving this balance between accuracy and timeliness. Action-uncertainty methods can produce early warnings but often with lower balanced accuracy. For example, Diff-DAgger reaches a T-det of $0.163$ on $\pi_0$/LIBERO-10, while its bACC is only $0.531$. In contrast, OOD-based methods often achieve stronger balanced accuracy but raise alarms later. FIPER obtains a bACC of $0.808$ on OpenVLA/LIBERO-10, but with a T-det of $0.812$. SAFE is competitive in several settings, yet its performance varies across policies. 

To assess how the two branches contribute to these gains, we also report the results of the detector and predictor.
Both branches of \methodname{} already provide strong standalone monitoring signals. The detector effectively recognizes failure evidence from observed execution states, while the predictor achieves comparable or stronger
performance through action-conditioned consequence prediction. Applying the OR rule further improves bTWA over both components in all four settings and consistently reduces normalized detection time. These consistent gains support the complementary roles of current-state failure recognition and action-conditioned future-risk prediction.




\subsection{Online Real-Robot Monitoring}
\label{sec:online}

\begin{table*}[t]
\centering

\caption{
Real-world failure monitoring and runtime efficiency.
(a) Failure monitoring results across three real-world manipulation tasks.
SR denotes the success rate of the deployed base policy.
(b) Mean runtime per monitoring evaluation for representative failure monitors on a single NVIDIA A800 80GB GPU. For methods that rely on policy-derived features or action-uncertainty signals, the reported runtime includes the policy inference required to obtain these signals.
}
\label{tab:real_robot_runtime}

\vspace{4pt}

\begin{minipage}[t]{0.67\textwidth}
\vspace{0pt}
\centering
\label{tab:real_robot}

\textbf{(a) Real-world Failure Monitoring}

\vspace{3pt}

\small
\setlength{\tabcolsep}{2.2pt}
\renewcommand{\arraystretch}{1.08}

\begin{adjustbox}{max width=\linewidth}
\begin{tabular}{
@{}
l
cc
cc
cc
@{}
}
\toprule

&
\multicolumn{2}{c}{\textbf{Fold Cloth (SR 60.0\%)}}
&
\multicolumn{2}{c}{\textbf{Stack Plates (SR 75.0\%)}}
&
\multicolumn{2}{c}{\textbf{Pick Cubes (SR 50.0\%)}}
\\

\cmidrule(lr){2-3}
\cmidrule(lr){4-5}
\cmidrule(lr){6-7}

Method
& bTWA $\uparrow$
& bACC $\uparrow$
& bTWA $\uparrow$
& bACC $\uparrow$
& bTWA $\uparrow$
& bACC $\uparrow$
\\
\midrule

STAC
& 0.542
& 0.613
& 0.599
& 0.721
& 0.518
& 0.546
\\

ACCEL
& 0.584
& 0.639
& 0.407
& 0.517
& 0.452
& 0.525
\\

\midrule

RND
& 0.566
& 0.664
& 0.597
& 0.721
& 0.608
& 0.695
\\

LogpZO
& 0.541
& 0.650
& 0.754
& 0.893
& 0.547
& 0.632
\\

FIPER
& 0.508
& 0.513
& 0.671
& 0.759
& 0.668
& \textbf{0.800}
\\

\midrule

SAFE-MLP
& \textbf{0.624}
& 0.688
& 0.801
& 0.892
& 0.572
& 0.681
\\

\midrule

\rowcolor{blue!4}
\textbf{\methodname{} (OR)}
& 0.603
& \textbf{0.713}
& \textbf{0.818}
& \textbf{0.900}
& \textbf{0.718}
& \textbf{0.800}
\\

\bottomrule
\end{tabular}
\end{adjustbox}

\end{minipage}
\hfill
\begin{minipage}[t]{0.30\textwidth}
\vspace{0pt}
\centering

\textbf{(b) Runtime Comparison}

\vspace{3pt}

\scriptsize
\setlength{\tabcolsep}{4.0pt}
\renewcommand{\arraystretch}{0.95}

\begin{adjustbox}{max width=\linewidth}
\begin{tabular}{@{}lc@{}}
\toprule
Method
& Runtime (ms) $\downarrow$
\\
\midrule

Base policy ($\pi_{0.5}$) & 67.07 \\
\midrule

STAC
& 223.92
\\

Diff-DAgger
& 775.51
\\

ACCEL
& 87.98
\\

RND
& 82.80
\\

LogpZO
& 76.68
\\

FIPER
& 224.69
\\

SAFE-MLP
& 67.00
\\

\midrule

\rowcolor{blue!4}
\textbf{\methodname{}~(Detector)}
& \textbf{9.72}
\\

\rowcolor{blue!4}
\textbf{\methodname{}~(Predictor)}
& 81.16
\\

\bottomrule
\end{tabular}
\end{adjustbox}

\end{minipage}
\end{table*}

We further evaluate \methodname{} on three real-world manipulation tasks, \textit{Fold Cloth}, \textit{Stack Plates}, and \textit{Pick Cubes}, using separately fine-tuned $\pi_{0.5}$ policies. For each task, we collect successful and failed policy rollouts to train and calibrate the monitor, and deploy \methodname{} during physical robot execution. During physical robot execution, the detector continuously monitors incoming observations, while the predictor evaluates each proposed action chunk before execution. Monitoring alarms are recorded without altering robot execution. To enable controlled comparison on identical physical trajectories, all baselines are evaluated post hoc using the same recorded rollouts and policy outputs. Additional details of the real-robot setup and deployment are provided in Appendix~\ref{app:real_robot_setup}.

Table~\hyperref[tab:real_robot_runtime]{\ref*{tab:real_robot_runtime}a} shows that \methodname{} maintains strong monitoring performance across all three real-world tasks. It achieves the highest bTWA on Stack Plates and Pick Cubes, and obtains the best or tied-best bACC on all three tasks. Averaged across the three tasks, \methodname{} achieves a bTWA of $0.713$ and a bACC of $0.804$, supporting its effectiveness in real-world failure monitoring.

We also benchmark monitoring latency using $\pi_{0.5}$ as the reference policy. As shown in Table~\hyperref[tab:real_robot_runtime]{\ref*{tab:real_robot_runtime}b}, the observation-only detector runs in $9.72$ ms without requiring policy inference. Predictor-based monitoring takes $81.16$ ms per policy query, including the $67.07$ ms base-policy inference, corresponding to an additional $14.09$ ms in this benchmark. In comparison, methods based on repeated policy sampling require substantially more computation. These measurements show that \methodname{} can provide effective online monitoring while adding limited latency to the deployed control pipeline.

\subsection{Component Analysis}
\label{sec:analysis}

\begin{figure*}[!t]
    \centering

    \begin{subfigure}[!t]{0.48\textwidth}
        \centering
        \includegraphics[width=\linewidth]{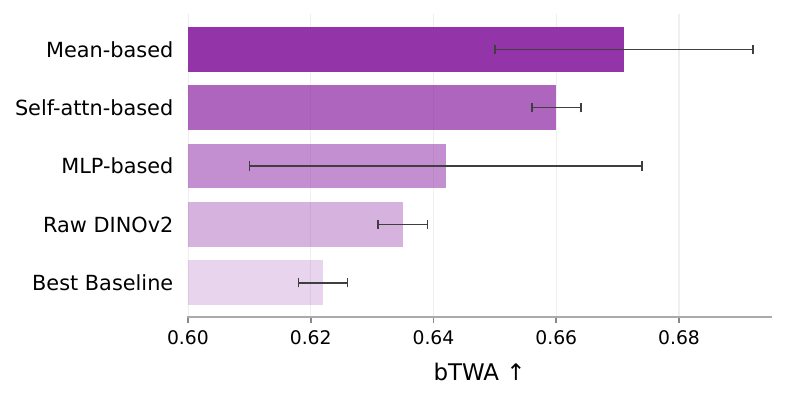}
        \caption{}
        \label{fig:detector_ablation}
    \end{subfigure}
    \hfill
    \begin{subfigure}[!t]{0.48\textwidth}
        \centering
        \includegraphics[width=\linewidth]{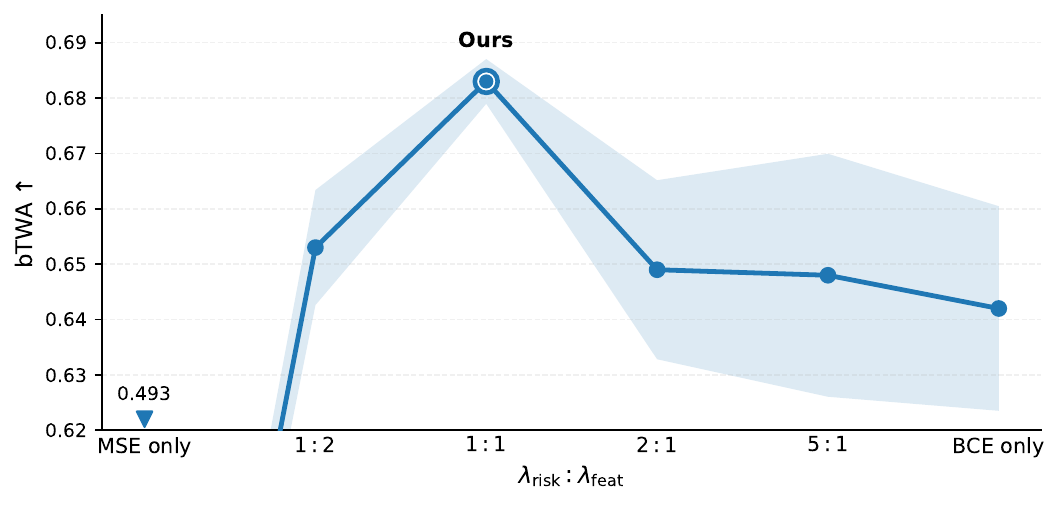}
        \caption{}
        \label{fig:predictor_loss}
    \end{subfigure}

    \caption{
    (a) Effect of different detector architectures.
    (b) Effect of future-risk and latent-transition loss weighting on predictor
    performance. The downward triangle indicates an off-scale value.
    }
    \label{fig:component_ablation}
\end{figure*}

This section examines the main factors underlying the effectiveness of \methodname{}. The detector analysis studies whether failure supervision produces a representation that is more suitable for failure monitoring than generic visual or policy features. The predictor analysis then evaluates how this representation benefits future-state prediction and how action conditioning improves monitoring performance. All component analyses are conducted on OpenVLA/LIBERO-10 under the same evaluation protocol as the offline benchmark. Additional analyses are provided in Appendix~\ref{app:additional_component}

\subsubsection{Detector Representation Learning}

\begin{figure}[t]
    \centering
    \includegraphics[width=\linewidth]{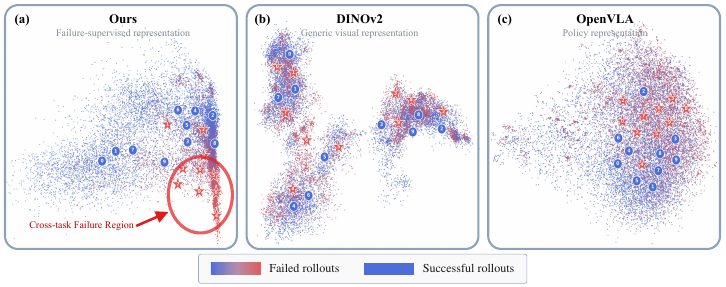}
    \caption{
    PCA visualization of representations on OpenVLA/LIBERO-10.
(a) Detector features.
(b) DINOv2 features.
(c) OpenVLA features.
Observations from successful rollouts are shown in blue, while those from failed rollouts are colored from blue to red as execution progresses.
Blue circles denote task-wise mean representations over observations from successful rollouts.
Red stars denote task-wise mean representations over the final $30\%$ of frames from failed rollouts.
    }
    \label{fig:detector_pca}
\end{figure}

\paragraph{Architecture Ablation.}

We ablate the feature extractor and latent-token aggregation in the observation-only detector. The \textit{Raw DINOv2} baseline directly mean-pools frozen DINOv2 patch features without the Perceiver-style feature extractor. The remaining variants use the Perceiver-style extractor to produce a fixed set of latent tokens and differ only in how these tokens are aggregated, using an MLP, self-attention, or mean pooling. As shown in Figure~\ref{fig:detector_ablation}, directly mean-pooling frozen DINOv2 features already achieves a bTWA of $0.635$, exceeding the strongest baseline in the main benchmark at $0.622$. Introducing the failure-supervised feature extractor further improves bTWA to $0.671$ with mean pooling, supporting the benefit of learning monitoring-relevant features on top of the frozen visual backbone. Mean pooling performs best among the aggregation variants, and we adopt it together with the learned feature extractor in the default detector.

\paragraph{Representation Structure.}

To examine how failure supervision separates successful and failed observations, we compare the learned detector representation with frozen DINOv2 patch features and OpenVLA internal representations in Figure~\ref{fig:detector_pca}. In the 2D PCA projection, DINOv2 features are organized primarily by task appearance, with successful and failed observations from the same task remaining close to each other. OpenVLA features similarly show limited separation by execution outcome. In contrast, the learned detector representation forms a more coherent failure region shared across tasks, with late-stage observations from failed rollouts clustering more closely in the projected space. Together with the quantitative gains in Figure~\ref{fig:detector_ablation}, this suggests that the learned representation captures failure-related structure shared across the evaluated tasks.

\subsubsection{Action-Conditioned Predictor Learning}

\paragraph{Training Objective.}


















\begin{table}[t]
\centering
\caption{
(a) Effect of the detector representation.
(b) Effect of action conditioning.
}
\label{tab:predictor_ablation}

\vspace{2pt}

\begin{minipage}[t]{0.49\textwidth}
\vspace{0pt}
\centering

\textbf{(a) Prediction Representation}

\vspace{2pt}

\small
\renewcommand{\arraystretch}{1.0}

\resizebox{\linewidth}{!}{%
\begin{tabular}{@{}lcc@{}}
\toprule
Method
& bTWA $\uparrow$
& bACC $\uparrow$
\\
\midrule

Generic PCA
& \res{0.644}{0.012}
& \res{0.747}{0.028}
\\

Random detector, $\lambda_{\mathrm{feat}}=0$
& \res{0.641}{0.012}
& \res{0.742}{0.027}
\\

Random detector
& \res{0.650}{0.005}
& \res{0.752}{0.009}
\\

\rowcolor{blue!4}
\textbf{Learned detector~(Ours)}
& \bestres{0.683}{0.007}
& \bestres{0.764}{0.017}
\\

\bottomrule
\end{tabular}%
}

\end{minipage}%
\hspace{0.020\textwidth}%
\begin{minipage}[t]{0.48\textwidth}
\vspace{0pt}
\centering

\textbf{(b) Action Conditioning}

\vspace{2pt}

\small
\renewcommand{\arraystretch}{1.0}

\resizebox{\linewidth}{!}{%
\begin{tabular}{@{}lcc@{}}
\toprule
Variant
& bTWA $\uparrow$
& bACC $\uparrow$
\\
\midrule

Zero action, MSE+BCE
& \res{0.520}{0.007}
& \res{0.603}{0.016}
\\

Zero action, BCE only
& \res{0.541}{0.008}
& \res{0.632}{0.015}
\\

Action shuffle
& \res{0.618}{0.005}
& \res{0.665}{0.006}
\\

\rowcolor{blue!4}
\textbf{Action-conditioned~(Ours)}
& \bestres{0.683}{0.007}
& \bestres{0.764}{0.017}
\\

\bottomrule
\end{tabular}%
}

\end{minipage}

\end{table}

The predictor is trained with two complementary objectives. The latent-transition loss encourages the predicted representation to match the detector representation of the actual future observation, while the future-risk loss directly supervises the predicted failure risk. We vary the relative weighting $\lambda_{\mathrm{risk}}:\lambda_{\mathrm{feat}}$ while keeping the predictor architecture fixed.
As shown in Figure~\ref{fig:predictor_loss}, using latent-transition supervision alone yields a bTWA of $0.493$, whereas future-risk supervision alone reaches $0.642$. Combining the two objectives further improves bTWA to $0.683$, with equal weighting performing best among the tested configurations. These results indicate that direct future-risk supervision is important for effective monitoring, and that jointly predicting future detector representations further improves performance over risk-only training.

\paragraph{Prediction Representation.}

Beyond the training objective, we assess how the representation space affects future-risk prediction by comparing three variants with the Stage~2 predictor architecture fixed. \textit{Generic PCA} compresses frozen DINOv2 features using PCA, \textit{Random detector} uses the same detector architecture without failure supervision, and \textit{Learned detector} uses the representation obtained after Stage~1 training. For the first two variants, the corresponding detector head is learned jointly with the predictor.
As shown in Table~\hyperref[tab:predictor_ablation]{\ref*{tab:predictor_ablation}a}, predicting in the learned detector space achieves the strongest performance, reaching a bTWA of $0.683$ and a bACC of $0.764$, compared with $0.650$ and $0.752$ for the random detector and $0.644$ and $0.747$ for Generic PCA. Adding the latent-transition loss to the random-detector variant yields only a small increase in mean bTWA. Together with the stronger performance of the learned detector, this suggests that failure-supervised detector features provide more effective prediction targets for future-risk estimation than generic or randomly initialized representations.

\paragraph{Action Conditioning.}

\begin{wraptable}{r}{0.5\textwidth}
\vspace{-\baselineskip}
\setlength{\abovecaptionskip}{0pt}
\centering
\caption{Future-latent prediction errors.}
\label{tab:action_latent_error}
\footnotesize
\setlength{\tabcolsep}{3pt}
\begin{tabular}{lcc}
\toprule
Action & $L_2$ error $\downarrow$ & Per-dim.\ MSE $\downarrow$ \\
\midrule
Shuffled
& \res{10.349}{0.038}
& \res{0.445}{0.003} \\
\textbf{Original}
& \bestres{9.680}{0.011}
& \bestres{0.394}{0.001} \\
\bottomrule
\end{tabular}
\end{wraptable}

Finally, we examine the contribution of action conditioning through two complementary ablations. First, we retrain the predictor with the entire action chunk set to zero. As shown in Table~\hyperref[tab:predictor_ablation]{\ref*{tab:predictor_ablation}b}, retaining the MSE+BCE objective reduces bTWA from $0.683$ to $0.520$ and bACC from $0.764$ to $0.603$. Training the zero-action variant with BCE alone also leaves a substantial gap to the full action-conditioned predictor. Second, we keep the trained predictor fixed and replace the action chunk at evaluation time with one sampled from another rollout of the same task, which preserves task identity while disrupting the correspondence between the current observation and the proposed action.
This action shuffle reduces bTWA to $0.618$ and bACC to $0.665$. It also increases both $L_2$ error and per-dimension MSE against the unchanged target future latent in all three seeds, as shown in Table~\ref{tab:action_latent_error}. Together, these results support the role of action conditioning in predicting the observed future representation and improving failure monitoring.

\section{Conclusion}

We presented \methodname{}, a future-aware runtime failure monitoring framework that builds on visual failure recognition to predict future action risks. A visual detector learns a failure-relevant representation space from successful and failed rollouts, within which an action-conditioned predictor estimates future representations and their associated risk. The combined monitor achieves the highest bTWA across all four simulated policy--benchmark settings, supporting the complementary value of current-state recognition and future-risk prediction in balancing failure detection, warning timeliness, and false alarms. Real-robot experiments demonstrate effective online monitoring, and runtime measurements indicate limited inference overhead. Component analyses further support the roles of failure-supervised representation learning and action conditioning in improving future-latent prediction and failure-risk estimation.

\subsection*{AI use statement}

In this work, we used generative AI tools to assist with manuscript drafting
and language refinement, literature retrieval and discovery, discussions of
research ideas and experimental design, and analysis and interpretation of
experimental results. Generative AI tools were also used to support code editing
and figure prototyping during the development of the paper. We also used Qwen3.5-9B to obtain approximate failure-onset annotations for the supplementary timing analysis~(Appendix~\ref{app:failure_onset_analysis}).
All research ideas,
methodological decisions, experiments, quantitative results, and scientific
claims were reviewed and verified by the authors. Literature suggestions were
checked against the original sources, AI-assisted code was manually inspected,
and AI-assisted interpretations of experimental results were verified against
the underlying data. 
We did not use generative AI tools to generate synthetic datasets, formulate or
prove mathematical claims, assist with mathematical proofs, or clean and
reformat datasets. We take full responsibility for all AI-assisted content in
this work.

\subsection*{Reproducibility statement}

The method formulation and training objectives are described in
Sec.~\ref{sec:problem} and~\ref{sec:method}, and the experimental setup
and evaluation protocol are provided in Sec.~\ref{sec:experiment_setup}.
Appendix~\ref{app:time_dependent_threshold} provides the detailed construction
and theoretical discussion of our threshold calibration.
Additional implementation details are provided in
Appendix~\ref{app:implementation}, including benchmark construction,
baseline implementations,
model architecture and training configurations, and the real-robot deployment
setup. 



\bibliography{iclr2027_conference}
\bibliographystyle{iclr2027_conference}

\newpage

\appendix

\section{Comparison with Model-Based Predictive Monitors}
\label{app:concurrent_world_model}

\citet{liu2024modelbased} combine latent-dynamics prediction,
OOD detection, and learned failure classification within interactive imitation
learning, using human intervention states to supervise failure detection.
Their follow-up work, Sirius-Fleet~\citep{liu2024multitaskinteractiverobotfleet},
extends predictive monitoring to multi-task robot fleets, using visual world
models and anomaly predictors to guide human intervention.
Concurrent works, Foresight~\citep{zhang2026foresight} and
ContactGuard~\citep{zheng2026contactguard}, also assess the anticipated
consequences of proposed actions. Below, we clarify how \methodname{}
relates to these approaches.

\paragraph{Failure supervision and prediction space.}
Foresight freezes a video-pretrained V-JEPA~2 encoder~\citep{assran2025vjepa2} and learns an
action-conditioned predictor before training a downstream failure detector.
ContactGuard learns a JEPA-style world model from unlabelled robot trajectories,
then freezes it and fits a linear failure probe.
In \methodname{}, failure supervision first shapes a compact representation
on top of a frozen visual backbone. We then freeze the detector and learn
action-conditioned transitions within this representation space, using both
latent-transition and future-risk supervision.
The same frozen detector head scores observed and predicted representations.
Thus, failure supervision determines the target space for prediction as well as the subsequent risk readout.

\paragraph{Monitoring scope and temporal context.}
Foresight aggregates predicted-latent tokens from the execution history with
a causal Transformer and uses trajectory-level outcome labels.
ContactGuard uses labelled grasp clips and a gripper-closure trigger to assess
predicted post-contact states before contact occurs.
\methodname{} also learns from episode-level outcomes, while combining
current-observation and predicted-state alarms throughout execution.
Its observation branch can assess incoming images independently of action
generation, and its prediction branch evaluates proposed action chunks.
This design requires neither a contact-event trigger nor a temporal
aggregation head.

\paragraph{Runtime considerations.}
Foresight reports $183.64$~ms per monitoring call with its Transformer head
on an NVIDIA H200 GPU~\citep[Table~14]{zhang2026foresight}.
ContactGuard reports $19.18$~ms for encoding, a 30-step latent rollout, and
classification on an NVIDIA RTX~5090~\citep[Table~4]{zheng2026contactguard}.
Both reported latencies exclude policy inference.
In our setup, the detector takes $9.72$~ms, while predictor-based monitoring
takes $81.16$~ms including policy inference on an NVIDIA A800 80GB GPU
(Table~\hyperref[tab:real_robot_runtime]{\ref*{tab:real_robot_runtime}b}).
Differences in hardware, prediction horizons, and timing boundaries prevent
a controlled speed comparison. Our measurements support the practicality
of a fast observation branch that can issue alarms without waiting for
policy inference.

\paragraph{Implications for generalization.}
Foresight's video pretraining may provide temporal and physical priors that help transfer to new settings. Our representation learns failure-related cues from labeled rollouts, so its transfer performance may depend on the range of tasks and failure modes covered during training. Evaluating both approaches under the same task and policy shifts would help establish how these differences affect their generalization.

\section{Comparison with Vision-Based Monitors}
\label{app:vision_based_monitors}

Several recent failure monitors assess robot execution directly from visual
observations or visually grounded representations. VLM-based approaches exploit
semantic reasoning to recognize and interpret manipulation failures.
AHA~\citep{duan2025aha} formulates failure detection as free-form visual
reasoning, while Guardian~\citep{pacaud2025guardian} learns detailed failure
detection and reasoning from automatically synthesized failure data.
FailSafe~\citep{lin2025failsafe} extends visual failure reasoning toward
recovery, and ViFailback~\citep{zeng2026vifailback} uses real-world failure
trajectories to train a VLM for failure diagnosis and visually grounded
correction guidance. SENTINEL~\citep{agia2025sentinel} incorporates a VLM-based
progress monitor alongside a separate action-consistency signal for runtime
failure detection. More broadly, recent progress and reward models learn from large-scale robot trajectory datasets to assess task progress and execution quality across successful and unsuccessful rollouts~\citep{liang2026robometer,acebrain2026,feng2026procvlm,lee2026roboreward}.

\begin{table}[t]
\centering
\caption{
Comparison with RoboMeter on LIBERO-10.
Latency denotes standalone inference time on a single NVIDIA A800 80GB GPU.
}
\label{tab:robometer_comparison}

\vspace{3pt}

\small
\setlength{\tabcolsep}{5pt}
\renewcommand{\arraystretch}{1.08}

\begin{tabular}{lccccc}
\toprule
&
\multicolumn{2}{c}{OpenVLA}
&
\multicolumn{2}{c}{$\pi_0$}
&
\\
\cmidrule(lr){2-3}
\cmidrule(lr){4-5}

Method
& bTWA $\uparrow$
& bACC $\uparrow$
& bTWA $\uparrow$
& bACC $\uparrow$
& Latency (ms) $\downarrow$
\\
\midrule

RoboMeter
& 0.669
& 0.739
& 0.714
& 0.831
& 109.40
\\

\rowcolor{blue!4}
\textbf{\methodname{}~(Detector)}
& \textbf{0.671}
& \textbf{0.744}
& \textbf{0.750}
& \textbf{0.838}
& \textbf{9.72}
\\

\bottomrule
\end{tabular}
\end{table}

Beyond large VLMs, observation-side monitoring has also been studied with
lighter learned models. RC-NF~\citep{zhou2026rcnf}, for example, combines
object-motion cues extracted from visual observations with robot state in a
robot-conditioned normalizing flow for real-time anomaly detection. Many existing vision-based monitors rely on multi-billion-parameter vision-language models, and some additionally require explicit semantic reasoning, leading to substantially higher computational cost than lightweight per-step monitors. RC-NF~\citep{zhou2026rcnf} is designed for real-time monitoring, but
its official implementation was not publicly available at the time of our
experiments. We therefore choose RoboMeter~\citep{liang2026robometer} for
quantitative comparison, as it has relatively low inference latency, is
pretrained on large-scale robot trajectory data, and its learned progress signal
can be directly adapted to failure detection.

We fine-tune RoboMeter on the full training split of our collected rollouts
and adapt its progress prediction for frame-level failure monitoring. At
timestep $t$, we sample eight frames from the observed trajectory prefix and
use RoboMeter to estimate the current task progress $p_t$, performing inference
every four environment frames. We then calibrate a time-varying lower band
$L_\alpha(t)$ from successful progress trajectories and raise an alarm when
$p_t \leq L_\alpha(t)$. Threshold selection and evaluation follow the same
protocol as our \methodname{}.

As shown in Table~\ref{tab:robometer_comparison}, our observation-only detector achieves higher
bTWA and bACC than RoboMeter in both LIBERO-10 policy settings, while requiring
substantially lower standalone inference latency. We also observe failed rollouts for which RoboMeter assigns relatively high
progress scores, suggesting that progress estimation may not always capture
fine-grained visual failure cues. This observation is
consistent with recent VLM-critic work showing that distinguishing fine-grained
visual differences between successful and failed executions remains challenging
for vision-language models~\citep{sudhakar2026robotcriticssweatsmall}.

\section{Implementation Details}
\label{app:implementation}

\subsection{Benchmark Details}

We construct disjoint training, validation, calibration, and evaluation splits
for each policy--benchmark setting, as summarized in Table~\ref{tab:data_splits}. The training split is used to train the
detector and predictor, while the validation split is additionally collected for hyperparameter
and model selection. The calibration split contains held-out successful rollouts used
only to fit the monitoring thresholds after model selection. All reported metrics
are computed on the separate evaluation split. The RoboTwin-8 task subsets are selected separately for $\pi_{0.5}$ and
Fast-WAM according to their policy success rates. The selected tasks and
their corresponding success rates are reported in
Table~\ref{tab:robotwin_tasks}.

For functional threshold calibration, we randomly divide the successful
calibration trajectories into two subsets using a 30/70 split, following the
functional calibration implementation of SAFE~\citep{gu2025safe}. The first
subset is used to estimate the mean trajectory and temporal modulation
function, while the second subset calibrates the width of the conformal band.
The detailed threshold construction is described in Appendix~\ref{app:time_dependent_threshold}.

\begin{table}[!ht]
\centering
\caption{
Dataset splits used for training, validation, threshold calibration, and
evaluation. Validation rollouts are used for hyperparameter and model selection,
while calibration contains held-out successful rollouts used only to fit the
monitoring thresholds.
}
\label{tab:data_splits}
\begin{tabular}{lcccc}
\toprule
Setting & Train & Validation & Calibration & Evaluation \\
\midrule
LIBERO-10   & 300 & 50 & 50 & 150 \\
RoboTwin-8  & 192 & 32 & 32 & 96 \\
Fold Cloth  & 200 & 50 & 50 & 40 \\
Stack Plates & 42 & 22 & 20 & 40 \\
Pick Cubes   & 40 & 23 & 20 & 40 \\
\bottomrule
\end{tabular}
\end{table}

\begin{table}[!ht]
\centering
\caption{
Selected RoboTwin-8 tasks and corresponding base-policy success rates.
The task subsets are selected separately for $\pi_{0.5}$ and Fast-WAM.
}
\label{tab:robotwin_tasks}

\small

\textbf{(a) $\bm{\pi_{0.5}}$}

\vspace{2pt}

\begin{tabularx}{\linewidth}{
    @{}>{\raggedright\arraybackslash}X r@{}
}
\toprule
Task instruction & Success Rate \\
\midrule
Lift the pot with both arms. & 25.0\% \\
Pick up one bottle with each arm. & 27.5\% \\
Pick up the hamburger and fries and place them on the tray. & 62.5\% \\
Place the container onto the plate. & 72.5\% \\
Pick up the playing card and move it away from the table. & 70.0\% \\
Press the top button of the alarm clock. & 72.5\% \\
Pick up the bread and place it into the skillet. & 20.0\% \\
Place the empty cup on the coaster. & 37.5\% \\
\bottomrule
\end{tabularx}

\vspace{7pt}

\textbf{(b) Fast-WAM}

\vspace{2pt}

\begin{tabularx}{\linewidth}{
    @{}>{\raggedright\arraybackslash}X r@{}
}
\toprule
Task instruction & Success Rate \\
\midrule
Grasp the red block with the left arm, hand it to the right arm, and place it on the blue pad. & 67.5\% \\
Pick up the mug, reorient it, and hang it on the rack. & 60.0\% \\
Move the stapler to a colored mat. & 52.5\% \\
Open the microwave with one arm. & 55.0\% \\
Pick up one bottle with each arm. & 67.5\% \\
Place the can into the basket and lift the basket. & 72.5\% \\
Stack the three bowls on top of each other. & 82.5\% \\
Press the switch. & 75.0\% \\
\bottomrule
\end{tabularx}

\end{table}

Our timestep supervision requires only the final success or failure outcome of
each rollout. Successful trajectories are assigned zero risk throughout
execution, while failed trajectories use the temporally increasing target
$y_t=t/(T-1)$ defined in Section~\ref{sec:problem}. The linear target is a weak trajectory-level risk proxy rather than a ground-truth failure-onset annotation. Compared with assigning the same binary failure label to every frame of a failed rollout, it avoids treating early, often nominal-looking frames as equally erroneous. We empirically found this linear supervision to perform better than binary targets for both ReAP and the supervised SAFE baselines in Appendix~\ref{app:additional_component}.

\subsection{Baseline Implementations}

For RND, LogpZO, SAFE-MLP, and SAFE-LSTM, we build on the public SAFE
implementation\footnote{\url{https://github.com/vla-safe/SAFE}} and follow its model architectures and optimization
hyperparameters unless otherwise specified. RND and LogpZO are trained only on successful rollouts and therefore do not use
failure-risk labels. For the supervised SAFE baselines, we replace the binary
failure target with the same temporally increasing target
$y_t=t/(T-1)$ used by our detector as discussed above.

For baselines that operate on policy representations, we follow SAFE for
feature extraction whenever applicable. For OpenVLA, we mean-pool the hidden states corresponding to the seven
autoregressive action tokens, yielding a 4096-D representation. For $\pi_0$ and $\pi_{0.5}$, we
average action-expert suffix hidden states across the action horizon and ten
flow-matching steps to obtain a 1024-D representation. For Fast-WAM, we
mean-pool the final video-MoT policy tokens into a 3072-D representation.

STAC, Diff-DAgger, ACCEL, and FIPER are implemented following their respective
public implementations. For methods that require repeated action sampling, we
generate 32 action chunks at each replanning step. We
retain the sampled action chunks and intermediate flow trajectories required
by the corresponding baselines. For Fast-WAM, repeated-sampling methods likewise use 32 action samples with an
action horizon of 32 and ten inference steps.

OpenVLA requires separate treatment because its standard rollout policy uses
deterministic autoregressive decoding. We apply this sampling procedure only to FIPER, whose action-uncertainty
component can be constructed directly from a set of stochastic action proposals. Diff-DAgger and ACCEL require policy-specific diffusion or
flow-matching quantities that are not defined for OpenVLA. STAC is not applicable to OpenVLA because it requires temporally overlapping
action chunks across consecutive policy queries, whereas OpenVLA predicts only
a single immediate action per query. We evaluate FIPER on OpenVLA using
stochastic action proposals following the sampling strategy of
RoboMonkey~\citep{kwok2025robomonkey}. Specifically, we first obtain multiple valid action
samples through stochastic autoregressive decoding, then fit a diagonal Gaussian
to the first six continuous action dimensions, and draw additional proposals
from the fitted distribution. The gripper action is determined by majority
vote, and continuous actions are clipped to the policy action-statistics
range. This sampling procedure is used only to construct the FIPER uncertainty
signal and does not alter the deterministic OpenVLA rollouts used for
evaluation.

FIPER uses the thresholding procedure from its original implementation.
For the remaining baselines, we select the time-varying calibration variant
using the validation split and fit its thresholds on the held-out successful
calibration rollouts using our six calibration variants.

\subsection{Architecture and Training Details}

We use the same monitor architecture and training configuration across all
policy--benchmark settings unless otherwise stated. The prediction horizon $H$ is measured in environment control steps.
Each training sample pairs the current observation $O_t$ with the
$H$ actions executed before the future observation $O_{t+H}$.
At inference, the predictor evaluates the first $H$ actions proposed by the current policy query.
Simulation experiments train
one monitor jointly across all tasks within each setting, while each real-world
task uses a separately trained monitor. The main architecture and training
configuration is summarized in Table~\ref{tab:reap_config}.
We use the LeRobot dataset interface\footnote{\url{https://github.com/huggingface/lerobot}}
to load our collected rollouts and sample training data.
Image augmentation, including
random cropping and small rotations for third-person views and color jitter for all camera views, is applied during training.
Each detector and predictor is trained on a single NVIDIA A800
80GB GPU.

\begin{table}[t]
\centering
\caption{
Architecture and training configuration of ReAP.
The same configuration is used across policy--benchmark settings unless
otherwise specified.
}
\label{tab:reap_config}
\small
\setlength{\tabcolsep}{5pt}
\renewcommand{\arraystretch}{1.08}

\begin{tabular}{
@{}p{0.24\linewidth}
@{\hspace{0.02\linewidth}}
>{\centering\arraybackslash}p{0.22\linewidth}
@{\hspace{0.04\linewidth}}
p{0.24\linewidth}
@{\hspace{0.02\linewidth}}
>{\centering\arraybackslash}p{0.22\linewidth}@{}
}
\toprule
\multicolumn{2}{c}{Architecture} &
\multicolumn{2}{c}{Training} \\
\cmidrule(lr){1-2}
\cmidrule(lr){3-4}
Setting & Value & Setting & Value \\
\midrule

Visual backbone
& DINOv2-Large
& Optimizer
& AdamW \\

Latent queries $M$
& 32
& Batch size
& 64 \\

Detector blocks $N_d$
& 2
& Training steps
& 10k \\

Detector latent width
& 1024
& Learning rate
& $1\times10^{-4}$ \\

Predictor blocks $N_p$
& 7
& Weight decay
& 0.01 \\

State / predictor width
& 256
& $(\lambda_{\mathrm{feat}},\lambda_{\mathrm{risk}})$
& $(1,1)$ \\

Prediction horizon $H$
& \begin{tabular}[t]{@{}c@{}}
    1 (OpenVLA) \\
    10 (others)
  \end{tabular}
& LR schedule
& \begin{tabular}[t]{@{}c@{}}
    Cosine \\
    500-step warmup
  \end{tabular} \\

\bottomrule
\end{tabular}
\end{table}

\subsection{Real-Robot Setup}
\label{app:real_robot_setup}

Our real-world experiments use a dual-arm robotic platform consisting of two
AgileX PIPER manipulators. Visual observations are captured by three Intel
RealSense D435i cameras, including one global third-person camera and one
wrist-mounted camera on each arm. For each manipulation task, we deploy a
separately fine-tuned $\pi_{0.5}$ policy and train a task-specific ReAP monitor
using the same model architecture and training configuration as in simulation.

Our deployment system is built on the Evo-RL framework~\citep{evorl2026}, which
we extend to support runtime failure monitoring. Observation acquisition, policy
inference, and robot control run in separate processes. A dedicated observation
server continuously provides the latest synchronized camera observations. The control process retrieves the latest observation and sends it
to a remote inference server, where the deployed policy generates the next action
chunk. The predictor runs alongside the policy on the inference server and
evaluates each proposed action chunk before execution, while the
observation-based detector runs independently on a local NVIDIA RTX 4070 GPU and
monitors incoming observations without waiting for policy inference. The action
chunk returned by the server is then executed by the robot. Whenever either
branch raises an alarm, we record the alarm together with the corresponding
rollout information for subsequent evaluation. \methodname{} is the only monitor integrated into the live deployment system. Baseline predictions are generated post hoc from the same recorded observations and policy outputs. Since monitoring alarms do not alter robot execution, all methods are evaluated on identical successful and failed trajectories, enabling paired comparison without collecting separate physical rollouts for each baseline.

\begin{figure*}[!ht]
    \centering

    \begin{subfigure}[t]{0.98\textwidth}
        \centering
        \includegraphics[width=\linewidth]{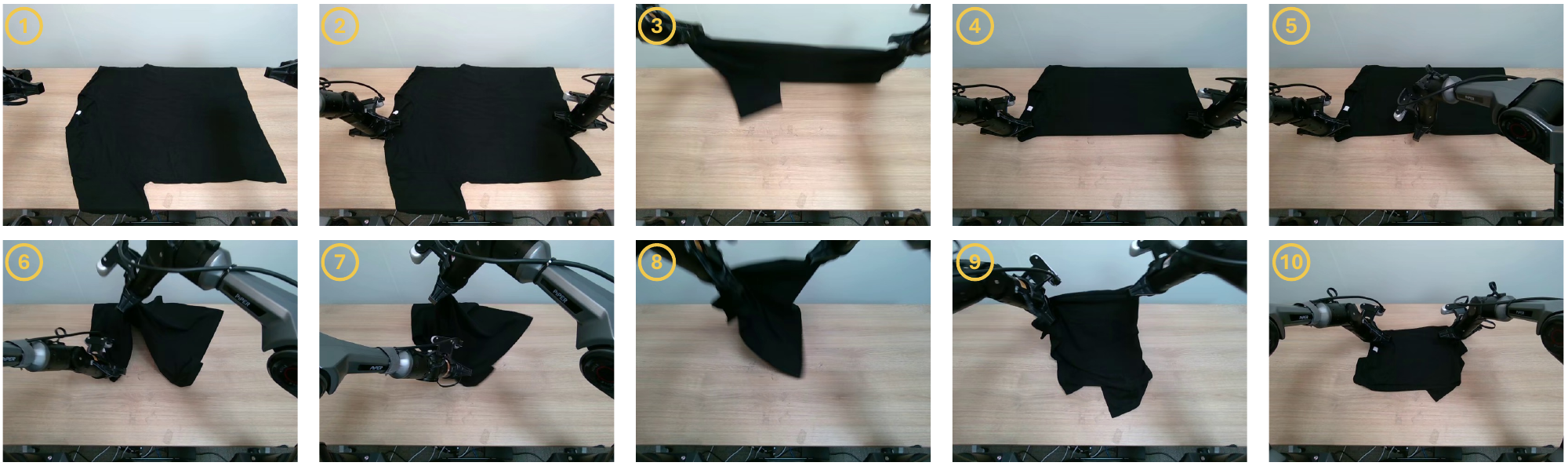}
        \caption{Fold Cloth.}
        \label{fig:real_fold_cloth}
    \end{subfigure}

    \vspace{4pt}

    \begin{subfigure}[t]{0.98\textwidth}
        \centering
        \includegraphics[width=\linewidth]{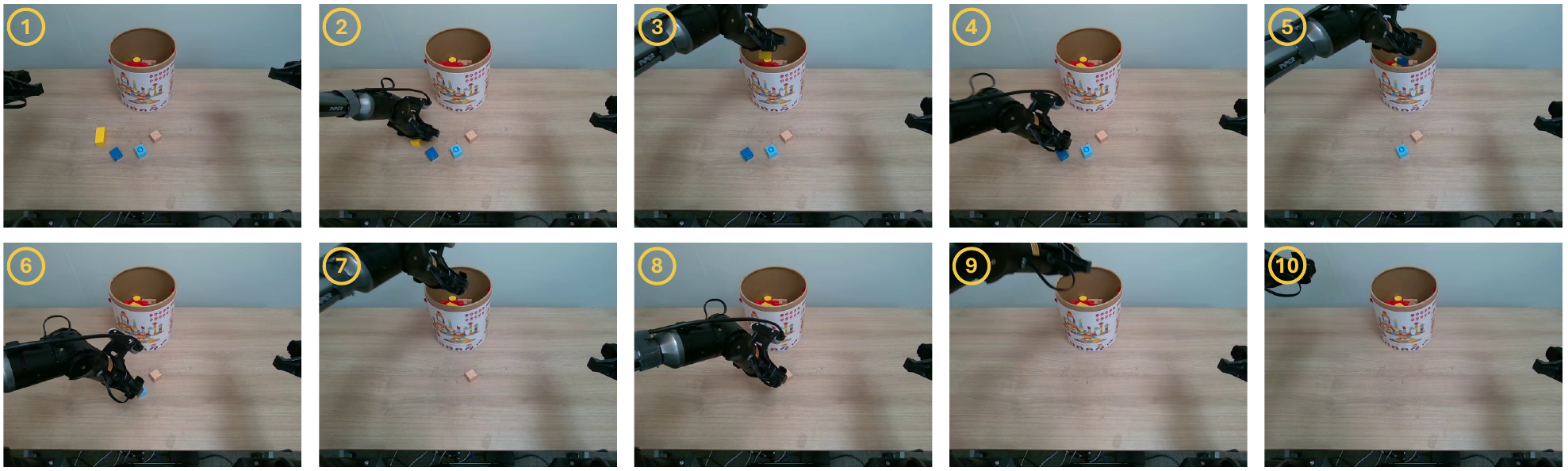}
        \caption{Pick Cubes.}
        \label{fig:real_pick_cubes}
    \end{subfigure}

    \vspace{4pt}

    \begin{subfigure}[t]{0.98\textwidth}
        \centering
        \includegraphics[width=\linewidth]{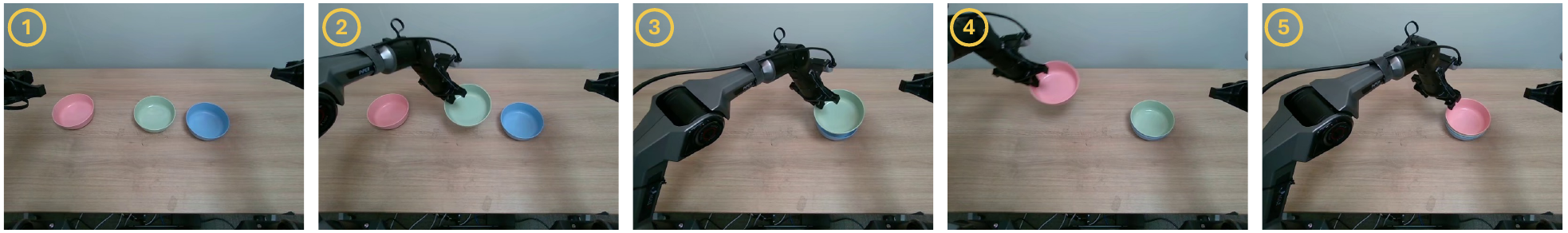}
        \caption{Stack Plates.}
        \label{fig:real_stack_plates}
    \end{subfigure}

    \caption{
    Representative real-world rollouts across three manipulation tasks.
    }
    \label{fig:real_robot_setup}
\end{figure*}

Figure~\ref{fig:real_robot_setup} illustrates the three
real-world manipulation tasks. In Fold Cloth, the two arms coordinate to lift,
reposition, and repeatedly fold a horizontally placed garment. In Stack Plates,
the robot manipulates three dishes and stacks them into a single pile. In Pick
Cubes, the robot picks four cubes from the tabletop and places them into a container.


\section{Additional Analyses}


\subsection{Generalization Analyses}

\label{app:generalization}

\paragraph{Seen-to-unseen generalization.}
A useful failure representation should preserve failure-relevant cues as task
appearance and interaction context change~\citep{gu2025safe}. We evaluate this
property on OpenVLA by randomly selecting three LIBERO-10 tasks (Task 1, 4, 7) as unseen tasks,
training all methods on the remaining seen tasks, calibrating corresponding thresholds on unseen tasks, and evaluating them on a disjoint set of unseen-task rollouts without further model training. As shown in
Table~\ref{tab:seen_unseen}, performance on unseen tasks is relatively close
across methods, reflecting the difficulty of transferring failure-monitoring
signals across task distributions. \methodname{} achieves the highest mean
performance, reaching 0.568 bTWA and 0.677 bACC. This suggests that the
failure-supervised representation captures failure cues that remain useful
beyond the tasks observed during training. However, the margin over the
strongest baselines is modest, indicating that cross-task transfer remains
challenging and that the learned failure representation is still affected by
task-specific visual and interaction patterns.

\begin{table*}[!ht]
\centering

\begin{minipage}[t]{0.45\textwidth}
\vspace{0pt}
\centering
\captionof{table}{Seen-to-unseen failure monitoring results.}
\label{tab:seen_unseen}
\vspace{2pt}

\small
\setlength{\tabcolsep}{4.0pt}
\renewcommand{\arraystretch}{1.05}

\begin{tabular}{lcc}
\toprule
Method & bTWA $\uparrow$ & bACC $\uparrow$ \\
\midrule
SAFE-MLP  & \res{0.537}{0.002} & \res{0.653}{0.007} \\
SAFE-LSTM & \res{0.556}{0.093} & \res{0.632}{0.090} \\
LogpZO    & \res{0.523}{0.070} & \res{0.592}{0.089} \\
RND       & \res{0.552}{0.030} & \res{0.661}{0.034} \\
FIPER     & \res{0.548}{0.000} & \res{0.600}{0.000} \\
\midrule
\rowcolor{blue!4}
\methodname{} (Ours)
          & \bestres{0.568}{0.064}
          & \bestres{0.677}{0.070} \\
\bottomrule
\end{tabular}
\end{minipage}
\hfill
\begin{minipage}[t]{0.53\textwidth}
\vspace{0pt}
\centering
\captionof{table}{Cross-policy failure monitoring on LIBERO-10.}
\label{tab:cross_policy_comparison}
\vspace{2pt}

\small
\setlength{\tabcolsep}{3.0pt}
\renewcommand{\arraystretch}{1.08}

\begin{tabular}{llcc}
\toprule
Method & Transfer & bTWA $\uparrow$ & bACC $\uparrow$ \\
\midrule
RoboMeter
& OpenVLA $\rightarrow \pi_0$
& 0.685 & 0.786 \\
RoboMeter
& $\pi_0 \rightarrow$ OpenVLA
& \textbf{0.630} & \textbf{0.685} \\
\midrule
\rowcolor{blue!4}
\methodname{}
& OpenVLA $\rightarrow \pi_0$
& \textbf{0.746} & \textbf{0.818} \\
\rowcolor{blue!4}
\methodname{}
& $\pi_0 \rightarrow$ OpenVLA
& 0.629 & 0.676 \\
\bottomrule
\end{tabular}
\end{minipage}

\vspace{-0.4\baselineskip}
\end{table*}

\paragraph{Cross-policy generalization.}
We next evaluate whether \methodname{} transfers across policies on LIBERO-10.
Since \methodname{} relies on external visual observations and proposed action
chunks without accessing policy-internal representations, the same monitor can
be applied to a different policy without architectural modification. We train
and calibrate the monitor using rollouts from one source policy and directly
evaluate it on another target policy without additional adaptation. For each transfer direction, we train the monitor on source-policy rollouts using the prediction horizon available at target-policy evaluation. For OpenVLA $\rightarrow \pi_0$, we construct ten-step action sequences from consecutive actions in the recorded OpenVLA rollouts and train the predictor with $H=10$. At evaluation, we use the first ten actions of each action chunk proposed by $\pi_0$. For $\pi_0 \rightarrow$ OpenVLA, we train and evaluate with $H=1$, matching OpenVLA's single-step action proposal. 

As shown in
Table~\ref{tab:cross_policy_comparison}, OpenVLA $\rightarrow \pi_0$ achieves
0.746 bTWA, compared with 0.780 under $\pi_0$ in-policy training in
Table~\ref{tab:main_offline_results}, while $\pi_0 \rightarrow$ OpenVLA reaches
0.629 compared with 0.701 under the corresponding OpenVLA in-policy setting.
Although performance decreases after transferring across policies, it remains
relatively close to the corresponding in-policy results in both settings.
This shows that \methodname{} can preserve useful failure-monitoring capability
across different deployed policies without target-policy adaptation. We further compare cross-policy transfer with RoboMeter, since it is likewise
independent of policy-internal signals. When transferring to $\pi_0$,
\methodname{} achieves 0.746 bTWA and 0.818 bACC, compared with 0.685 and
0.786 for RoboMeter. When transferring to OpenVLA, the two methods perform
comparably, with \methodname{} obtaining 0.629 bTWA and 0.676 bACC and
RoboMeter obtaining 0.630 and 0.685. Across the two transfer settings,
\methodname{} therefore maintains a relatively stable advantage or comparable
performance against RoboMeter.

\subsection{Additional Component Ablations}
\label{app:additional_component}

\paragraph{Visual backbone.}
To examine whether the detector depends on a particular pretrained visual
representation, we replace the visual backbone while keeping the detector
architecture and training protocol unchanged. Table~\ref{tab:visual_backbone_ablation}
shows relatively stable performance across CLIP~\citep{radford2021clip} and different DINOv2 variants,
suggesting that the detector is not highly sensitive to the choice of pretrained
visual encoder. DINOv2-Small achieves the highest mean performance, reaching
0.672 bTWA and 0.760 bACC, while the default DINOv2-Large obtains a nearly
identical bTWA of 0.671. Performance remains relatively stable across backbone choices, suggesting that ReAP can effectively use features from different pretrained visual backbones.

\begin{table}[!ht]
\centering

\begin{minipage}[t]{0.49\textwidth}
\vspace{0pt}
\centering

\captionof{table}{
Visual backbone ablation on OpenVLA/LIBERO-10.
}
\label{tab:visual_backbone_ablation}

\vspace{2pt}

\scriptsize
\setlength{\tabcolsep}{2.3pt}
\renewcommand{\arraystretch}{1.10}

\begin{tabular}{lccc}
\toprule
Visual Backbone
& bTWA $\uparrow$
& bACC $\uparrow$
& T-det $\downarrow$
\\
\midrule

CLIP ViT-B/32
& \res{0.654}{0.026}
& \res{0.740}{0.030}
& \res{0.462}{0.062}
\\

DINOv2-Small
& \bestres{0.672}{0.025}
& \bestres{0.760}{0.017}
& \bestres{0.432}{0.059}
\\

DINOv2-Base
& \res{0.657}{0.013}
& \res{0.747}{0.016}
& \res{0.450}{0.043}
\\

\rowcolor{blue!4}
\textbf{DINOv2-Large}
& \res{0.671}{0.021}
& \res{0.744}{0.020}
& \res{0.450}{0.041}
\\

\bottomrule
\end{tabular}

\end{minipage}
\hfill
\begin{minipage}[t]{0.49\textwidth}
\vspace{0pt}
\centering

\captionof{table}{
Ablation of the prediction horizon on $\pi_{0}$/LIBERO-10.
}
\label{tab:predictor_horizon}

\vspace{2pt}

\scriptsize
\setlength{\tabcolsep}{2.3pt}
\renewcommand{\arraystretch}{1.00}

\begin{tabular}{lccc}
\toprule
Prediction Horizon
& bTWA $\uparrow$
& bACC $\uparrow$
& T-det $\downarrow$
\\
\midrule

$H=1$
& \res{0.745}{0.002}
& \res{0.834}{0.014}
& \res{0.464}{0.027}
\\

$H=5$
& \res{0.744}{0.004}
& \bestres{0.837}{0.010}
& \res{0.464}{0.011}
\\

\rowcolor{blue!4}
\bm{$H=10$}
& \bestres{0.755}{0.009}
& \res{0.831}{0.010}
& \bestres{0.461}{0.017}
\\

$H=20$
& \res{0.736}{0.002}
& \res{0.826}{0.008}
& \res{0.475}{0.021}
\\

$H=30$
& \res{0.736}{0.004}
& \res{0.829}{0.005}
& \res{0.479}{0.018}
\\

\bottomrule
\end{tabular}

\end{minipage}

\end{table}

\paragraph{Prediction horizon.}
The prediction horizon $H$ determines how far ahead the predictor estimates the consequences of a proposed action chunk. We evaluate horizons from $H=1$ to $H=30$. As shown in Table~\ref{tab:predictor_horizon}, performance does not improve monotonically with the prediction horizon. $H=5$ achieves the highest bACC of 0.837, while $H=10$ achieves the highest bTWA of 0.755 and the numerically lowest T-det of 0.461. Increasing the horizon to $H=20$ or $H=30$ reduces bTWA to 0.736 and results in somewhat later alarms. We therefore use $H=10$ as the default horizon for chunked-action policies, based on its overall balance of detection accuracy and timeliness.



\begin{figure}
    \vspace{-8pt}
    \centering
    \includegraphics[
        width=\linewidth
    ]{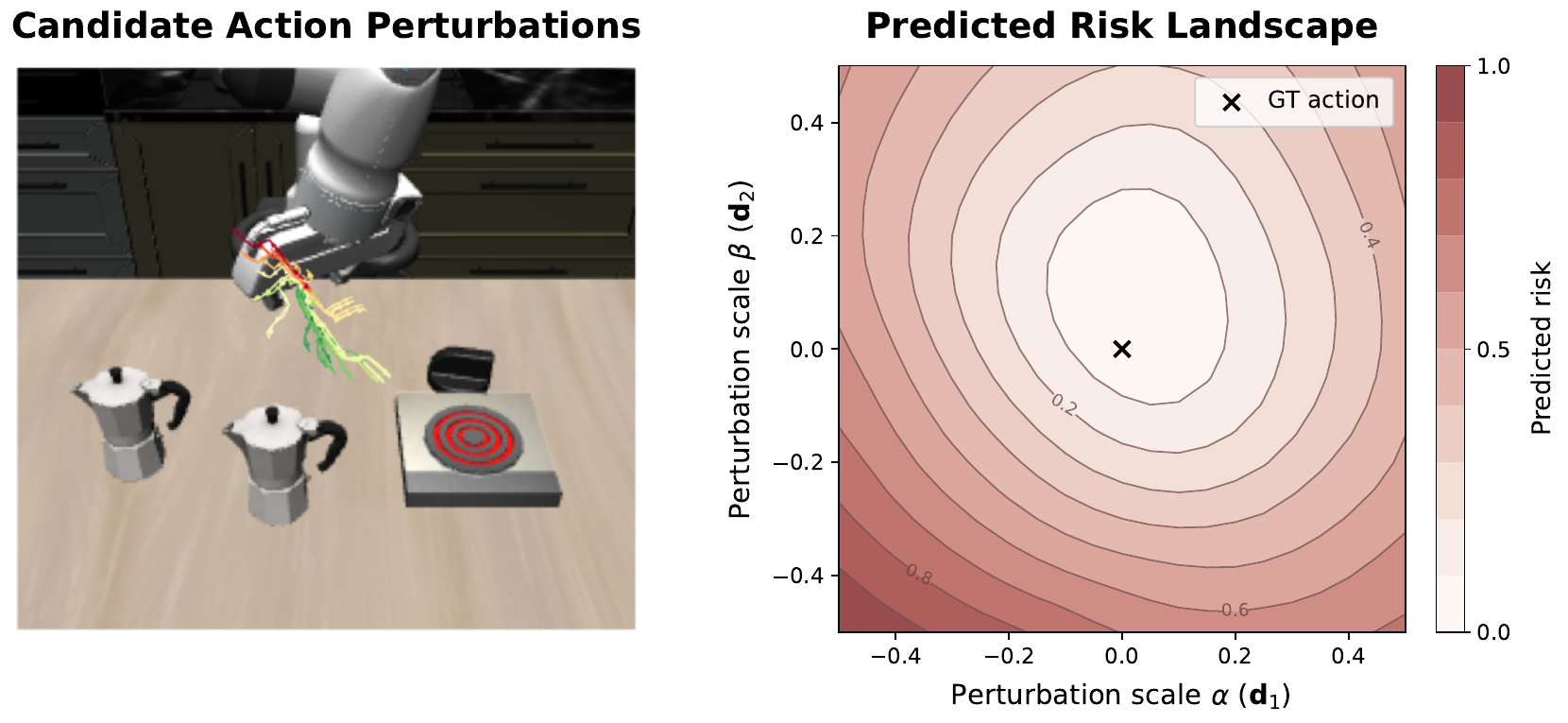}
    \vspace{-7pt}
    \caption{
    Action-sensitivity visualization on $\pi_0$/LIBERO-10 under controlled perturbations of a candidate action chunk.
    }
    \label{fig:action_landscape}
    \vspace{-8pt}
\end{figure}

\paragraph{Action Sensitivity.}

To qualitatively examine how the predictor responds to changes in the candidate action, we fix an observation from $\pi_0$/LIBERO-10 and visualize its predictions in Fig.~\ref{fig:action_landscape}. The left panel shows candidate action trajectories around the current gripper pose. We take a reference action chunk $\mathbf a_0$ from a held-out successful rollout and perturb it along two temporally smooth and orthogonal directions $\mathbf d_1$ and $\mathbf d_2$, forming $\mathbf a_{\alpha,\beta}=\mathbf a_0+\alpha\mathbf d_1+\beta\mathbf d_2$. The right panel shows the resulting predicted risk landscape, with lower risk near the reference action and generally higher risk for larger deviations. This visualization illustrates the predictor's sensitivity to action deviations under a fixed observation, complementing the quantitative action-conditioning results in Table~\hyperref[tab:predictor_ablation]{\ref*{tab:predictor_ablation}b} and Table~\ref{tab:action_latent_error}.


\paragraph{Effect of Training Labels}

\begin{table}[t]
\centering
\caption{Effect of training labels on OpenVLA/LIBERO-10.}
\label{tab:label_ablation}
\vspace{3pt}

\small
\setlength{\tabcolsep}{5pt}
\renewcommand{\arraystretch}{1.08}

\begin{tabular}{llccc}
\toprule
Model & Label
& bTWA $\uparrow$
& bACC $\uparrow$
& T-det $\downarrow$ \\
\midrule
SAFE-MLP & Binary
& \res{0.498}{0.005}
& \bestres{0.797}{0.013}
& \res{0.925}{0.009} \\
SAFE-MLP & Temporal
& \res{0.622}{0.004}
& \res{0.687}{0.001}
& \res{0.641}{0.019} \\
\midrule
\textbf{\methodname{}~(Detector)} & Binary
& \res{0.465}{0.020}
& \res{0.669}{0.044}
& \res{0.946}{0.034} \\
\rowcolor{blue!4}
\textbf{\methodname{}~(Detector)} & Temporal
& \bestres{0.671}{0.021}
& \res{0.744}{0.020}
& \bestres{0.450}{0.041} \\
\bottomrule
\end{tabular}
\end{table}

We compare binary and temporally increasing risk targets on
OpenVLA/LIBERO-10. The binary target assigns a risk label of one to
every observation in a failed rollout, while our default target
increases linearly from zero to one over that rollout. Successful
rollouts receive a target of zero in both cases. Table~\ref{tab:label_ablation} shows that the effect of the label
depends on the monitored model and the evaluation metric. Although binary labels yield higher bACC for SAFE-MLP, they lead to substantially later alarms and consequently lower bTWA. The \methodname{} detector trained with binary labels also raises alarms much later and achieves lower bTWA than its temporally supervised counterpart. These results support the temporally increasing target for the accuracy–timeliness trade-off measured by bTWA.

\subsection{Alarm Timing Relative to Annotated Failure Onset}
\label{app:failure_onset_analysis}

To provide a temporal reference for evaluating alarm timing, we use Qwen3.5-9B~\citep{qwen35} to obtain approximate failure-onset annotations for failed rollouts on LIBERO-10. We adapt the annotation prompt from Appendix F.1 of \citet{park2026hideandseek} and use a coarse-to-fine procedure with two successive model calls per rollout. In both calls, the model receives image grids containing the main-camera and wrist-camera views at each sampled time point, labeled with the frame index and timestamp. The first call locates the approximate onset using frames sampled throughout the rollout every 25 timesteps, including the first and last frames. Each 535-frame failed rollout yields 23 sampled time points, arranged into four image grids. The second call then refines this estimate using denser samples taken every three frames within a window extending 30 frames before and after the initial estimate. These model-generated annotations serve only as an approximate
temporal reference for this supplementary analysis and are not used for training or threshold calibration.

We compute the
normalized detection lag for each method and calibration level
as
\begin{equation}
\Delta t =
\frac{1}{|\mathcal{TP}|}
\sum_{i \in \mathcal{TP}}
\frac{t_i^{*} - t_{\mathrm{alarm},i}}{T_i},
\label{eq:onset_detection_lag}
\end{equation}
where $\mathcal{TP}$ denotes correctly detected failed rollouts,
$t_i^{*}$ is the annotated failure onset,
$t_{\mathrm{alarm},i}$ is the first alarm time, and $T_i$ is the
rollout length. Positive values indicate alarms before the
annotated onset, while negative values indicate delayed alarms.
Missed failures are excluded from $\Delta t$ and are reflected
in bACC. We report both metrics averaged over
$\alpha \in \{0.05, 0.10, 0.15, 0.20, 0.25\}$.

\begin{table}[t]
\centering
\caption{
Detection timing and failure coverage on LIBERO-10.
Detected failures reports the mean number of correctly detected failures
out of 75 for OpenVLA and 60 for $\pi_0$.
Positive $\Delta t$ indicates alarms before the annotated failure onset,
while negative values indicate delayed alarms.
$\Delta t$ is computed only over correctly detected failed rollouts.
}
\label{tab:failure_onset}
\small
\setlength{\tabcolsep}{4pt}
\renewcommand{\arraystretch}{1.08}

\resizebox{\linewidth}{!}{%
\begin{tabular}{l ccc @{\hspace{12pt}} ccc}
\toprule
& \multicolumn{3}{c}{OpenVLA}
& \multicolumn{3}{c}{$\pi_0$} \\
\cmidrule(lr){2-4}
\cmidrule(lr){5-7}

Method
& Detected failures $\uparrow$ & Lag $\Delta t\uparrow$ & bACC $\uparrow$
& Detected failures $\uparrow$ & Lag $\Delta t\uparrow$ & bACC $\uparrow$ \\
\midrule

STAC
& -- & -- & --
& \res{56.4}{0.0}
& \res{+0.076}{0.000}
& \res{0.578}{0.000}
\\

Diff-DAgger
& -- & -- & --
& \bestres{59.4}{0.0}
& \bestres{+0.120}{0.000}
& \res{0.531}{0.000}
\\

ACCEL
& -- & -- & --
& \res{46.4}{0.0}
& \res{-0.118}{0.000}
& \res{0.777}{0.000}
\\

\midrule

RND
& \res{49.9}{10.3}
& \res{-0.138}{0.038}
& \res{0.715}{0.048}
& \res{50.3}{0.8}
& \res{-0.044}{0.023}
& \bestres{0.874}{0.010}
\\

LogpZO
& \res{22.7}{13.6}
& \res{-0.074}{0.059}
& \res{0.531}{0.019}
& \res{50.6}{0.3}
& \res{-0.097}{0.010}
& \res{0.862}{0.007}
\\

FIPER
& \res{51.3}{5.5}
& \res{-0.487}{0.007}
& \bestres{0.808}{0.035}
& \res{50.9}{1.5}
& \res{-0.177}{0.021}
& \res{0.866}{0.005}
\\

\midrule

SAFE-MLP
& \res{36.7}{2.1}
& \res{-0.050}{0.005}
& \res{0.687}{0.001}
& \res{50.6}{0.3}
& \res{-0.019}{0.005}
& \res{0.869}{0.005}
\\

SAFE-LSTM
& \res{12.3}{10.9}
& \bestres{+0.100}{0.130}
& \res{0.528}{0.037}
& \res{54.0}{1.1}
& \res{-0.057}{0.014}
& \res{0.871}{0.010}
\\

\midrule
\rowcolor{blue!4}
ReAP (Ours)
& \bestres{60.9}{2.3}
& \res{+0.081}{0.011}
& \res{0.770}{0.008}
& \res{46.6}{2.3}
& \res{+0.045}{0.010}
& \res{0.865}{0.015}
\\

\bottomrule
\end{tabular}%
}
\end{table}

Table~\ref{tab:failure_onset} highlights the trade-off between
warning timeliness, failure coverage, and false alarms.
On OpenVLA, SAFE-LSTM's larger mean warning lead is accompanied
by limited failure coverage, while FIPER achieves higher bACC
with delayed alarms.
ReAP detects the most failures while still raising alarms before
the annotated onset on average.
On $\pi_0$, baselines that warn earlier than ReAP incur substantially
more false alarms, whereas those with slightly higher bACC
raise alarms after the annotated onset on average.
Across both policies, ReAP combines positive mean warning leads
with competitive balanced accuracy, supporting its ability to
provide advance warnings while controlling the false-alarm cost.

\subsection{Threshold Calibration Analysis}

\begin{table}[t]
\centering
\caption{
Threshold-calibration ablation for the detector on LIBERO-10.
}
\label{tab:threshold_ablation}

\vspace{3pt}

\small
\setlength{\tabcolsep}{3.2pt}
\renewcommand{\arraystretch}{1.08}

\begin{tabular}{lccc@{\hspace{10pt}}ccc}
\toprule
&
\multicolumn{3}{c}{\textbf{OpenVLA}}
&
\multicolumn{3}{c}{\textbf{$\pi_0$}}
\\
\cmidrule(lr){2-4}
\cmidrule(lr){5-7}

Calibration
& bTWA $\uparrow$
& bACC $\uparrow$
& T-det $\downarrow$
& bTWA $\uparrow$
& bACC $\uparrow$
& T-det $\downarrow$
\\
\midrule

\rowcolor{blue!4}
\textbf{Normalized functional}
& \bestres{0.671}{0.021}
& \bestres{0.744}{0.020}
& \res{0.450}{0.041}
& \bestres{0.750}{0.005}
& \bestres{0.838}{0.013}
& \res{0.466}{0.014}
\\

Normalized pointwise
& \res{0.574}{0.031}
& \res{0.618}{0.033}
& \bestres{0.147}{0.021}
& \res{0.703}{0.013}
& \res{0.790}{0.008}
& \res{0.318}{0.007}
\\

Absolute functional
& \res{0.535}{0.010}
& \res{0.645}{0.032}
& \res{0.685}{0.036}
& \res{0.690}{0.034}
& \res{0.825}{0.023}
& \res{0.532}{0.037}
\\

Absolute pointwise
& \res{0.486}{0.020}
& \res{0.546}{0.021}
& \res{0.203}{0.020}
& \res{0.711}{0.008}
& \res{0.791}{0.009}
& \bestres{0.299}{0.008}
\\

Stage functional
& \res{0.495}{0.003}
& \res{0.527}{0.017}
& \res{0.982}{0.010}
& \res{0.633}{0.016}
& \res{0.786}{0.019}
& \res{0.704}{0.042}
\\

Stage pointwise
& \res{0.522}{0.004}
& \res{0.638}{0.009}
& \res{0.784}{0.027}
& \res{0.657}{0.014}
& \res{0.782}{0.025}
& \res{0.488}{0.056}
\\

\bottomrule
\end{tabular}
\end{table}

The calibration procedure determines how continuous risk scores are converted
into failure alarms and therefore directly affects the trade-off between
detection accuracy and warning timeliness. We evaluate six variants obtained by
combining three temporal parameterizations with pointwise or functional
calibration. As shown in Table~\ref{tab:threshold_ablation}, normalized-time
functional calibration achieves the highest bTWA on both OpenVLA and $\pi_0$.
Pointwise calibration generally produces earlier alarms, as reflected by lower
T-det, but this gain is often accompanied by lower bACC and overall bTWA.

The distinction between pointwise and trajectory-level calibration helps explain these results. As described in Appendix~\ref{app:time_dependent_threshold}, pointwise calibration estimates a separate quantile at each temporal position. However, a successful rollout is counted as a false alarm if its score crosses the threshold at any monitored timestep, so controlling exceedances separately at each position does not directly control false alarms over the full rollout. Functional calibration instead uses the maximum standardized deviation along each successful trajectory to determine the band width, explicitly accounting for threshold crossings across time. This provides a possible explanation for the later alarms but higher bACC and bTWA observed with normalized-time functional calibration. Normalized time may further improve calibration by reducing temporal misalignment between rollouts of different durations. However, different monitoring methods may exhibit different temporal patterns in their risk scores, so the most effective calibration variant may vary across methods. We therefore select the calibration variant separately for each method on the validation split before fitting the final thresholds on held-out successful calibration rollouts.

\subsection{Fusion Analysis}

\paragraph{Fusion rule.}
\begin{table}[t]
\centering
\caption{
Comparison of OR and AND fusion rules across four policy--benchmark settings.
}
\label{tab:fusion_ablation}

\vspace{3pt}

\small
\setlength{\tabcolsep}{6pt}
\renewcommand{\arraystretch}{1.08}

\begin{tabular}{llccc}
\toprule
Benchmark
& Fusion
& bTWA $\uparrow$
& bACC $\uparrow$
& T-det $\downarrow$
\\
\midrule

OpenVLA / LIBERO-10
& \textbf{OR}
& \bestres{0.701}{0.010}
& \bestres{0.770}{0.008}
& \bestres{0.326}{0.029}
\\
& AND
& \res{0.653}{0.035}
& \res{0.737}{0.042}
& \res{0.530}{0.072}
\\

\midrule

$\pi_0$ / LIBERO-10
& \textbf{OR}
& \bestres{0.780}{0.008}
& \bestres{0.865}{0.015}
& \bestres{0.395}{0.025}
\\
& AND
& \res{0.726}{0.006}
& \res{0.804}{0.006}
& \res{0.532}{0.013}
\\

\midrule

$\pi_{0.5}$ / RoboTwin-8
& \textbf{OR}
& \bestres{0.701}{0.013}
& \res{0.785}{0.011}
& \bestres{0.306}{0.021}
\\
& AND
& \res{0.683}{0.024}
& \bestres{0.802}{0.023}
& \res{0.392}{0.032}
\\

\midrule

Fast-WAM / RoboTwin-8
& \textbf{OR}
& \bestres{0.651}{0.014}
& \bestres{0.779}{0.008}
& \bestres{0.592}{0.051}
\\
& AND
& \res{0.614}{0.014}
& \res{0.734}{0.014}
& \res{0.725}{0.036}
\\

\bottomrule
\end{tabular}
\end{table}

Table~\ref{tab:fusion_ablation} compares OR and AND fusion across four policy--benchmark settings. AND flags a rollout once both branches have raised at least one alarm, with the fused alarm time given by the later of their first-alarm times. If either branch never alarms, AND does not flag the rollout.
OR achieves higher bTWA and earlier
detection in every setting, together with higher bACC in three of
the four settings. Since bTWA explicitly penalizes false alarms on
successful rollouts, its consistent improvement indicates that
the gains in time-weighted failure detection outweigh the additional
false-alarm cost under this metric. OR also outperforms both
standalone branches in bTWA across all four settings in
Table~\ref{tab:main_offline_results}.

\begin{table}[t]
\centering
\caption{
Trajectory-level alarm overlap and fusion outcomes at $\alpha=0.15$
for seed~2. Each policy has 150 evaluation rollouts, including
75 failures for OpenVLA and 60 for $\pi_0$.
OR and AND denote the union and intersection of trajectory-level branch alarms, respectively. OR triggers at the first branch alarm, while AND triggers once both branches have alarmed. Alarms on successful rollouts are false alarms.
}
\label{tab:alarm_overlap}
\small
\setlength{\tabcolsep}{6pt}
\renewcommand{\arraystretch}{1.08}
\begin{tabular}{@{}lcccccc@{}}
\toprule
& \multicolumn{3}{c}{OpenVLA}
& \multicolumn{3}{c}{$\pi_0$} \\
\cmidrule(lr){2-4}\cmidrule(l){5-7}
Alarm pattern
& Total & Failed & Successful
& Total & Failed & Successful \\
\midrule
Both alarm
& 47 & 41 & 6
& 39 & 38 & 1 \\
Predictor only
& 24 & 17 & 7
& 5 & 4 & 1 \\
Detector only
& 12 & 4 & 8
& 10 & 9 & 1 \\
Neither alarms
& 67 & 13 & 54
& 96 & 9 & 87 \\
\midrule
OR
& 83 & 62 & 21
& 54 & 51 & 3 \\
AND
& 47 & 41 & 6
& 39 & 38 & 1 \\
\bottomrule
\end{tabular}
\end{table}

\begin{figure*}[t]
    \centering

    \begin{subfigure}[t]{0.98\textwidth}
        \centering
        \includegraphics[width=\linewidth]{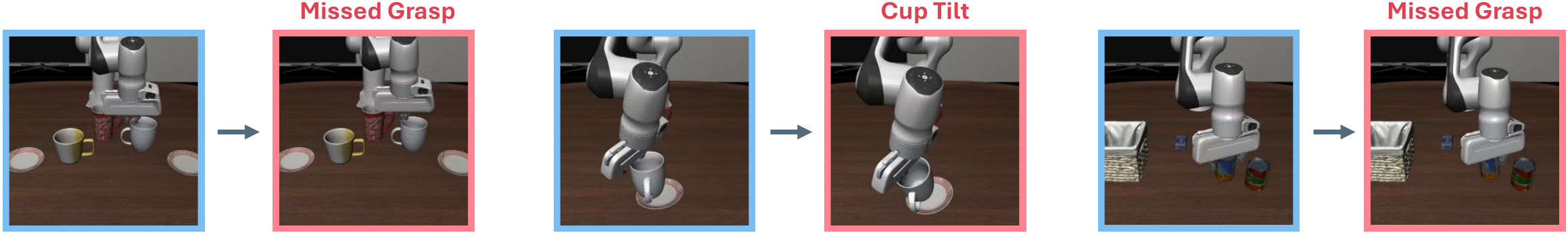}
        \caption{}
        \label{fig:predictor_only_cases}
    \end{subfigure}

    \vspace{5pt}

    \begin{subfigure}[t]{0.98\textwidth}
        \centering
        \includegraphics[width=\linewidth]{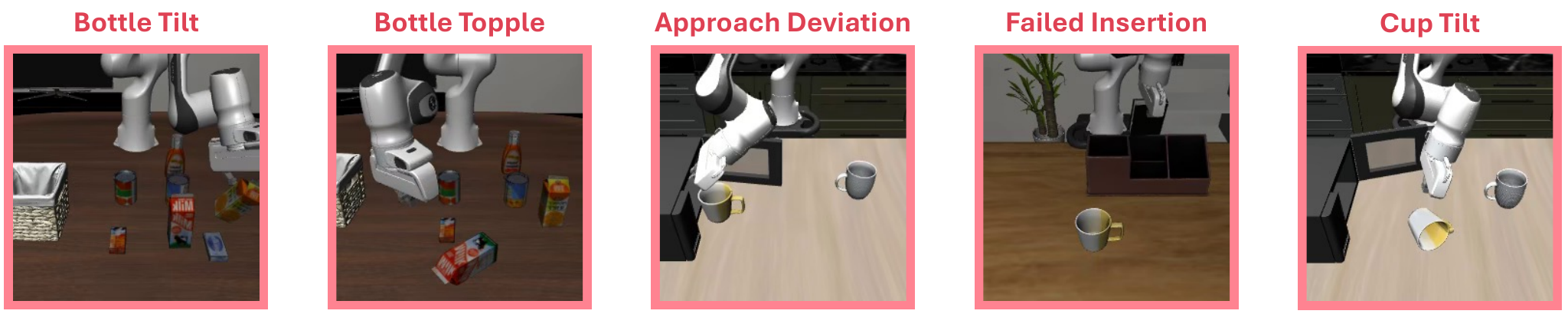}
        \caption{}
        \label{fig:detector_only_cases}
    \end{subfigure}

    \caption{
    Qualitative examples of complementary detector and predictor alarms.
    (a) Predictor-only cases. Blue boxes show the observation when the predictor
    raises an alarm, while red boxes show a later state where the failure becomes
    visually apparent.
    (b) Detector-only cases, where the observed state already contains clear
    visual failure evidence.
    }
    \label{fig:complementary_cases}
\end{figure*}

To examine the source of this improvement,
Table~\ref{tab:alarm_overlap} summarizes branch alarm overlap across
150 evaluation rollouts for each LIBERO-10 policy at $\alpha=0.15$. On OpenVLA, the predictor uniquely detects 17 failed
rollouts, while the detector uniquely detects four. On $\pi_0$,
the corresponding counts are four and nine. By retaining these
branch-specific warnings, OR detects 21 additional failed OpenVLA
rollouts and 13 additional failed $\pi_0$ rollouts compared with
AND, while introducing 15 and two additional false alarms,
respectively. AND misses failures detected by only one branch, while OR retains these warnings and provides broader failure coverage.

Figure~\ref{fig:complementary_cases} provides qualitative examples of these
branch-specific alarms. In predictor-only cases, the current observation can
remain visually plausible while the proposed action already points toward an
undesirable future outcome, allowing the predictor to raise an alarm before
clear failure evidence becomes observable. In detector-only cases, the failure
has already manifested as a salient visual deviation, such as a tipped object
or failed insertion, making it more directly recognizable from the current
observation. These examples are consistent with the overlap analysis and
illustrate the complementary roles of current-state failure recognition and
action-conditioned future-risk prediction.

\section{Time-Varying Threshold Calibration}
\label{app:time_dependent_threshold}

We calibrate alarm thresholds using held-out successful rollouts to account for variation in risk scores over the course of execution.
For rollout $i$, let $T_i$ denote its full length in environment frames. A monitoring branch records $L_i$ risk scores, forming the sequence $\mathbf{s}_i=(s_{i,0},\ldots,s_{i,L_i-1})$. Each score $s_{i,j}$ is associated with an original frame index $f_{i,j}$, and larger scores indicate higher predicted failure risk. For the predictor, $f_{i,j}$ refers to the decision frame at which the proposed action chunk is evaluated.

We calibrate thresholds separately for the detector and predictor. For each branch, let $\mathcal{D}_{\mathrm{cal}}=\{\mathbf{s}_1,\ldots,\mathbf{s}_N\}$ denote the risk-score sequences from held-out successful rollouts. These sequences are used to estimate a time-varying upper threshold $\delta_j(\alpha)$, where $j$ indexes monitored positions under the chosen temporal parameterization. We evaluate $\alpha\in\{0.05,0.10,0.15,0.20,0.25\}$ and use the same $\alpha$ for both branches. Each branch raises an alarm at the first position satisfying $s_{i,j}\geq\delta_j(\alpha)$, and the two branches' alarms are combined using the OR rule described in the main paper.

\begin{figure*}[t]
    \centering
    \includegraphics[
        width=0.96\textwidth
    ]{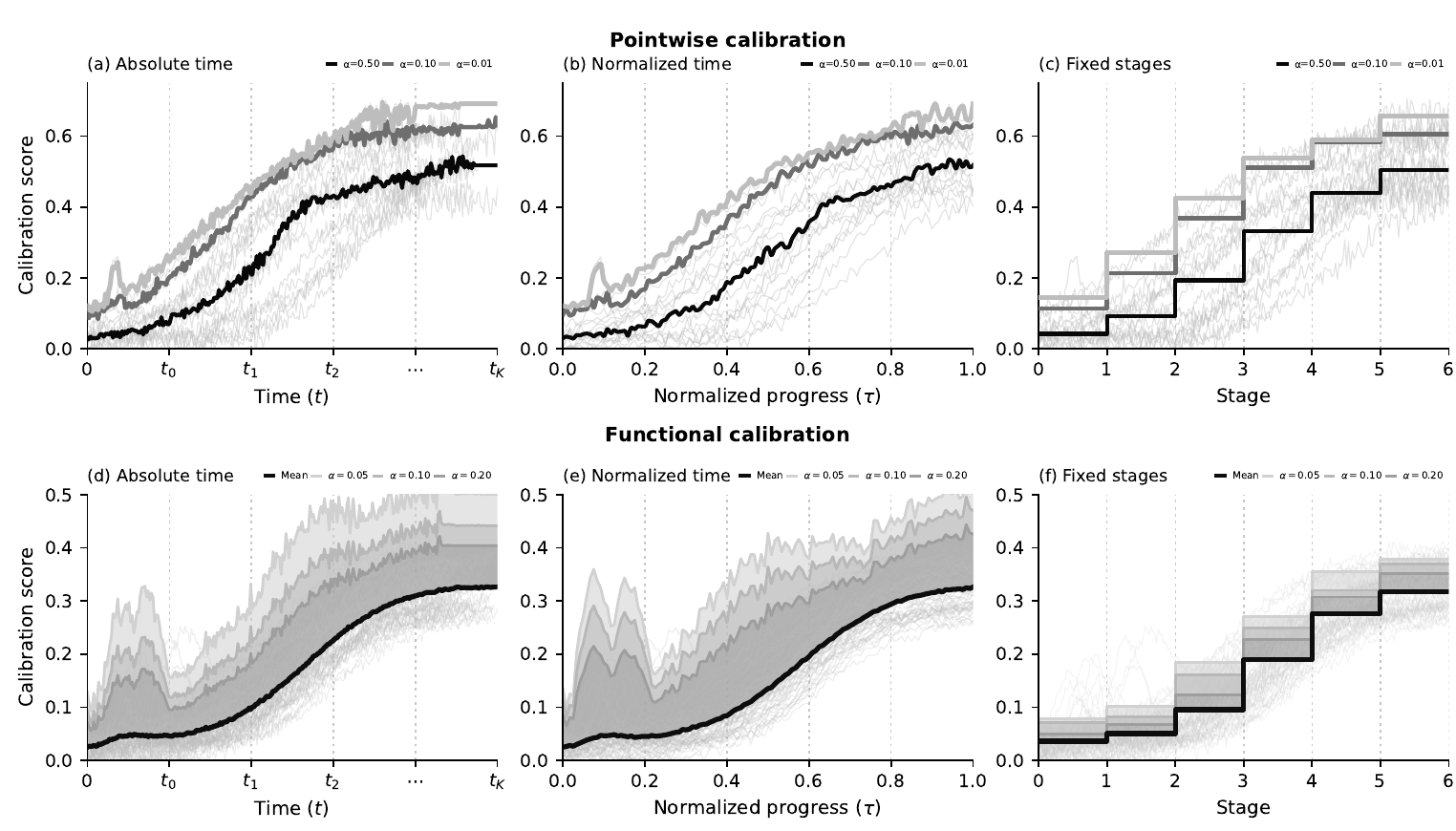}
    \caption{
    Illustration of the six time-varying threshold calibration variants.
    The top row shows pointwise empirical-quantile calibration and the bottom
    row shows functional calibration.
    From left to right, the columns use absolute time, normalized execution
    progress, and fixed execution stages.
    Thin gray curves denote successful calibration trajectories.
    Dark curves in the top row indicate thresholds at representative
    significance levels $\alpha$, while the bottom row shows the nominal
    trajectory together with the corresponding functional upper bands.
    Smaller $\alpha$ produces a more conservative upper boundary.
    }
    \label{fig:conformal}
\end{figure*}

A single global threshold ignores temporal variation in the score distribution
over the course of an execution.
Successful manipulation trajectories can exhibit different nominal risk levels
during phases such as free-space motion, contact, grasping, and placement.
We therefore consider three temporal parameterizations consisting of normalized
time, absolute time, and fixed execution stages. Each temporal parameterization is paired with either pointwise empirical-quantile calibration or functional conformal calibration, resulting in the six variants illustrated in Figure~\ref{fig:conformal}.

\subsection{Temporal Parameterizations}

\paragraph{Absolute time.}
Absolute-time calibration aligns successful trajectories by their
monitored score index $j$. Let
$L_{\max}=\max_i L_i$ be the maximum number of recorded scores among
successful calibration trajectories. Shorter score sequences are
extended to $L_{\max}$ by repeating their final score. Thresholds are
then constructed as a function of the absolute score index.

This parameterization can be applied online using only the current timestep and
does not require knowing the final duration of the test rollout. However, when
rollout lengths vary, the same absolute timestep may correspond to different
stages of task execution across trajectories. If a test rollout contains more than $L_{\max}$ recorded scores,
the final calibrated threshold is used for all subsequent positions. 

\paragraph{Normalized time.}
Normalized-time calibration aligns trajectories according to relative execution
progress rather than absolute timestep. For calibration rollout $i$, we define
normalized progress as
\[
u_{i,t}=\frac{t}{T_i-1},
\]
and linearly interpolate its risk trajectory onto a common grid of $G$
normalized positions between 0 and 1. We use $G=100$ in all experiments.
This representation removes differences in raw rollout duration during
calibration and aligns trajectories according to their relative progress.

At deployment, the final duration of an ongoing rollout is unavailable. We
therefore define a reference rollout length
\[
T_{\mathrm{ref}}
=
\frac{1}{N}\sum_{i=1}^{N} T_i
\]
using only successful calibration rollouts. The current timestep is mapped to
normalized progress using $T_{\mathrm{ref}}$ and then to the corresponding
position on the calibrated grid. If execution exceeds $T_{\mathrm{ref}}$, the
threshold at the final normalized position is retained. In this way, no information about
the final duration of the test rollout is used.

\paragraph{Staged time.}
Staged-time calibration provides a coarser representation of relative execution
progress. Each successful calibration trajectory is divided into $K$ equally
spaced stages according to normalized time, and all risk scores within the same
stage share one threshold. We use $K=10$ in all experiments. This produces a
piecewise-constant threshold trajectory while preserving the coarse progression
of risk over an execution.

For pointwise calibration, successful scores assigned to the same stage
are pooled when estimating that stage's upper quantile. For functional
calibration, we instead take the maximum score within each stage of
each successful rollout. This yields one $K$-dimensional stage-level
trajectory per rollout, to which the functional-band procedure is
applied.
At deployment, the current timestep is mapped to a stage using the same
reference rollout length $T_{\mathrm{ref}}$. If execution extends beyond
$T_{\mathrm{ref}}$, the threshold of the final stage is retained.

\subsection{Pointwise Empirical Quantile Calibration}

For normalized-time and absolute-time parameterizations, pointwise calibration
estimates an upper empirical quantile independently at each aligned temporal
position.
Let $\widetilde{s}_{i,j}$ denote the aligned score of calibration rollout $i$
at temporal position $j$.
The threshold is

\begin{equation}
\delta_j(\alpha)
=
Q_{1-\alpha}
\left(
\widetilde{s}_{1,j},
\ldots,
\widetilde{s}_{N,j}
\right),
\label{eq:pointwise_threshold}
\end{equation}

where $Q_{1-\alpha}$ denotes the empirical $(1-\alpha)$-quantile.
For the staged-time parameterization, the same construction is applied to all
successful scores assigned to each stage.

Pointwise calibration follows local changes in the nominal score distribution,
but different temporal positions are calibrated independently.
A pointwise significance level therefore does not imply that an entire
successful trajectory remains below the threshold simultaneously.
We use empirical upper quantiles directly and interpret $\alpha$ as the
significance level indexing the resulting operating point.

\subsection{Functional Conformal Calibration}

Conformal prediction provides a distribution-free framework for calibration
under exchangeability
\citep{vovk2005algorithmic,angelopoulos2023conformal}.
Functional conformal prediction extends this construction to function-valued
observations and simultaneous prediction bands
\citep{diquigiovanni2022conformal,diquigiovanni2025band}.
Trajectory-level calibration has also been adopted in recent runtime failure
monitoring methods
\citep{xu2025faildetect,gu2025safe,park2026hideandseek,
zhang2026foresight}.

In our setting, an entire successful risk trajectory is treated as one
functional observation.
For each temporal parameterization, the calibration set is divided into two
disjoint subsets, $\mathcal{D}_A$ and $\mathcal{D}_B$.
The first estimates the nominal trajectory and its temporal variation, while
the second calibrates the width of the upper band. In practice, approximately 30\% of successful
trajectories estimate the nominal trajectory and modulation function,
while approximately 70\% calibrate the final band width.

For normalized-time and absolute-time representations, let
$\widetilde{s}_{i,j}$ denote the aligned score at temporal position $j$.
We compute the mean successful trajectory from $\mathcal{D}_A$ as
$\mu_j =
|\mathcal{D}_A|^{-1}
\sum_{\mathbf{s}_i\in\mathcal{D}_A}\widetilde{s}_{i,j}$.

Successful trajectories can exhibit different variability across execution.
We therefore use a temporal modulation function $m_j$.
For each trajectory in $\mathcal{D}_A$, we compute its maximum absolute
deviation
$d_i=\max_j|\widetilde{s}_{i,j}-\mu_j|$.
To estimate the modulation function, let $n_A=|\mathcal D_A|$ and
$k_\alpha=\lceil(n_A+1)(1-\alpha)\rceil$.
If $k_\alpha\leq n_A$, let $\gamma_\alpha$ be the
$k_\alpha$-th smallest value among the deviations $\{d_i\}$ and retain
$\mathcal H=\{\mathbf s_i\in\mathcal D_A:d_i\leq\gamma_\alpha\}$.
Otherwise, we retain all trajectories, setting $\mathcal H=\mathcal D_A$.
The modulation function is then
$m_j=\max_{\mathbf s_i\in\mathcal H}|\tilde s_{i,j}-\mu_j|+\epsilon$.

For each trajectory in $\mathcal{D}_B$, the one-sided functional
nonconformity score is

\begin{equation}
R_i
=
\max_j
\frac{
\widetilde{s}_{i,j}-\mu_j
}{
m_j
}.
\label{eq:functional_nonconformity}
\end{equation}

Let $q_{1-\alpha}$ denote the empirical $(1-\alpha)$-quantile of these
nonconformity scores.
The resulting upper band is

\begin{equation}
\delta_j(\alpha)
=
\mu_j
+
q_{1-\alpha}m_j.
\label{eq:functional_band}
\end{equation}

The coefficient $q_{1-\alpha}$ is shared across the temporal domain, while
$m_j$ allows the band width to vary with the nominal score variability.

The same construction is applied to the staged-time parameterization.
In this case, the center and modulation are indexed by stage, yielding
piecewise-constant values $\mu_k$ and $m_k$.
For trajectory $i$, the nonconformity score is computed by taking the maximum
standardized upward deviation across all timesteps using the corresponding
stage statistics, and the threshold for stage $k$ is
$\delta_k(\alpha)=\mu_k+q_{1-\alpha}m_k$.

We estimate the final band coefficient using empirical quantiles. The resulting band targets a coverage level of $1-\alpha$ for successful trajectories, without a finite-sample coverage guarantee. We assess its actual false-alarm behavior on the held-out evaluation split.

\subsection{Independent Calibration and OR Fusion}

The significance level $\alpha$ controls the sensitivity of each calibrated
monitoring branch.
A smaller $\alpha$ generally produces a higher and more conservative threshold,
which tends to reduce false alarms but may delay or miss failures.
A larger $\alpha$ lowers the threshold and increases sensitivity, typically
yielding earlier alarms at the cost of more false positives.

Previous calibration-based monitors typically associate $\alpha$ with the
monitoring score that directly determines the final alarm
\citep{gu2025safe,xu2025faildetect,park2026hideandseek,
zhang2026foresight}.
For example, SAFE~\citep{gu2025safe} uses functional conformal prediction to bound the probability of raising any alarm during a successful rollout by $\alpha$, under the exchangeability assumption. In \methodname{}, the detector and predictor are calibrated separately at the same $\alpha$, and their alarms are combined using an OR rule. A branch-level guarantee therefore does not directly imply the same guarantee for the combined monitor.

Let $A_{\mathrm{det}}$ and $A_{\mathrm{pred}}$ denote the events that the respective branches raise an alarm on a successful rollout. If each branch satisfies $\Pr(A_{\mathrm{det}})\leq\alpha$ and $\Pr(A_{\mathrm{pred}})\leq\alpha$, the union bound gives
\[
\Pr(A_{\mathrm{det}}\cup A_{\mathrm{pred}})
\leq \Pr(A_{\mathrm{det}})+\Pr(A_{\mathrm{pred}})
\leq 2\alpha.
\]
This bound does not require independence between the two branches. It describes how branch-level false-alarm guarantees would translate under OR fusion. In our implementation, $\alpha$ serves as a nominal branch-level calibration parameter, and the false-alarm rate of the combined monitor is evaluated empirically on the held-out evaluation split.



\section{Rollout Visualization}

\label{app:visualization}

\begin{figure}[!htbp]
    \centering
    \includegraphics[
        width=0.88\linewidth
    ]{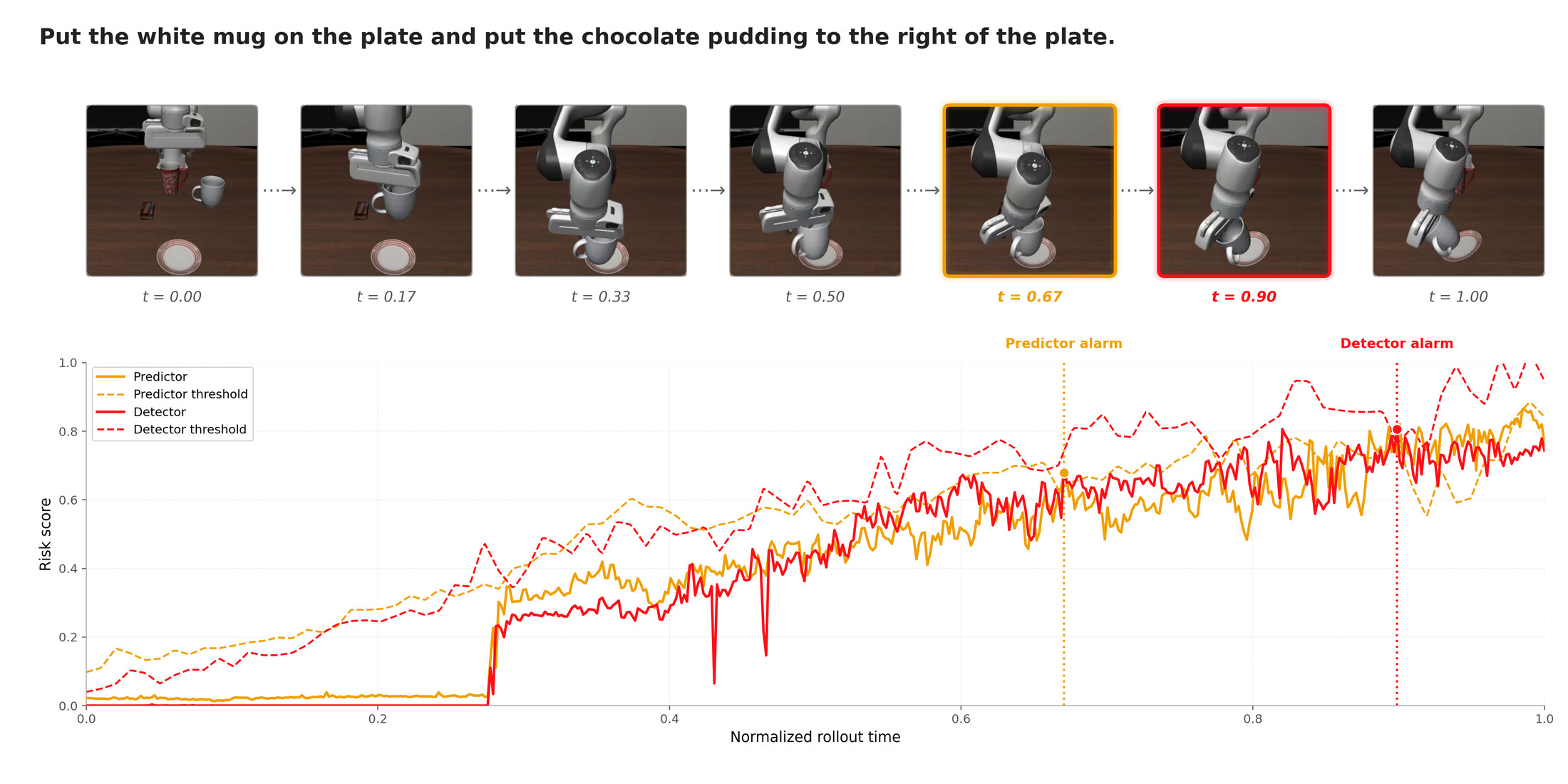}
    \caption{Qualitative monitoring example on LIBERO-10 with OpenVLA. The mug tilts during placement, leaving the task incomplete. The predictor raises an alarm first, followed by the detector.}
    \label{fig:qual_libero_both_1}
\end{figure}

\begin{figure}[!htbp]
    \centering
    \includegraphics[
        width=0.88\linewidth
    ]{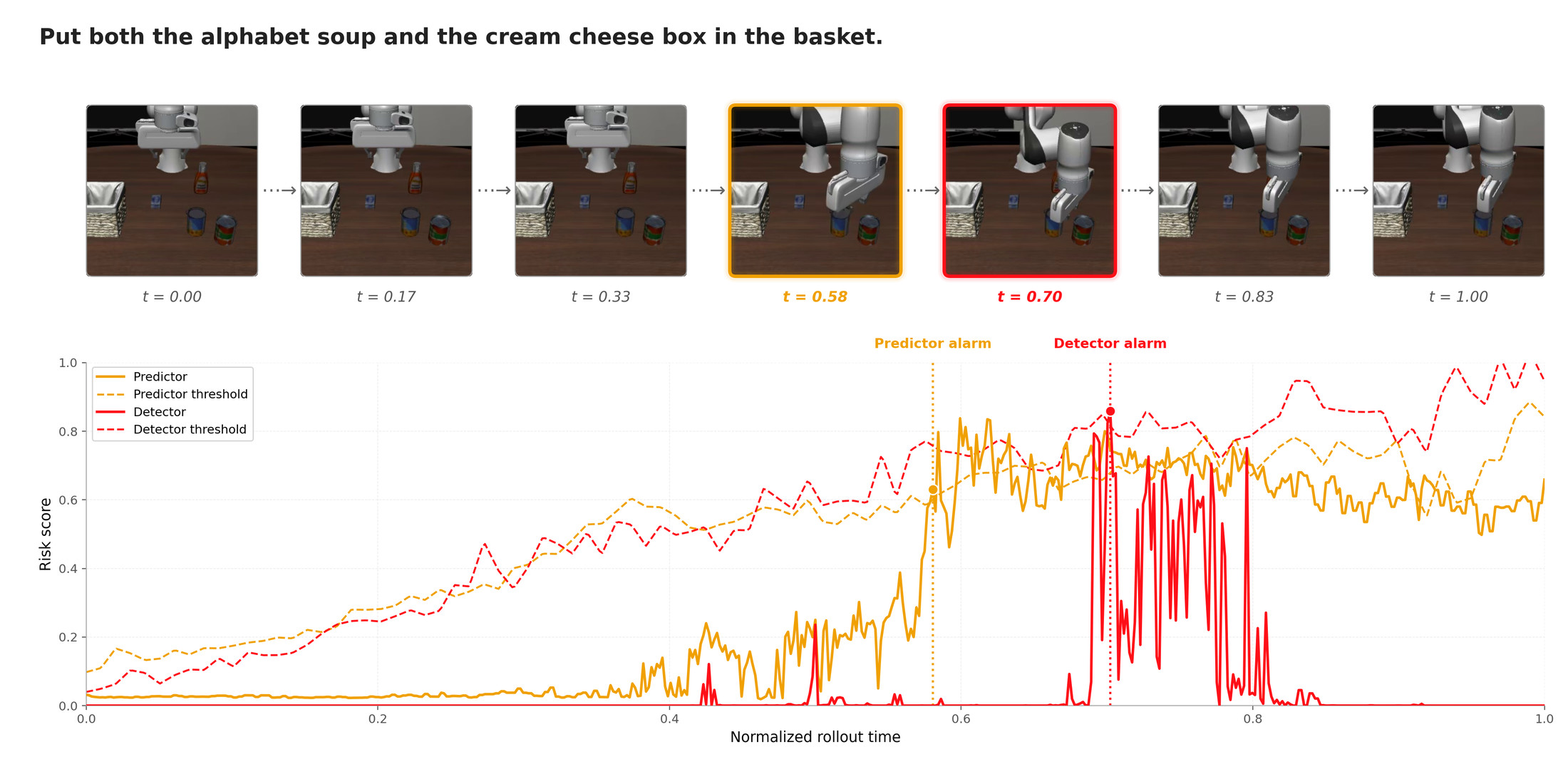}
    \caption{Qualitative monitoring example on LIBERO-10 with OpenVLA. The gripper deviates while approaching the target objects. The predictor raises an alarm first, followed by the detector.}
    \label{fig:qual_libero_both_2}
\end{figure}

\begin{figure}[!htbp]
    \centering
    \includegraphics[
        width=0.88\linewidth
    ]{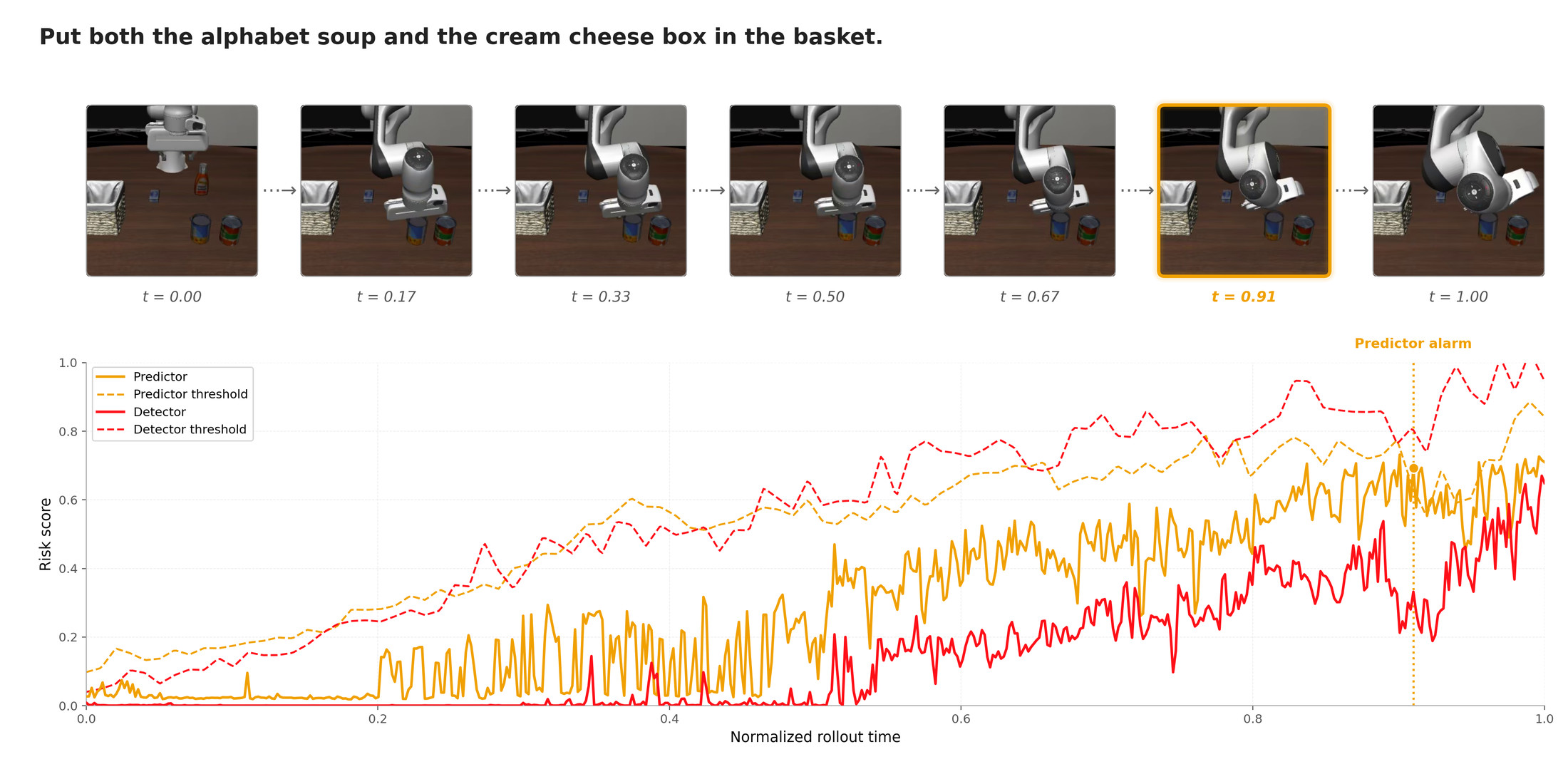}
    \caption{Qualitative monitoring example on LIBERO-10 with OpenVLA. The gripper rotates excessively during grasping. Only the predictor raises an alarm.}
    \label{fig:qual_libero_predictor}
\end{figure}

\begin{figure}[!htbp]
    \centering
    \includegraphics[
        width=0.88\linewidth
    ]{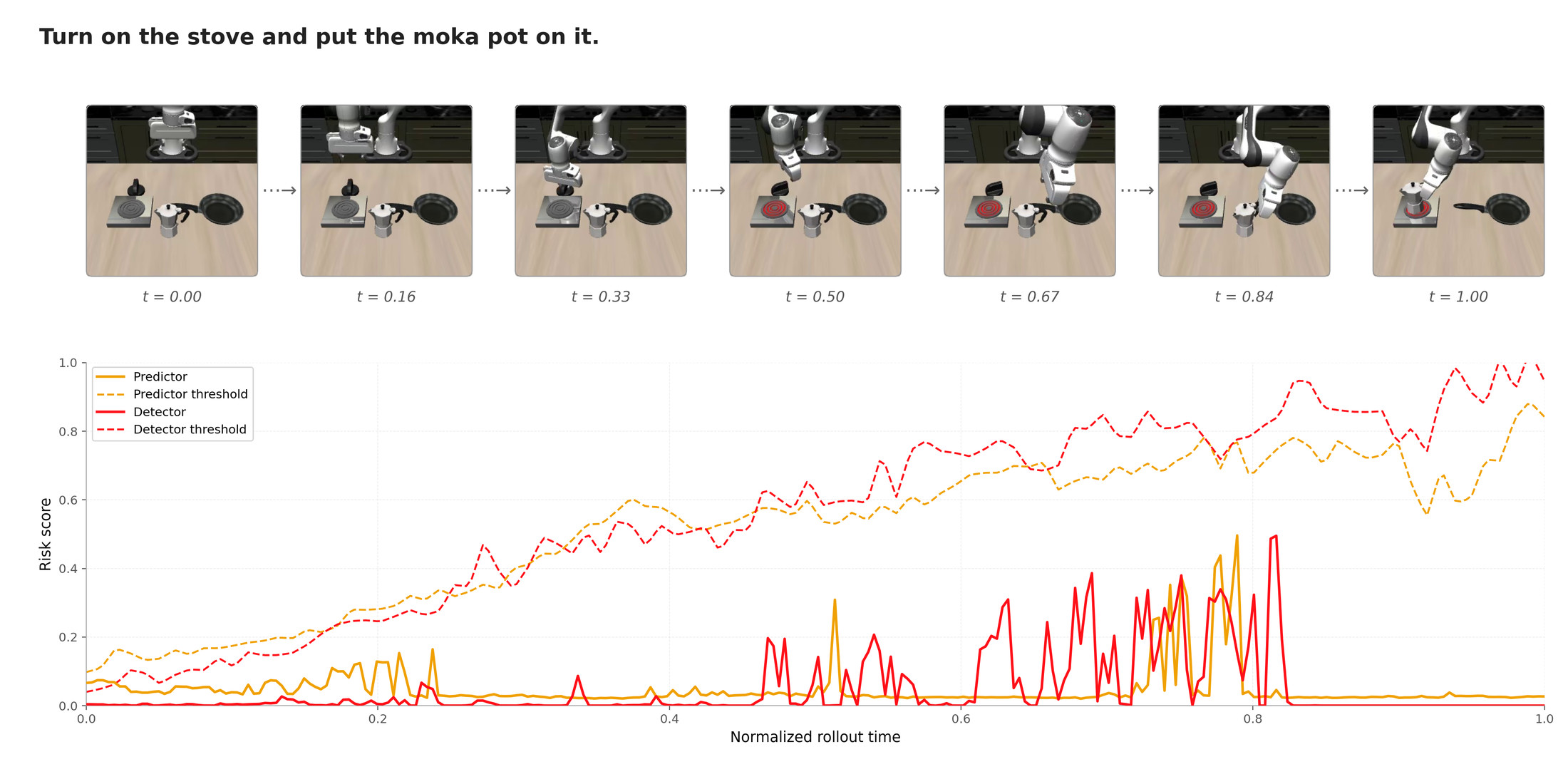}
    \caption{Qualitative monitoring example on LIBERO-10 with OpenVLA. The robot successfully turns on the stove and places the moka pot on it.}
    \label{fig:qual_libero_success_1}
\end{figure}

\begin{figure}[!htbp]
    \centering
    \includegraphics[
        width=0.88\linewidth
    ]{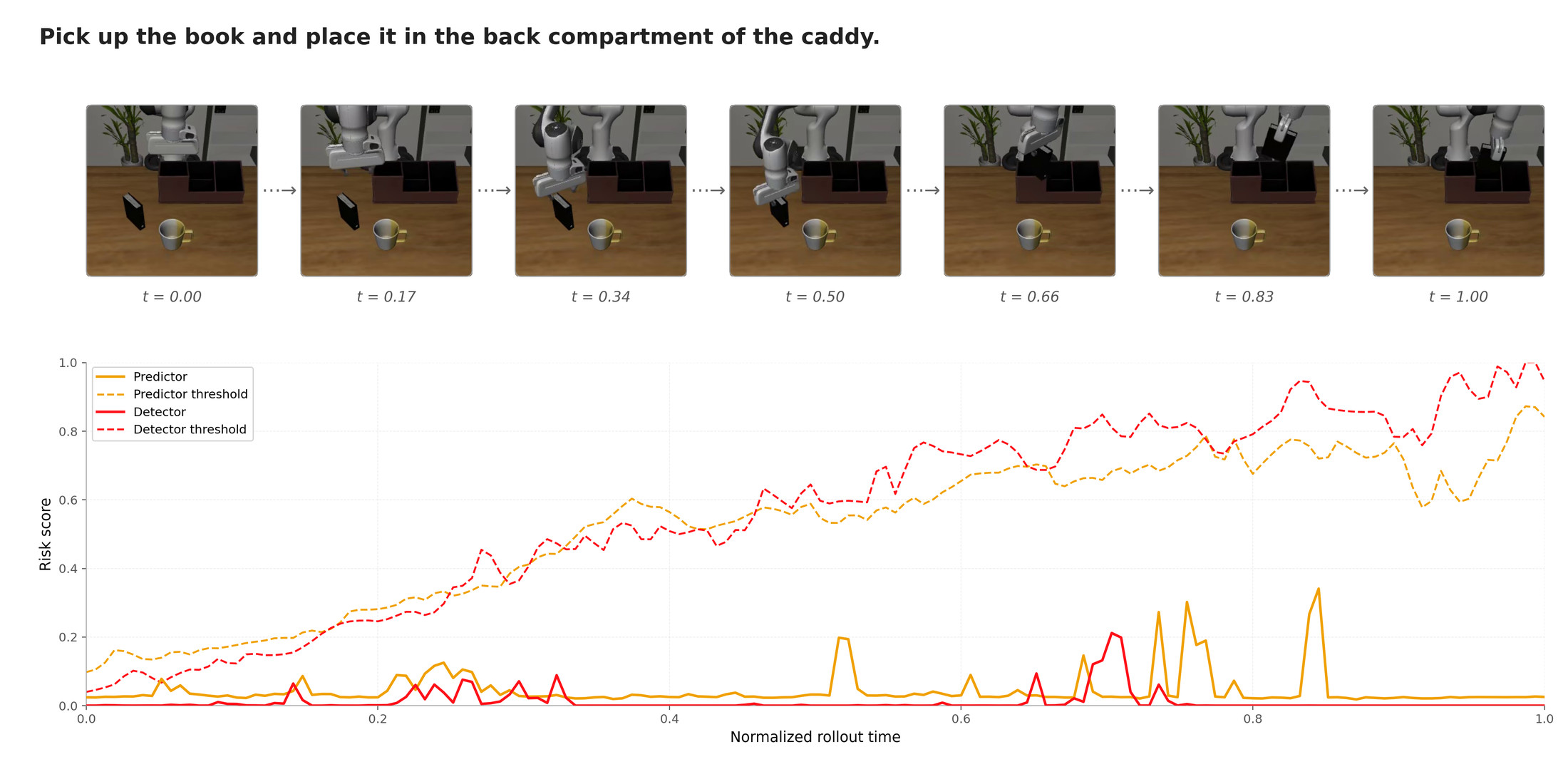}
    \caption{Qualitative monitoring example on LIBERO-10 with OpenVLA. The robot successfully picks up the book and places it in the back compartment of the caddy.}
    \label{fig:qual_libero_success_2}
\end{figure}

We visualize five representative OpenVLA rollouts on LIBERO-10 in Figures~\ref{fig:qual_libero_both_1}--\ref{fig:qual_libero_success_2}, including three failed and two successful executions. The first two failures involve a mug tilting during placement and a gripper deviating from the intended trajectory while approaching the target object. In both cases, the predictor raises an alarm first, followed by the detector as execution continues and the failure becomes more visually apparent. In the third failure, the gripper rotates excessively during grasping. The detector remains below its threshold, while the action-conditioned predictor raises an alarm for such excessive rotation. Neither component raises an alarm in the two successful rollouts. These examples illustrate how the detector and predictor can provide complementary monitoring signals during execution.

\begin{figure}[!htbp]
    \centering
    \includegraphics[
        width=0.88\linewidth
    ]{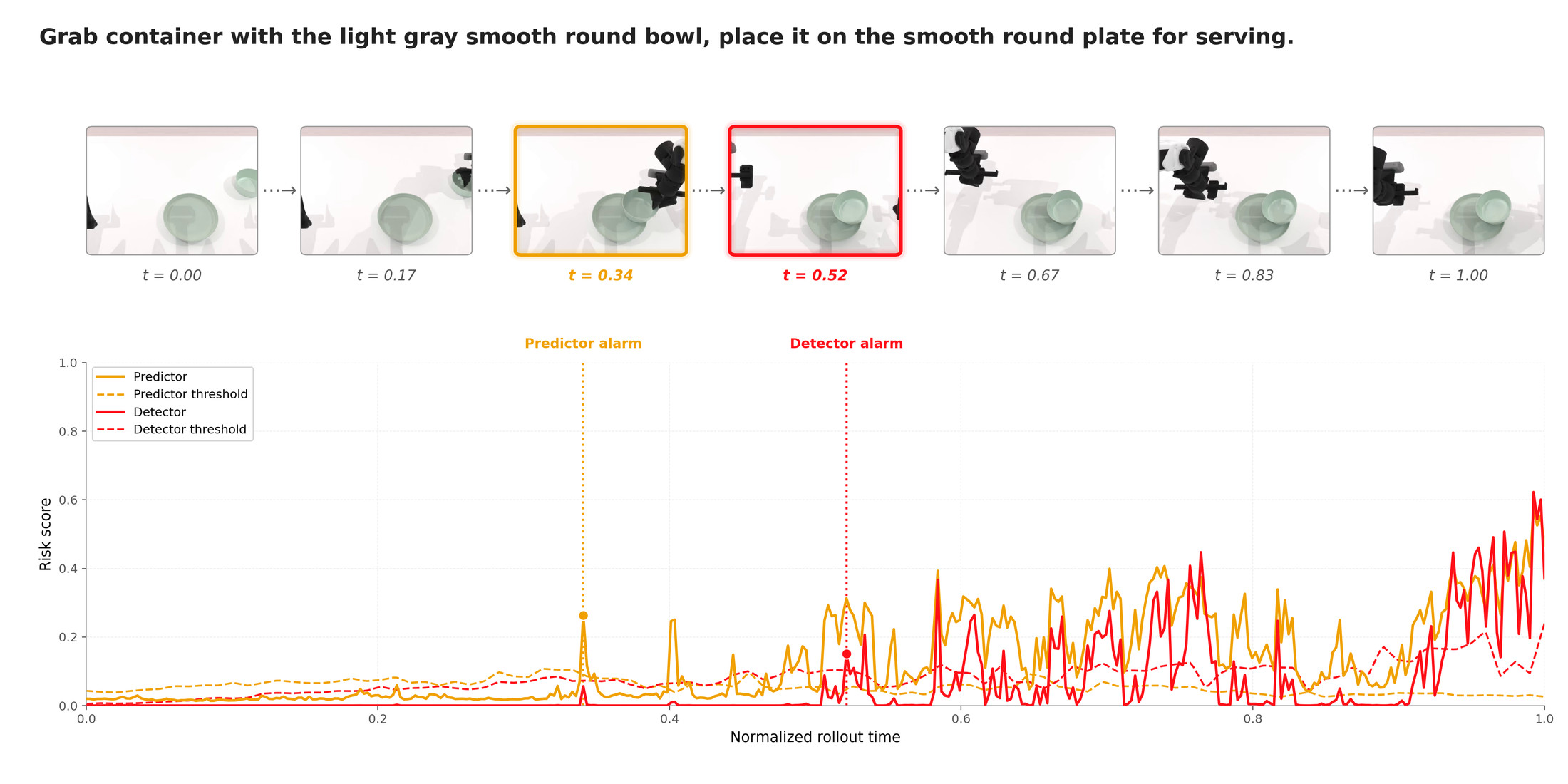}
    \caption{Qualitative monitoring example on RoboTwin-8 with $\pi_{0.5}$. The bowl is placed off-center at the edge of the plate. The predictor raises an alarm first, followed by the detector.}
    \label{fig:qual_robotwin_both_1}
\end{figure}

\begin{figure}[!htbp]
    \centering
    \includegraphics[
        width=0.88\linewidth
    ]{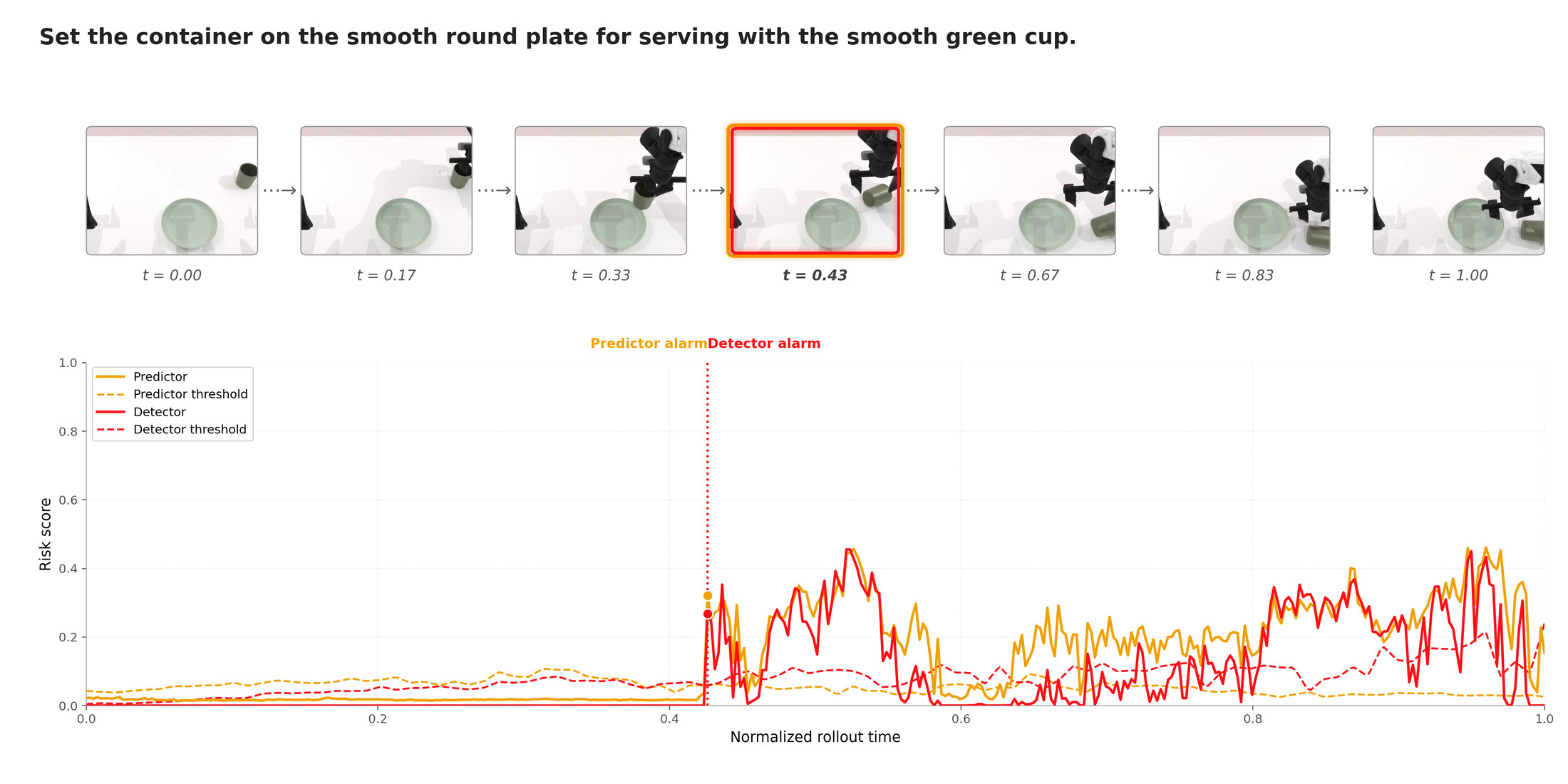}
    \caption{Qualitative monitoring example on RoboTwin-8 with $\pi_{0.5}$. The cup tips over beside the plate. The detector and predictor raise alarms at nearly the same time.}
    \label{fig:qual_robotwin_both_2}
\end{figure}

\begin{figure}[!htbp]
    \centering
    \includegraphics[
        width=0.88\linewidth
    ]{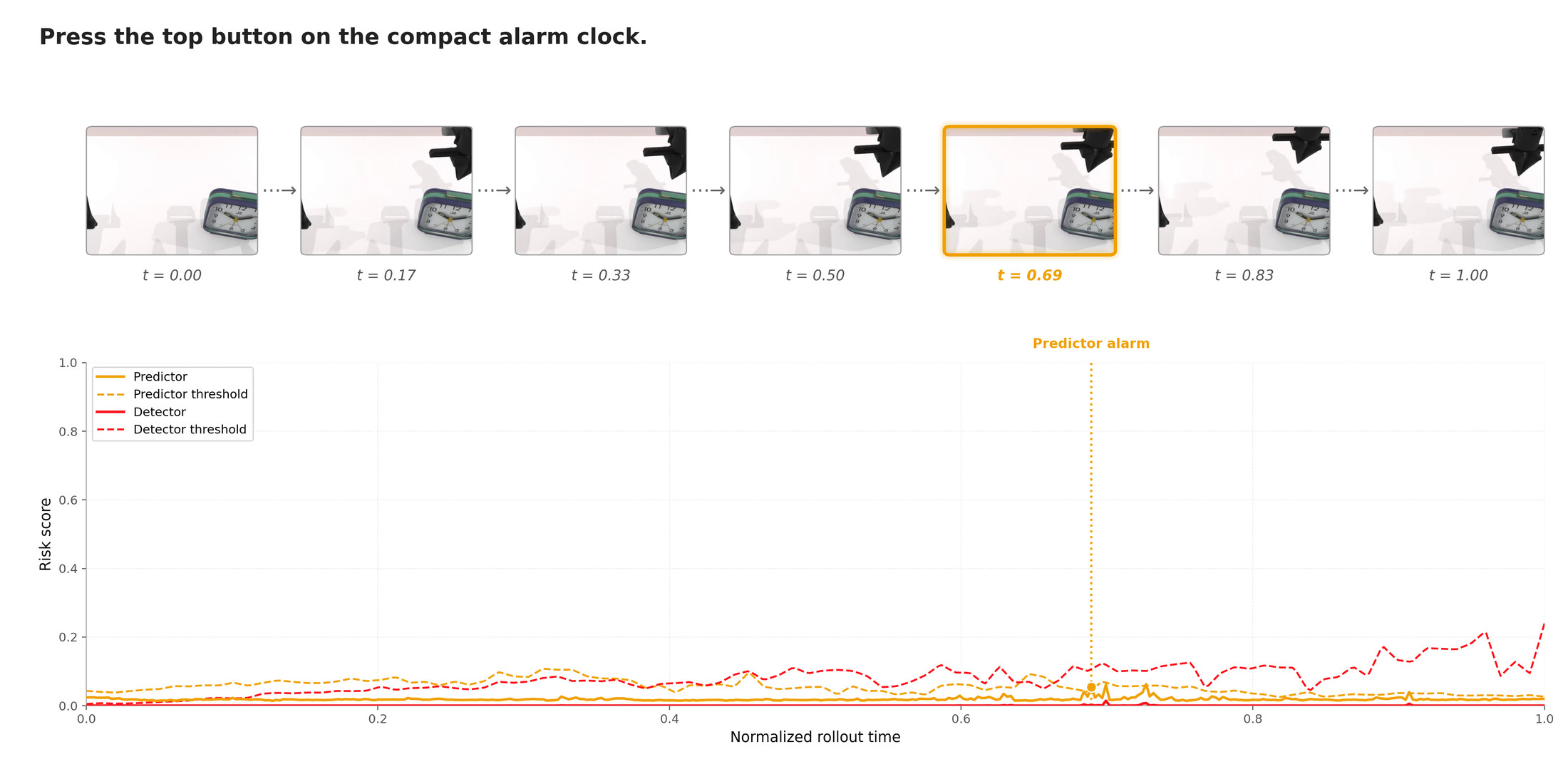}
    \caption{Qualitative monitoring example on RoboTwin-8 with $\pi_{0.5}$. The gripper remains near the clock without visibly completing the button press. Only the predictor raises an alarm.}
    \label{fig:qual_robotwin_predictor}
\end{figure}

\begin{figure}[!htbp]
    \centering
    \includegraphics[
        width=0.88\linewidth
    ]{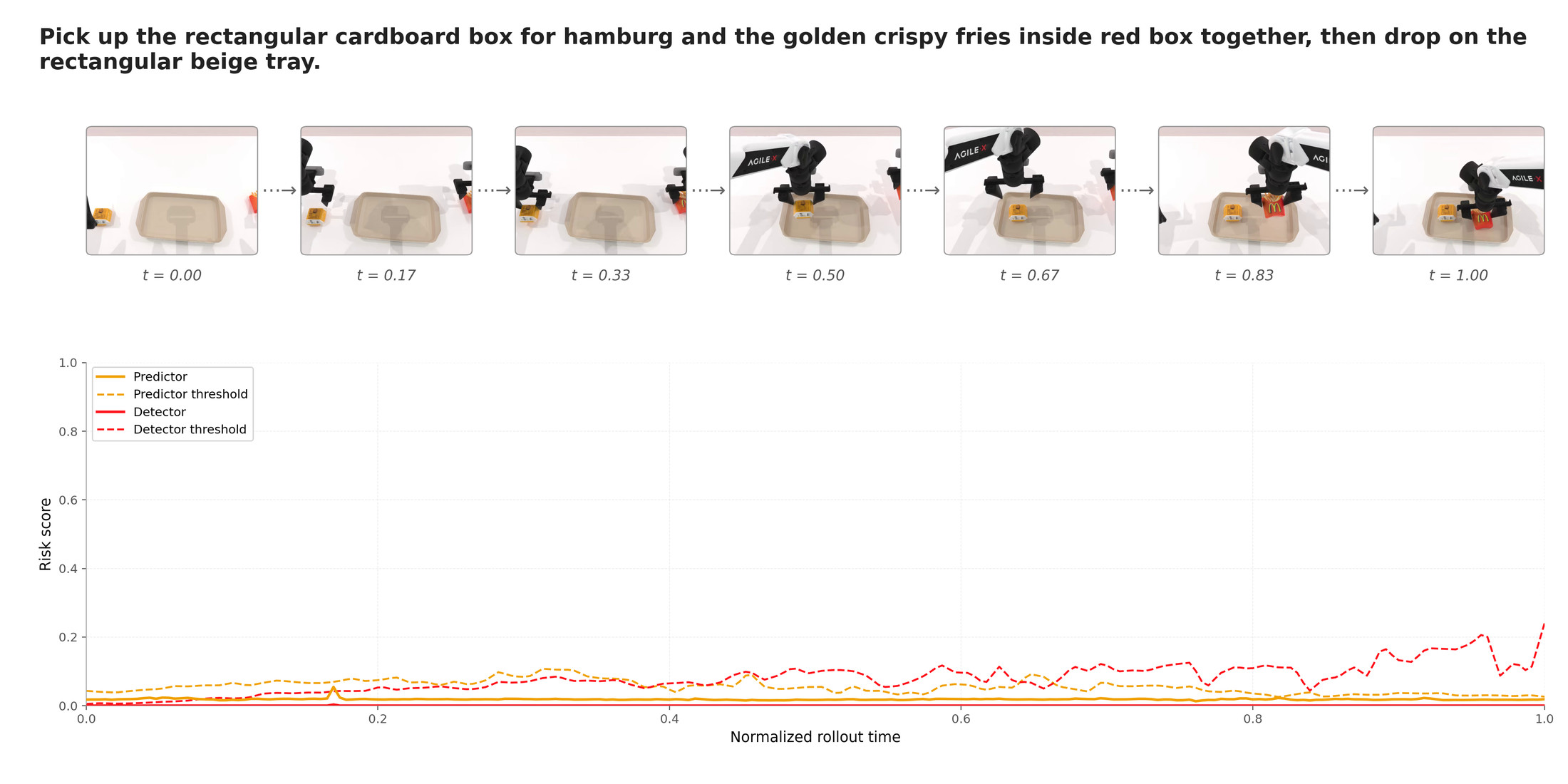}
    \caption{Qualitative monitoring example on RoboTwin-8 with $\pi_{0.5}$. The robot successfully transfers the hamburger box and fries container onto the tray.}
    \label{fig:qual_robotwin_success}
\end{figure}

\begin{figure}[!htbp]
    \centering
    \includegraphics[
        width=0.88\linewidth
    ]{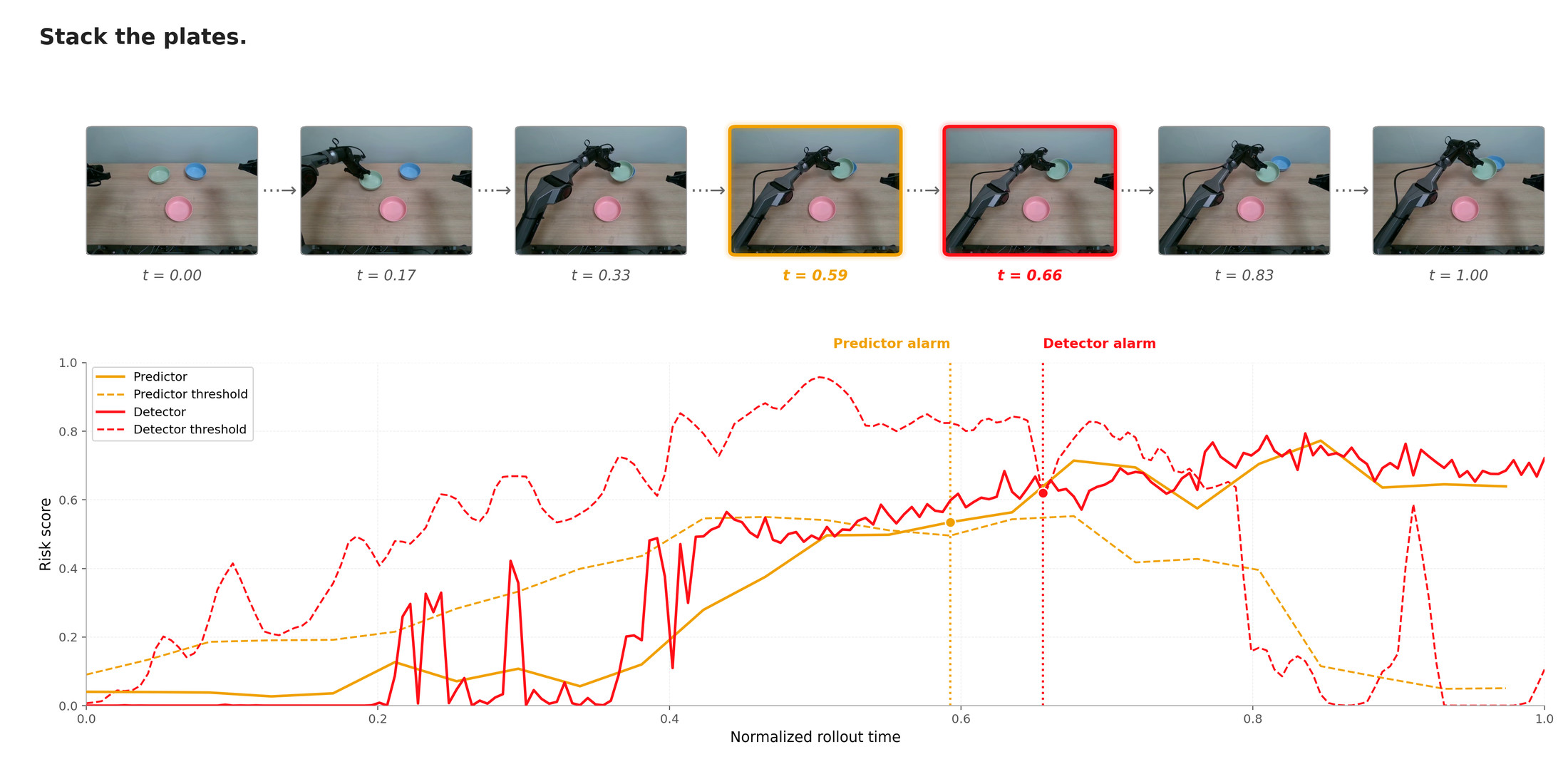}
    \caption{Qualitative monitoring example on Stack Plates with $\pi_{0.5}$. The robot repeatedly attempts to manipulate the plates. The predictor raises an alarm first, followed by the detector.}
    \label{fig:qual_stack_both_1}
\end{figure}

\begin{figure}[!htbp]
    \centering
    \includegraphics[
        width=0.88\linewidth
    ]{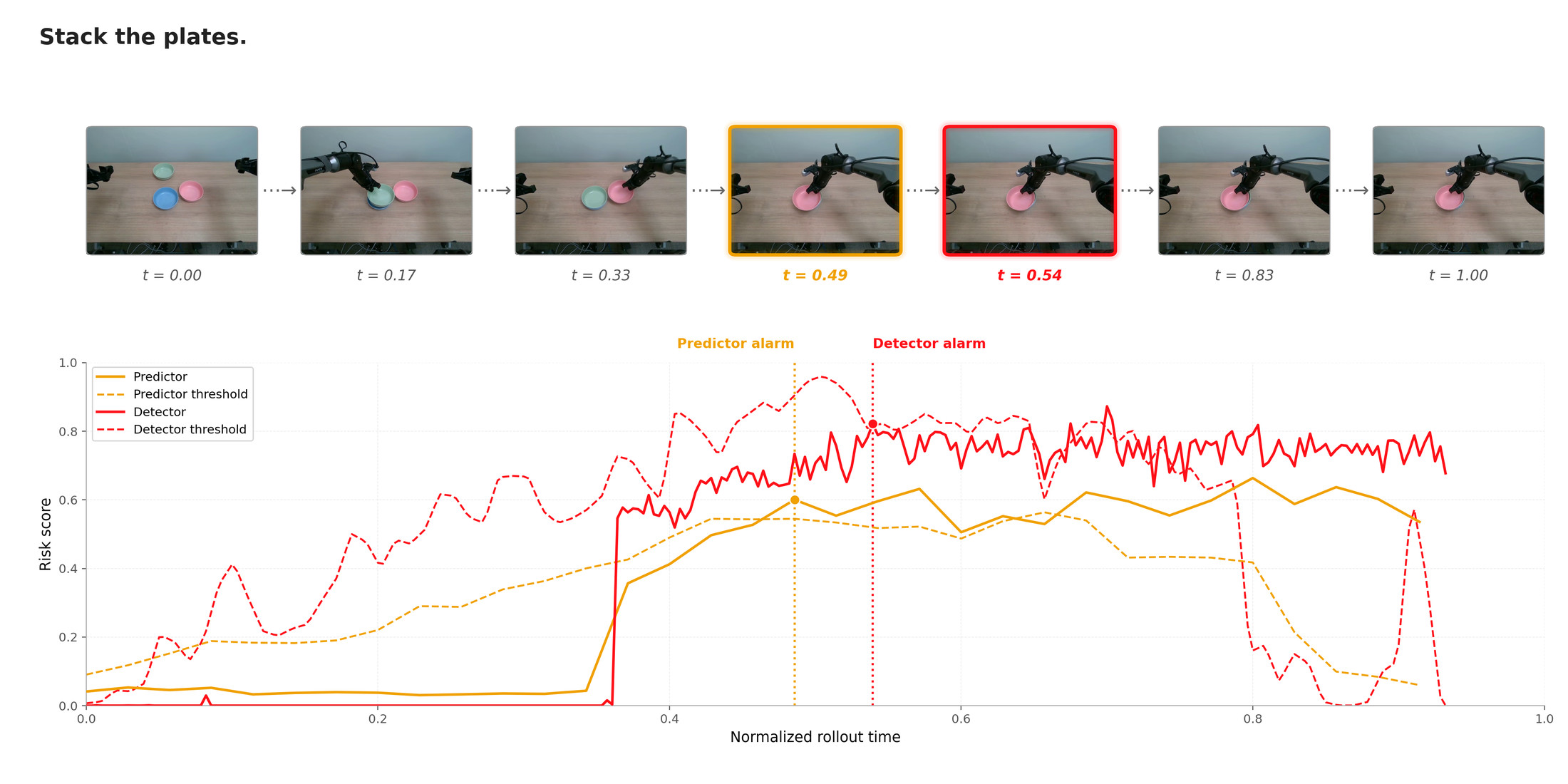}
    \caption{Qualitative monitoring example on Stack Plates with $\pi_{0.5}$. The robot stalls while attempting to grasp the plate. The predictor raises an alarm first, followed by the detector.}
    \label{fig:qual_stack_both_2}
\end{figure}

\begin{figure}[!htbp]
    \centering
    \includegraphics[
        width=0.88\linewidth
    ]{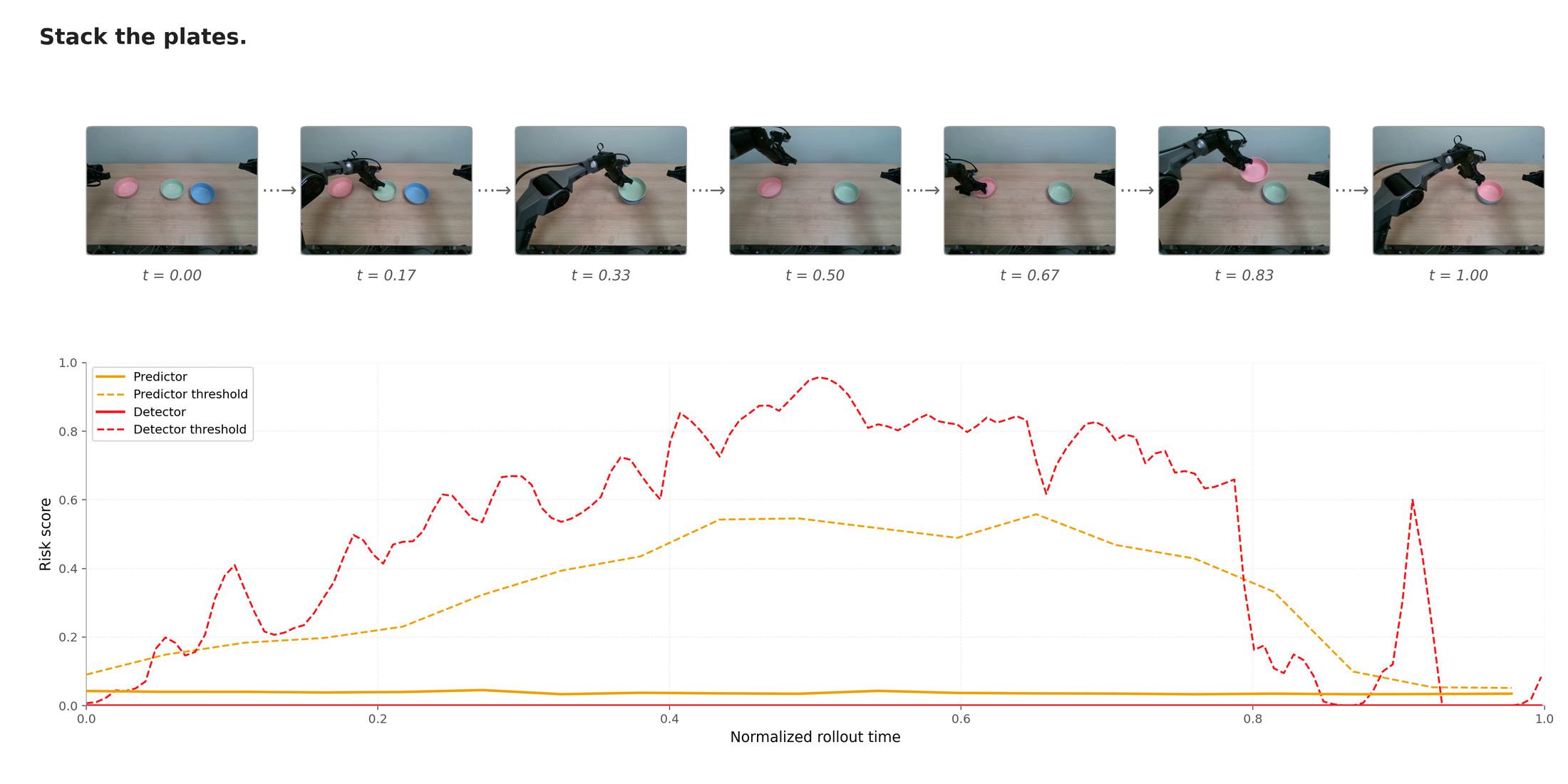}
    \caption{Qualitative monitoring example on Stack Plates with $\pi_{0.5}$. The robot successfully stacks the plates together.}
    \label{fig:qual_stack_success_1}
\end{figure}

\begin{figure}[!htbp]
    \centering
    \includegraphics[
        width=0.88\linewidth
    ]{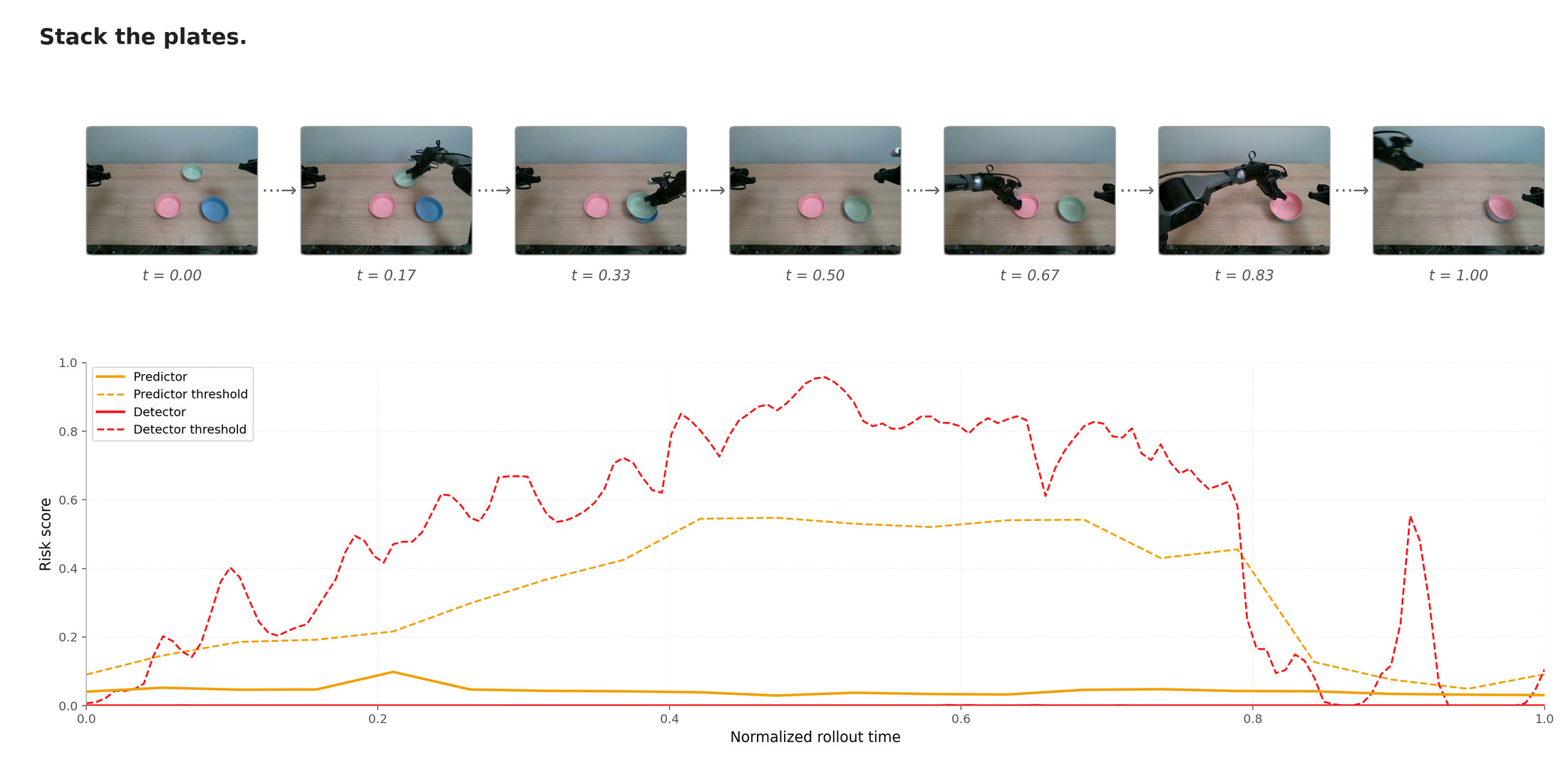}
    \caption{Qualitative monitoring example on Stack Plates with $\pi_{0.5}$. The robot successfully stacks the plates together.}
    \label{fig:qual_stack_success_2}
\end{figure}

\begin{figure}[!htbp]
    \centering
    \includegraphics[
        width=0.88\linewidth
    ]{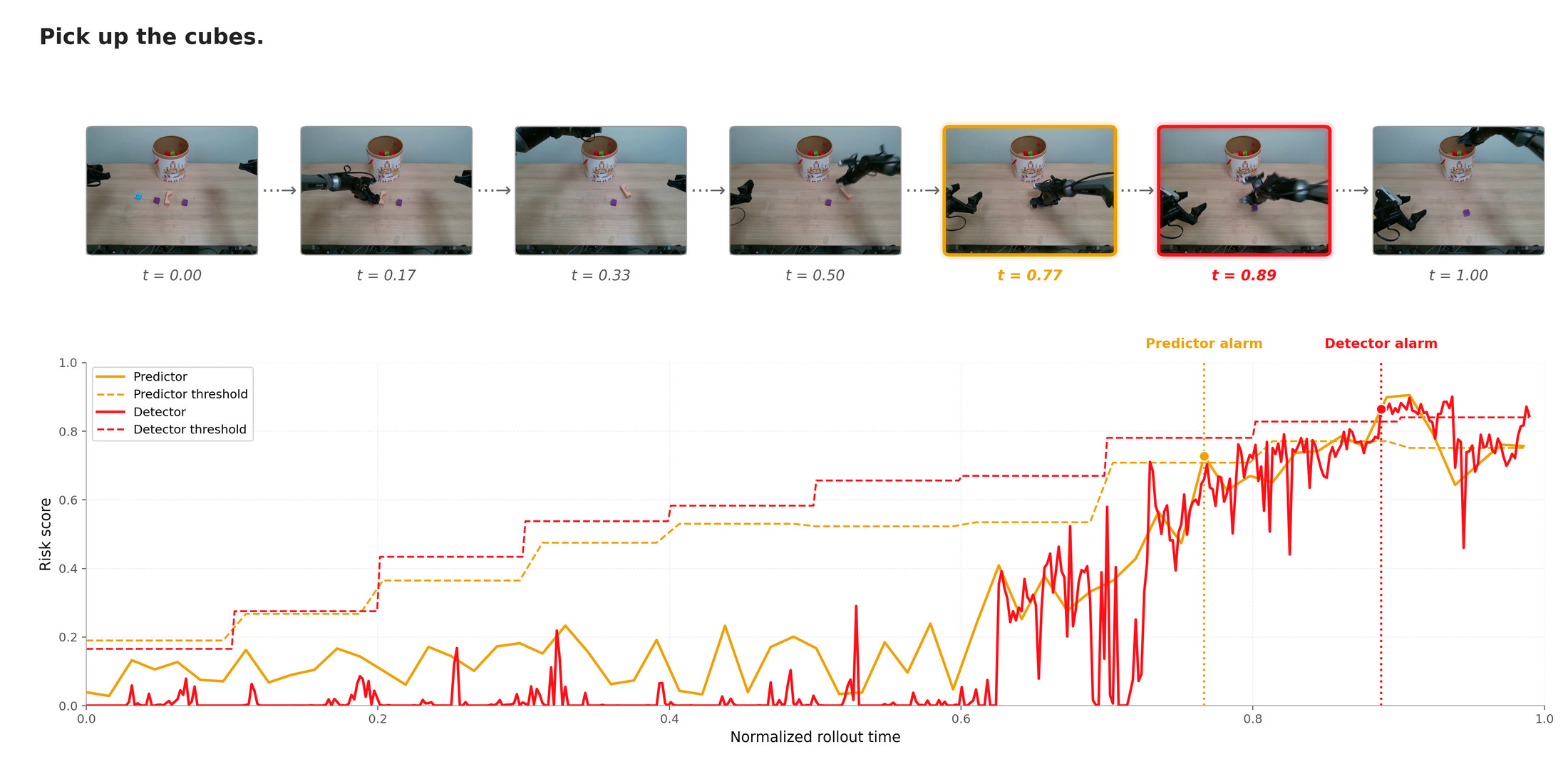}
    \caption{Qualitative monitoring example on Pick Cubes with $\pi_{0.5}$. The robot leaves a cube on the table after further grasp attempts. The predictor raises an alarm first, followed by the detector.}
    \label{fig:qual_pick_both_1}
\end{figure}

\begin{figure}[!htbp]
    \centering
    \includegraphics[
        width=0.88\linewidth
    ]{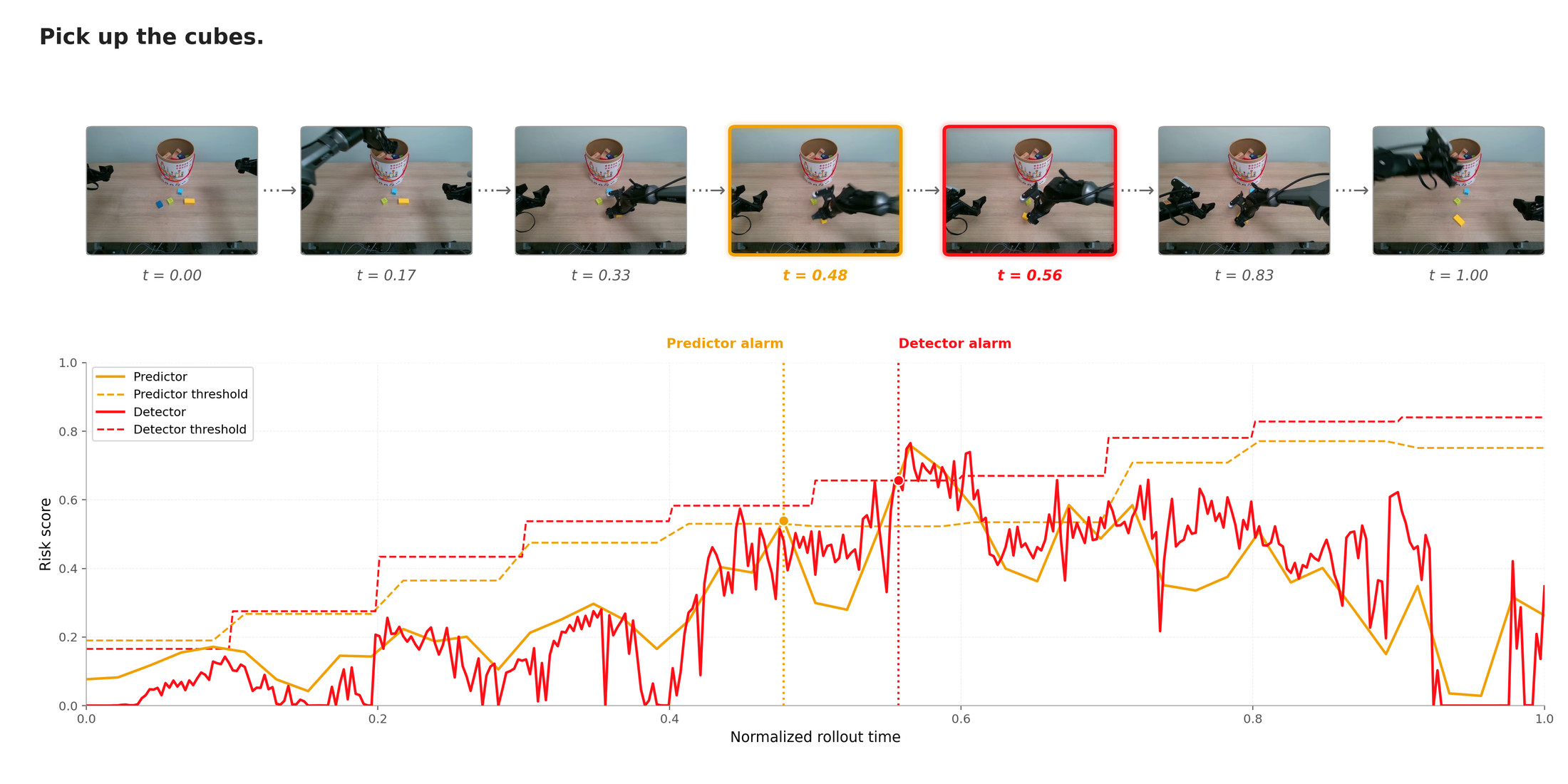}
    \caption{Qualitative monitoring example on Pick Cubes with $\pi_{0.5}$. Repeated grasp attempts leave cubes outside the container. The predictor raises an alarm first, followed by the detector.}
    \label{fig:qual_pick_both_2}
\end{figure}

\begin{figure}[!htbp]
    \centering
    \includegraphics[
        width=0.88\linewidth
    ]{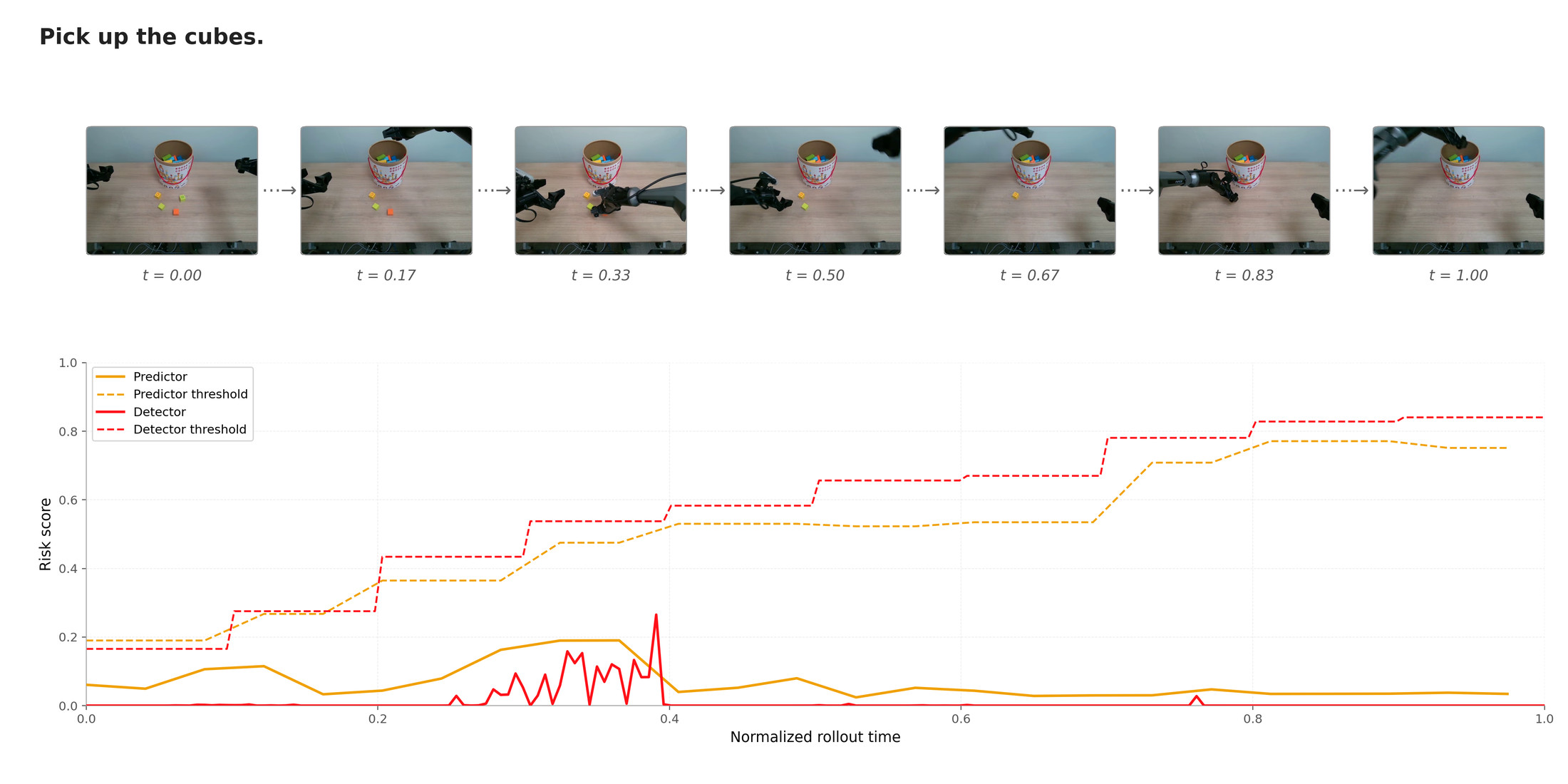}
    \caption{Qualitative monitoring example on Pick Cubes with $\pi_{0.5}$. The robot successfully picks up the cubes and transfers them into the container.}
    \label{fig:qual_pick_success_1}
\end{figure}

\begin{figure}[!htbp]
    \centering
    \includegraphics[
        width=0.88\linewidth
    ]{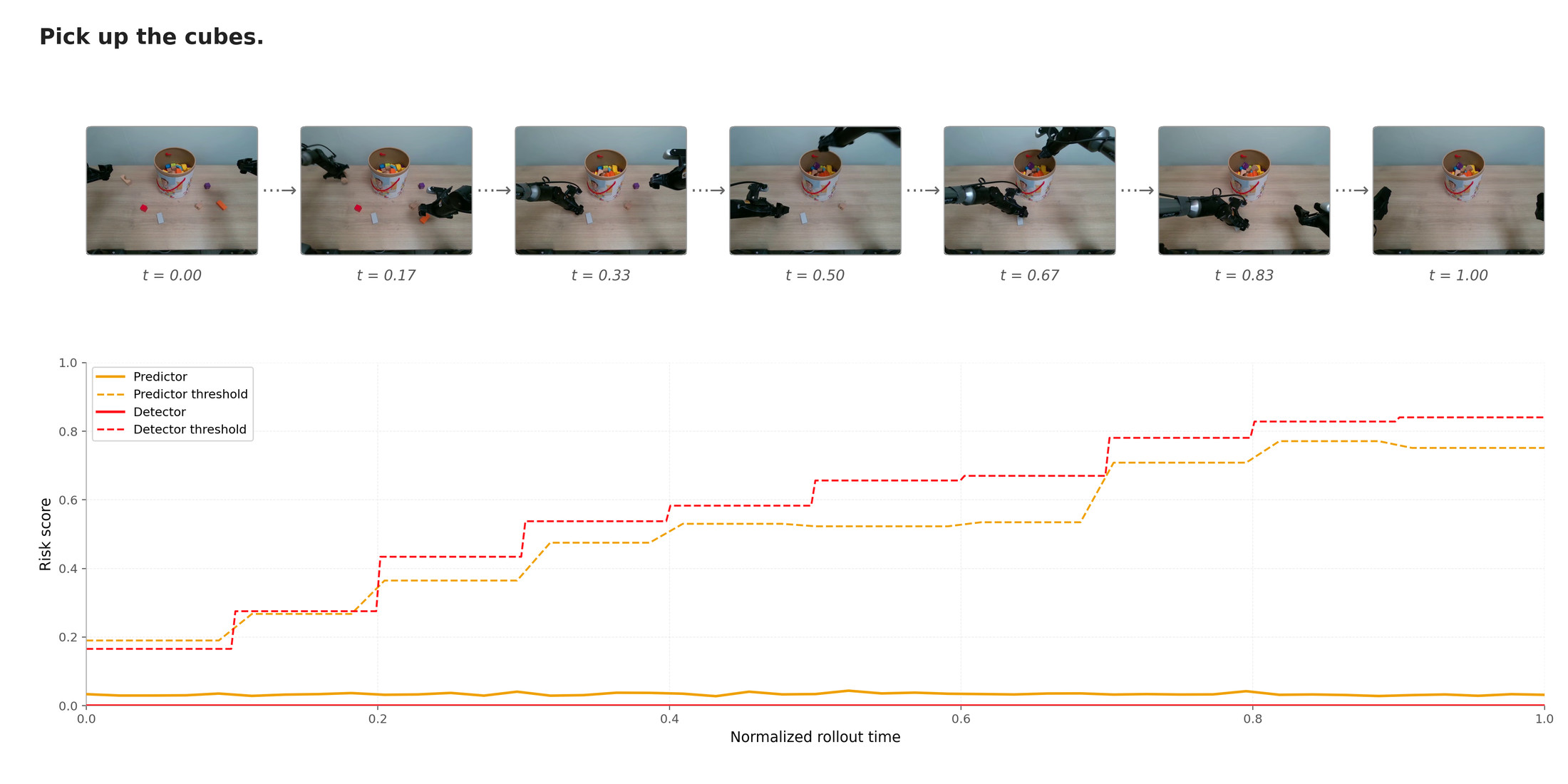}
    \caption{Qualitative monitoring example on Pick Cubes with $\pi_{0.5}$. The robot successfully picks up the cubes and transfers them into the container.}
    \label{fig:qual_pick_success_2}
\end{figure}

\begin{figure}[!htbp]
    \centering
    \includegraphics[
        width=0.88\linewidth
    ]{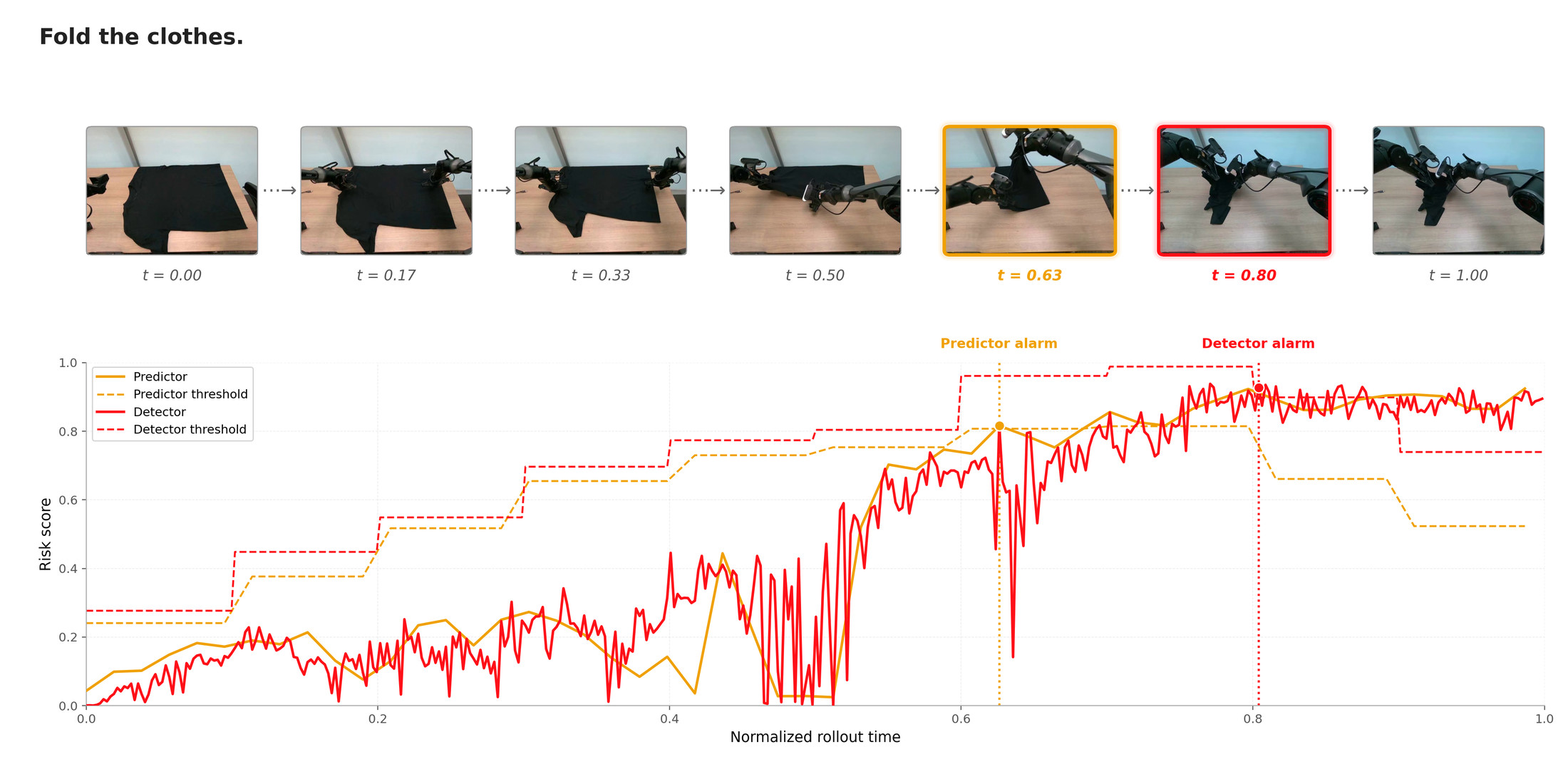}
    \caption{Qualitative monitoring example on Fold Cloth with $\pi_{0.5}$. The cloth slips from the right gripper. The predictor raises an alarm first, followed by the detector.}
    \label{fig:qual_fold_both_1}
\end{figure}

\begin{figure}[!htbp]
    \centering
    \includegraphics[
        width=0.88\linewidth
    ]{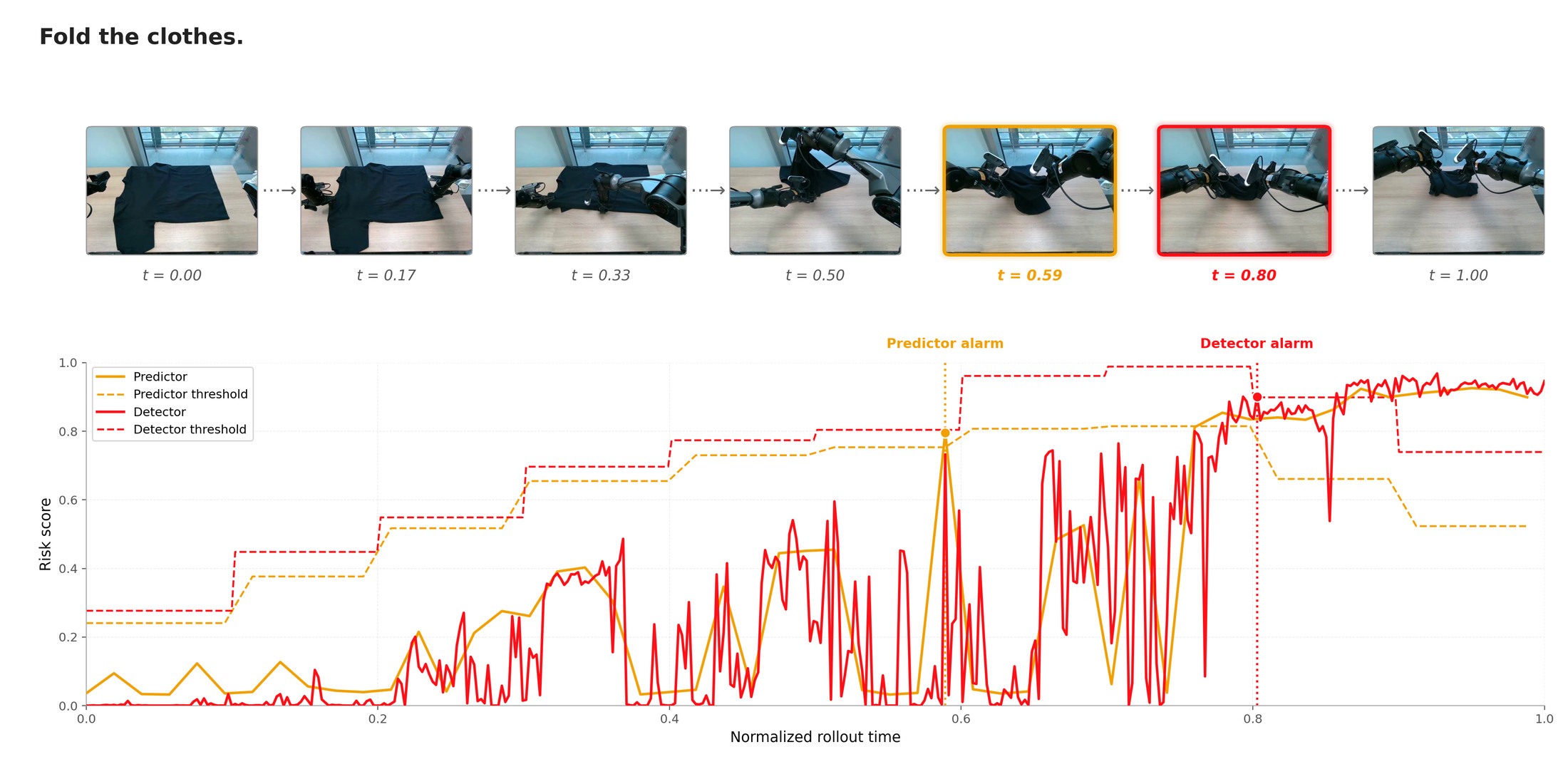}
    \caption{Qualitative monitoring example on Fold Cloth with $\pi_{0.5}$. The cloth bunches between the grippers. The predictor raises an alarm first, followed by the detector.}
    \label{fig:qual_fold_both_2}
\end{figure}

\begin{figure}[!htbp]
    \centering
    \includegraphics[
        width=0.88\linewidth
    ]{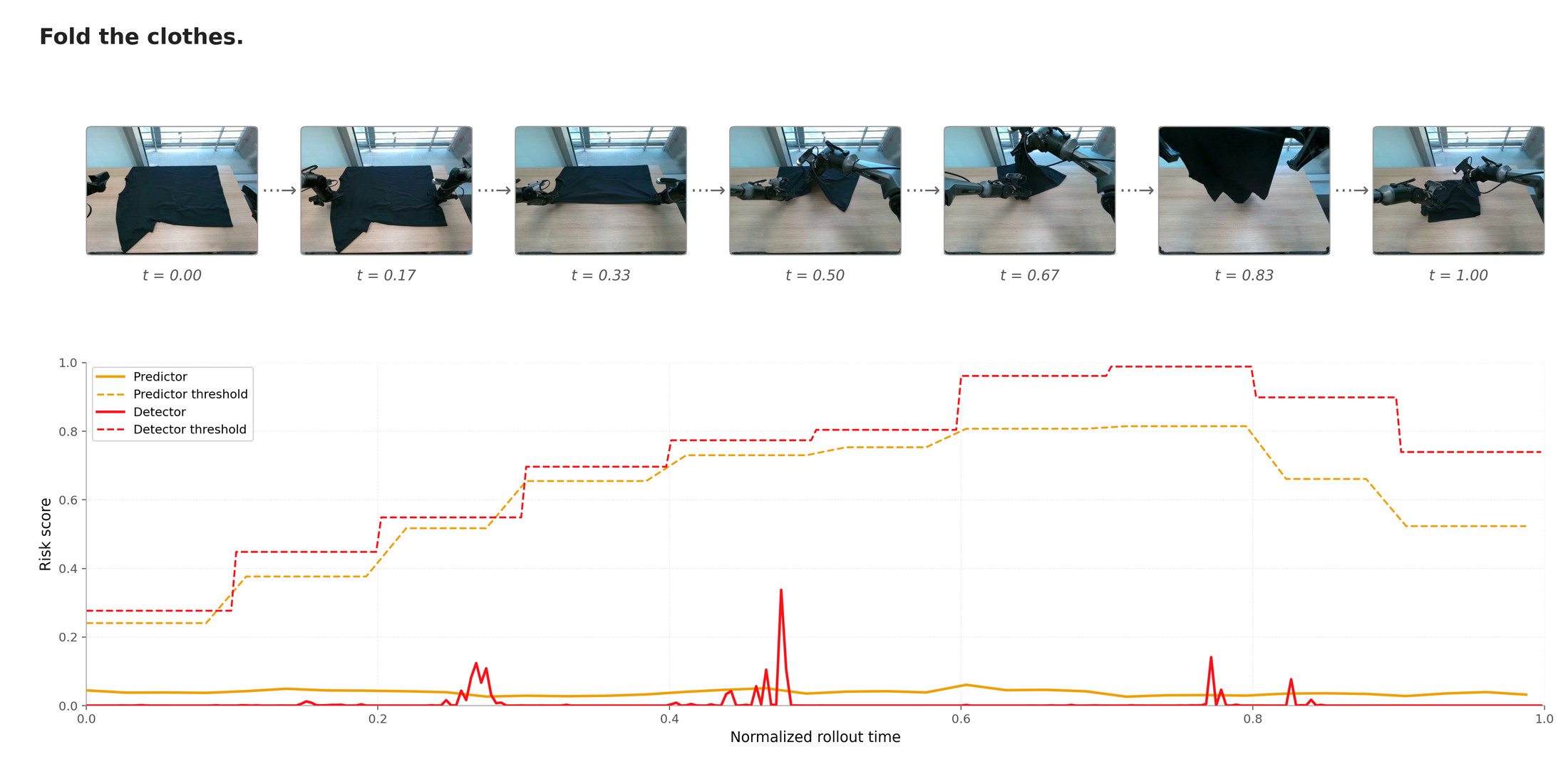}
    \caption{Qualitative monitoring example on Fold Cloth with $\pi_{0.5}$. The robot successfully folds the cloth through coordinated lifting and placement.}
    \label{fig:qual_fold_success_1}
\end{figure}

We additionally visualize rollouts from RoboTwin-8~(Figures~\ref{fig:qual_robotwin_both_1}--\ref{fig:qual_robotwin_success}) and the three real-world tasks, Stack Plates~(Figures~\ref{fig:qual_stack_both_1}--\ref{fig:qual_stack_success_2}), Pick Cubes~(Figures~\ref{fig:qual_pick_both_1}--\ref{fig:qual_pick_success_2}), and Fold Cloth~(Figures~\ref{fig:qual_fold_both_1}--\ref{fig:qual_fold_success_1}). Within each setting, all visualized rollouts use the same calibration variant and significance level $\alpha$. Detector and predictor thresholds are calibrated separately and then held fixed across the visualized rollouts in that setting.


\FloatBarrier


\section{Representation Visualization}

\begin{figure}[!htbp]
    \centering
    \includegraphics[
        width=0.92\linewidth
    ]{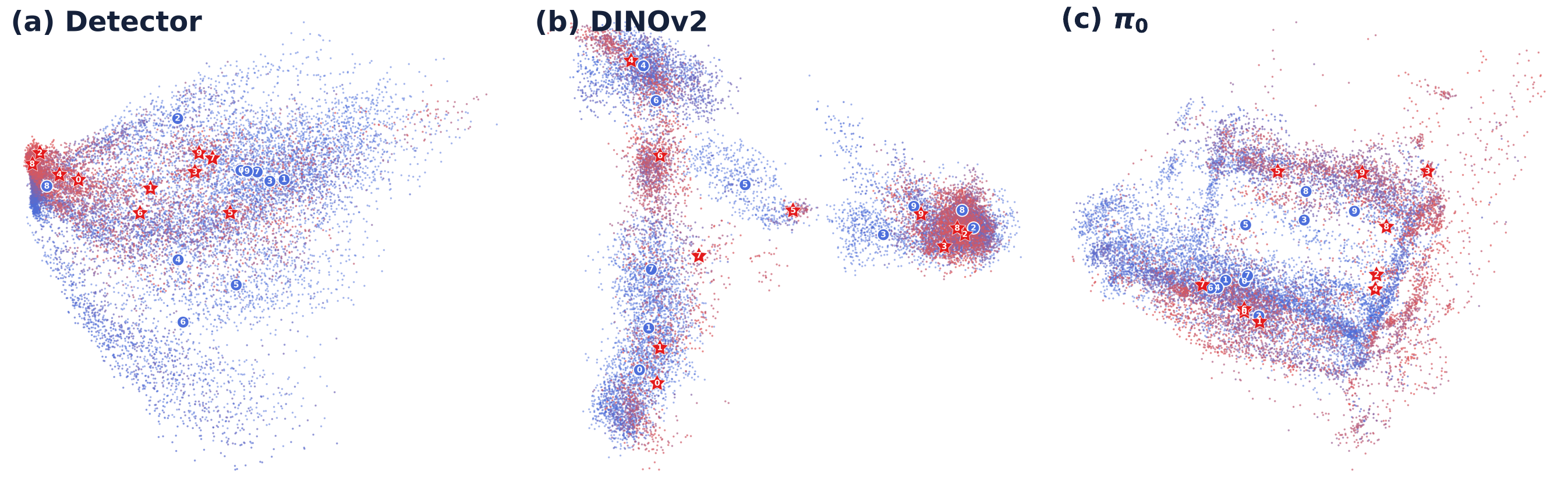}
    \caption{
    PCA visualization of representations on $\pi_0$/LIBERO-10.
    }
    \label{fig:pi0_libero}
\end{figure}

\begin{figure}[!htbp]
    \centering
    \includegraphics[
        width=0.92\linewidth
    ]{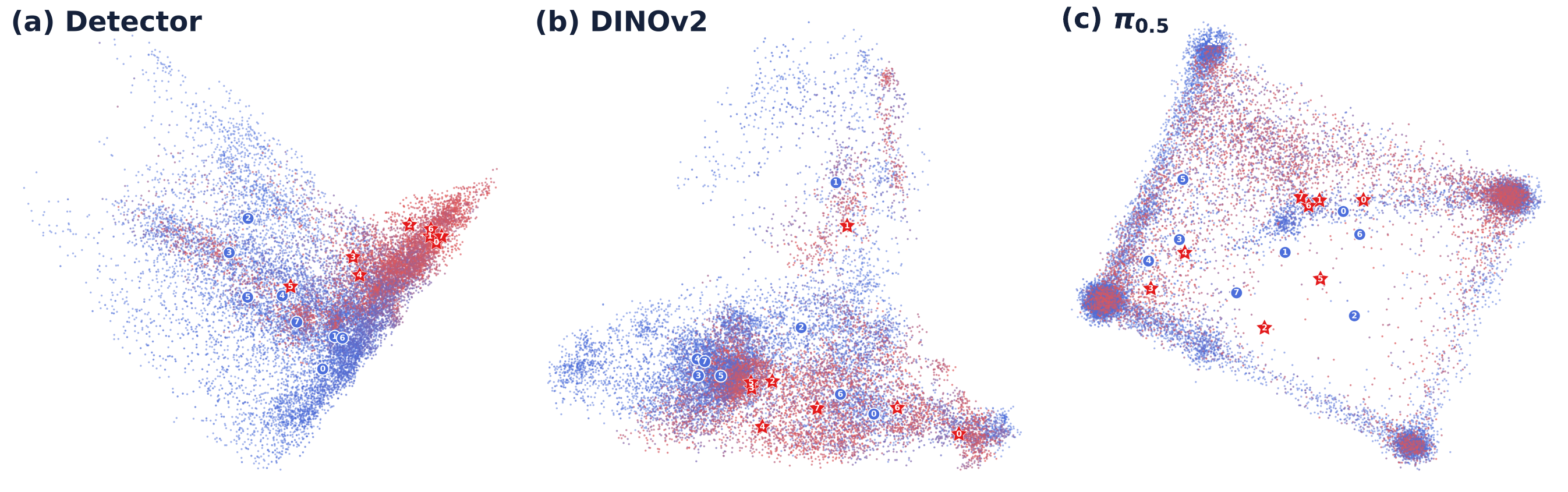}
    \caption{
    PCA visualization of representations on $\pi_{0.5}$/RoboTwin-8.
    }
    \label{fig:pi05_robotwin}
\end{figure}

\begin{figure}[!htbp]
    \centering
    \includegraphics[
        width=0.92\linewidth
    ]{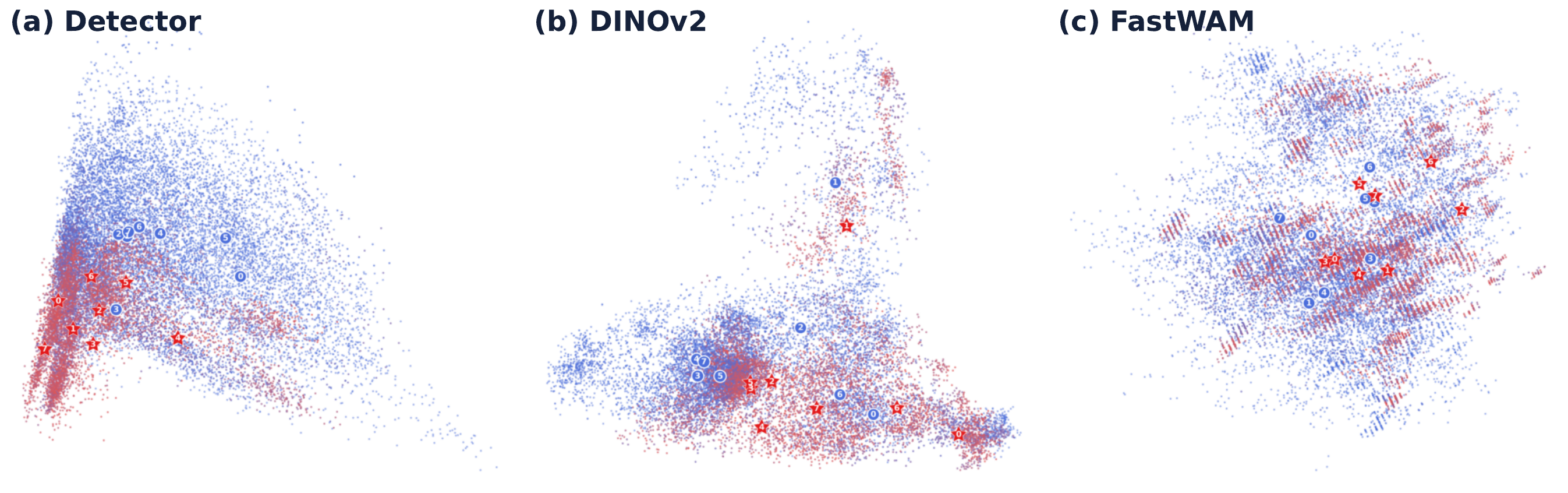}
    \caption{
    PCA visualization of representations on Fast-WAM/RoboTwin-8.
    }
    \label{fig:fastwam_robotwin}
\end{figure}

\FloatBarrier

\end{document}